\documentclass{article}
\usepackage[preprint]{neurips_2026}

\usepackage[utf8]{inputenc} 
\usepackage[T1]{fontenc}    
\usepackage{hyperref}       
\usepackage{url}            
\usepackage{booktabs}       
\usepackage{amsfonts}       
\usepackage{nicefrac}       
\usepackage{microtype}      
\usepackage{xcolor}         
\usepackage{colortbl}       

\usepackage{graphicx}
\usepackage{algorithm}
\usepackage{algpseudocode}
\usepackage{amsmath}
\usepackage{amssymb}
\usepackage{amsthm}
\usepackage{subcaption}
\usepackage{wrapfig}
\usepackage{enumitem}
\usepackage{xspace}
\usepackage[normalem]{ulem}

\newcommand{\methodstyle}[1]{\textsc{#1}\xspace}
\newcommand{\TabKDE}{\methodstyle{TabKDE}}
\newcommand{\CoresetTabKDE}{\methodstyle{CoreTabKDE}}
\newcommand{\RandomCoresetTabKDE}{\methodstyle{RandCoreTabKDE}}
\newcommand{\VAESimpleKDE}{\methodstyle{VAESimpleKDE}}
\newcommand{\CopulaDiff}{\methodstyle{CopulaDiff}}
\newcommand{\VAETabKDE}{\methodstyle{VAETabKDE}}
\newcommand{\simpKDE}{\methodstyle{simple-KDE}}
\newcommand{\TabSYN}{\methodstyle{TabSYN}}
\newcommand{\SMOTE}{\methodstyle{smote}}
\newcommand{\TabDiff}{\methodstyle{TabDiff}}
\newcommand{\PGETabsyn}{\methodstyle{PGE-Tabsyn}}
\newcommand{\FreqTabsyn}{\methodstyle{Freq-Tabsyn}}
\newcommand{\UniTabsyn}{\methodstyle{Uni-Tabsyn}}
\newcommand{\FreqTabKDE}{\methodstyle{Freq-TabKDE}}
\newcommand{\UniTabKDE}{\methodstyle{Uni-TabKDE}}
\newcommand{\PGE}{\methodstyle{PGE}}

\usepackage{pifont}  
  
\newcommand{\kde}{\textsc{kde}}  
\def\Cat{\mathsf{Cat}}
\def\Num{\mathsf{Num}}
\def\Ord{\mathsf{Ord}}
\def\R{\mathbb{R}}
\def\TT{\mathcal{T}}
\def\eps{\varepsilon}

\title{TabKDE:  Simple and Scalable Tabular Data Generation with Kernel Density Estimates}

\author{ 
   Meysam Alishahi \\ 
        University of Utah \\
  \And
   Yan Zheng\\
  Visa Research\\
  \And 
   Junpeng Wang \\ 
  Visa Research\\
  \And
   Chin-Chia Michael Yeh \\ 
  Visa Research\\
  \And
   Jeff M. Phillips\\
    University of Utah \\
}

\begin{document}

\maketitle

\begin{abstract}
    Tabular data generation considers a large table with multiple columns -- each column comprised of numerical, categorical, or sometimes ordinal values. The goal is to produce new rows for the table that replicate the distribution of rows from the original data -- without just copying those initial rows.  The last 4 years have seen enormous progress on this problem, mostly using computational expensive methods that employ one-hot encoding, VAEs, and diffusion.  
    
    This paper describes a new approach to the problem of tabular data generation.   By employing copula transformations and modeling the distribution as a kernel density estimate we can nearly match the accuracy and leakage-avoidance achievements of the previous methods, but with almost no training time.  Our method is very scalable, and can be run on data sets orders of magnitude larger than prior state-of-the-art on a simple laptop.  Moreover, because we employ kernel density estimates, we can store the model as a coreset of the original data -- we believe the first for generative modeling -- and as a result, require significantly less space as well.  Our code is available here:  \url{https://github.com/tabkde/tabkde-main}
\end{abstract}

\section{Introduction}\label{sec:intro}

Tabular data is a fundamental format in many domains, including finance, healthcare, and social sciences, and has seen much recent attention \citep{Fonseca2023, Assefa2021, HERNANDEZ202228, ouyang2023missdiff}, focusing on challenges in scalability and accuracy in its diverse structural characteristics~\citep{Xu2019, borisov2023language, liu2023goggle}. In fact, there are now several surveys on generative methods for this data~\citep{jiang2026representation,stoian2025survey,shi2025comprehensive}. Unlike image or text data, which follow well-defined spatial or sequential relationships, tabular data consists of mixture of varied features that may be numerical, categorical, or ordinal. This heterogeneity poses difficulties in modeling feature dependencies and joint distributions effectively. Traditional generative approaches, such as GANs and VAEs, have been applied with mixed success, and may need to be paired with careful preprocessing and one-hot encoding, which can lead to an explosion in dimensionality and loss of information~\citep{Xu2019, tabsyn2024}. Moreover, adversarial training in GANs can be unstable~\citep{Arjovsky2017}, while VAEs may struggle to generate realistic samples due to overly restrictive latent space constraints~\citep{dai2018diagnosing}. 

Copula-based data generators~\citep{Patki2016,Majdara2020} provide another approach towards transforming differently structured and scaled columns into common format.  The synthetic data vault (SDV)~\citep{Patki2016} includes generative modeling through a variety of approaches including low-rank modeling, GANs, and vine-copula~\citep{Meyer2021}.  These extensions can achieve improved fidelity, but sometimes at a heavy computational expense.

Diffusion models have recently emerged as powerful generative frameworks, demonstrating impressive performance in domains such as image synthesis and molecular generation~\citep{Ho2020, Rombach2022, dhariwal2021diffusion,Morehead2024,Luo2024}. These models operate by progressively transforming noise into structured data through a learned denoising process. 
Recent advances, such as TabDDPM~\citep{TabDDPM2023}, \TabSYN~\citep{tabsyn2024}, and \TabDiff~\citep{tabdiff2025} have made significant progress in adapting diffusion to the tabular setting. TabDDPM~\citep{TabDDPM2023} applies a diffusion model directly to tabular data, effectively capturing complex distributions but requiring a high number of sampling steps. \TabSYN~\citep{tabsyn2024} introduced a latent-space diffusion approach (similar to stable diffusion model approach~\citep{Rombach2022}): it first encodes categorical features with a one-hot encoding, then invokes a VAE to map to a structured representation before applying a diffusion model. This approach has demonstrated remarkable improvements in synthetic data quality, outperforming previous methods in terms of statistical fidelity. 
\TabDiff~\citep{tabdiff2025} extends this by applying a discrete state-space diffusion for the categorical features.

\subsection{Our Contribution}

\vspace{-2mm}
We propose a new approach to tabular data generation, \TabKDE, that nearly matches the accuracy in distributional statistics of the modern diffusion methods, but is significantly faster and more scalable.      
Notably, it only uses classic tools (carefully assembled): copula transformation, covariance estimation, kernel density estimation.  
\\
{\bf Interpretability} is arguably improved because this simple process ensures each original data feature (a column in the table) always corresponds to a single identifiable coordinate throughout the process.  
\\    
{\bf Scalability and efficiency} is improved over methods (like TABDDPM~\citep{TabDDPM2023} and \TabSYN~\citep{tabsyn2024}) which for categorical features rely on one-hot encoding and need to train an expensive diffusion model.   These falter by running out of memory on datasets with many categories since they must then operate in a very high-dimensional setting. \TabKDE avoids these issues by requiring each original feature is encoded in exactly one dimension.  
\\
{\bf Data leakage} is avoided in that \TabKDE produces similar data distributions as the input without replicating (or near-replicating) input data points.   In contrast \SMOTE~\citep{smote2002}, which also does not have a training step, often generates data too close to the data it was trained on.

\begin{figure}[b]
    \vspace{-5mm}
    \begin{center}
        \includegraphics[width=.9\linewidth]{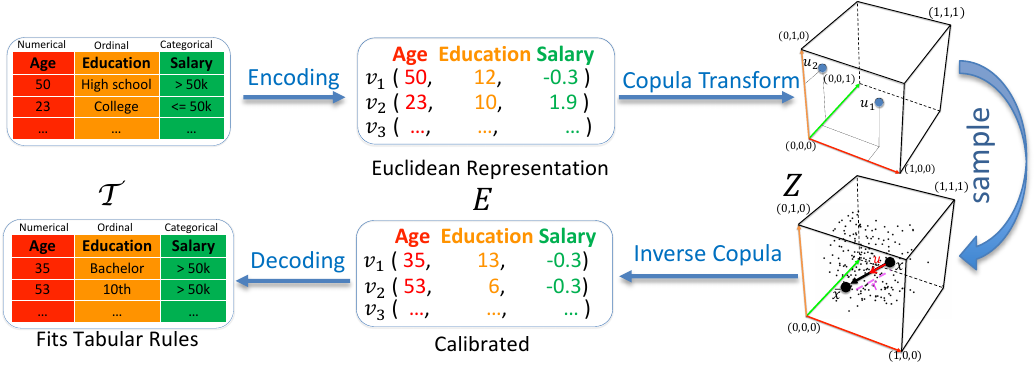}
    \end{center}

    \vspace{-5mm}
    \caption{An overview of the proposed \TabKDE.  
    }
    \label{fig:TabKDE-overview}
\end{figure}

\textbf{Overview of our approach: \TabKDE.}
We follow the general three step paradigm of \citep{Rombach2022}, applied to tabular data.  
\textbf{(1) Encoding} converts the input into a standardized continuous representation.  After this step each of categorical, ordinal, and numerical features are then represented in the same continuous format with well-defined order.  
\textbf{(2) Embedding into a Distance-Aware Latent Space} uses a continuous mapping into another continuous space where now Euclidean distance between objects is representative of how similar they are.  
\textbf{(3) Generative Modeling} maps the discrete distribution of training data in the latent space to a continuous distribution from which we can sample from.  
The samples are then made in the latent space, inverted to the encoded format, and decoded to be in the format of the input table.  
Figure \ref{fig:TabKDE-overview} illustrates how we implement this. 

Our method \TabKDE first converts all features—numerical, ordinal, and categorical—into a unified numerical format; importantly, it does so in a careful way, so each column in the table is represented as a single numerical value.  This is important for two reasons: first, it ensures the dimensionality of the latent space does not blow up, and second, it allow us to carefully control the marginal distributions.  
To this end, this encoding is followed by a copula-based transformation that maps the data into a unit hypercube for easier (marginal) density estimation.  Here its covariance is also calibrated.  
Then the generative process uses Kernel Density Estimation (KDE) modeling to represent and sample from a distribution.  The only training is in learning the shape of the kernel to match the distance to closest record.  Note that sampling from a KDE is simple, in that it needs only to choose a data point, a (covariance scaled) direction, and an offset distance.  
Then generated samples are mapped back by inverting the copula transform, and decoded.  One key additional step is performed to ensure all samples are within the margins identified by the copula, otherwise partial resampling is performed.

\TabKDE avoids one-hot encoding (which can blow up memory requirements), VAEs and diffusion (which can be slow to train), to achieve a scalable method which can still achieve high accuracy and avoid leakage.  
Our extensive experiments swap out each step with expensive and opaque methods (e.g., VAEs, diffusion) and we observe these typically decrease accuracy and are less efficient.

\section{TabKDE Algorithm}\label{sec:tabkde}

\vspace{-3mm}
We consider a tabular dataset $\mathcal{T} = \{X_1, X_2, \dots, X_m\}$, where each row $X_i$ represents an independent and identically distributed (i.i.d.) sample from an unknown joint distribution $P(X)$. 
Each row $X_i = (x_1, x_2, \dots, x_d)$ is a $d$-dimensional vector where each feature $x_j$ belongs to one of three categories: Numerical ($\Num$), Ordinal ($\Ord$), and Categorical ($\Cat$).  
Each numerical feature is in $\R$, but both categorical and ordinal features come from a discrete domain; the difference being that ordinal features have a specified ordering (e.g., grades $A > B > C$).  
    
\subsection{Order Encoding of Tabular Features: $\TT \to E$}
\label{subsec:initialencoding}

\vspace{-2mm}
For tabular data $\TT =\{X_1,\ldots, X_n\}$ with $d$ coordinates, we first encode each row $X_i$ into a space $E \subset \R^d$.  Importantly the space $E$ has one dimension for each column in $\TT$.  
Numerical values are left as they are, and ordinal values are assigned values $1, 2, 3, \ldots$ for the rank of the ordinal category.  

For categorical data we avoid one-hot encoding, and instead apply \textsc{PrincipalGuidedEncoding} (\PGE); detailed in Algorithm \ref{app:alg:categorical_embedding} in Appendix \ref{app::subsec:initialencoding}.  It first computes a vector $v \in \R^{d'}$, the top principal vector of the $d_1 \leq d$ numerical features matrix $X_\Num \in \R^{n \times d_1}$ ($z$-score normalized); this captures the largest mode of variation among the numerical data.  
Then all of the data $X_\Num$ is projected onto this top principal direction $v$, yielding an ``best effort" $1$-dimensional representation of each row $u = X_\Num v \in \mathbb{R}^{n}$. Then for each categorical dimension $j \in \Cat$ with category set $C_j$, it uses $u$ to numerically encode each category. That is, for each category $c \in C_j$ it represents $j$ as $v_{j,c}$ the average of all values with that category.  The ordering of these values $v_{j,c}$ thus provides an ordering of categories $c \in C_j$ which best corresponds with global variation of the non-categorical data, and does not increase the dimension of the representation.  
These categorical dimension can be retokenized by randomly rounding to one of the two nearest values proportional to how close it is.

To our knowledge, this is a new method in this context of generative modeling of tabular data, but similar ideas appearing in applied work (c.f.~\citep{Saukani2019}).  We tested other ordering options in Appendix \ref{app:ablation-cat-encode} and found they do not outperform \PGE, and are sometimes worse.

\subsection{Map to Numerical Latent Space: $E \to Z$ with Copula Transform}
\label{subsec:maptoE}

\vspace{-2mm}
Next we map to a latent space $Z$ where the native distance measures closeness between objects; we employ the classic method of a copula transform~\citep{Patki2016,nelsen2006introduction,rosenblatt1952remarks}: for each $j$th coordinate we build a CDF $F_j$ on the encoded training data points $E_{i,j}$, then the $i$th point is assigned as its $j$th coordinate in the latent space $Z_{i,j} = F_j(E_{i,j})$ as its value in $[0,1]$ from the empirical CDF $F_j$.  
That is, $Z_{i,j}$ is the fraction of $E_{i',j} \leq E_{i,j}$.  A more detailed description is in Appendix \ref{app:EtoZ}.  
These copula transforms have nice properties: they spread out the data in the $[0,1]$ domain, can handle mixed units since they are \emph{only dependent on the ordering}, and are very efficient~\citep{Patki2016}.  
We store the sorted order of the values $E_{i,j}$ so we can invert this.

This copula representation also captures data covariance, and we calibrate to this.  
After having built $Z$ via a copula map in each coordinate, we then compute the sample covariance $\Sigma$ of $Z$.  
This induces a Mahalanobis distance, and $\Sigma$ will be important for generative sampling.

\subsection{Learning Distance to Closest Record (DCR)}\label{subsec:dcrdefin}

\vspace{-2mm}
In generative sampling, one often computes the  \emph{distance to closest record (DCR)}~\citep{MateoSanz2004, Steier2025} to evaluate how similar a synthetic record $s$ is to a real one from a set $Z$. 
\[
\text{DCR}(s, Z) = \min_{z \in Z} \|s-z\|
\]
A DCR of $0$ may indicate an identical match, posing a significant privacy risk~\citep{chen2020gan}. We can compare DCR values from synthetic data to both training $Z_T$ and holdout $Z_H$ datasets. Ideally, for synthetic data $S$,  DCR distributions $\{\text{DCR}(s,Z_T)\}_{s \in S}$ and $\{\text{DCR}(s,Z_H)\}_{s \in S}$ should heavily overlap, showing that synthetic data reflects general patterns rather than replicating specific records.

As part of our generative process, we learn this distribution 
$\{\text{DCR}(s,Z)\}_{s \in S}$ 
where $Z$ is the training data in the copula space. 
Then we generate synthetic data to mirror this scale of variation.  
We repeatedly randomly split the data, and compute the DCR distribution between splits.  We fit a simple mixture of $k$ Gaussians $f$; 
choosing $k$ (from $1, \ldots, 10$) using Bayesian Information Criterion.

We use the learned DCR distribution $f$ as the perturbation-radius in the KDE sampler (Alg.~\ref{alg:SampleKDE}). This estimated $f$ is an empirical model of the nearest-neighbor spacing from the underlying data distribution at finite sample size. Drawing radii $r \sim f$ therefore calibrates the typical displacement of a synthetic point from its seed $z_i$: radii that are too small tend to collapse toward memorization, whereas radii that are too large push samples out of distribution.

\subsection{Tabular Kernel Density Estimation: $Z \to$ Sample}\label{subsec:ZtoSample}

\vspace{-2mm}
A \emph{kernel density estimation (KDE)} is a continuous estimate of a probability density function built by smoothing finite data samples with a kernel function $K$ (often Gaussian) and bandwidth $h$.  For $n$ data points $X = \{x_1, \ldots, x_n\} \sim P$, if we appropriately adjust $h$ as $n$ grows, then the KDE defined $\kde_X(x) = \frac{1}{n} \sum_{i=1}^n K((x - x_i)/h)$ will converge to $P$~\citep{Silverman86,Scott2015}.  
Moreover, we can generate synthetic data (in a manner that approaches the unknown distribution $P$) by drawing a random point $x_i$ and then adding an offset defined by $K$ and $h$.  
 


{\bf Sampling from KDE with DCR kernel.} 
In our \TabKDE method we adapt this sampling from a KDE of the data in a few subtle ways.  First, instead of a Gaussian, we use a kernel to generate offset radius $r$ that matches the learned DCR distribution.  
Second, instead of selecting the offset direction $u$ uniformly, we draw it proportional to the learned covariance $\Sigma$.  These two modifications are sketched in Algorithm \ref{alg:SimpleKDE} with a single sample in latent space generated via Algorithm \ref{alg:SampleKDE}.

\begin{minipage}{0.49\linewidth}
\vspace{-2mm}
\begin{algorithm}[H]
\caption{\textsc{SimpleKDE}($\TT$)}
\label{alg:SimpleKDE}
\begin{algorithmic}[1]
\State $Z \in [0,1]^{n\times d} \Leftarrow$ Copula-Transform($E(\TT)$) 
\State $\Sigma \gets \text{Covariance}(Z)$
\State Estimate empirical DCR distribution $f$
\State \textbf{for} $i= 1,\ldots, m$: 
\State\hspace{0.5cm}\( z'_i = \textsc{SampleKDE}(Z, f, \Sigma) \) 
\State \hspace{0.5cm} \(y_{j} \gets E^{-1}(\textsc{InverseCopula}(z'_{j}))\)
\State \Return \( Y=\{y_1,\ldots, y_m\} \)
\end{algorithmic}
\end{algorithm}
\end{minipage}
\hfill
\begin{minipage}{0.49\linewidth}
\begin{algorithm}[H]
\caption{$\textsc{SampleKDE}(Z,f,\Sigma)
$}
\label{alg:SampleKDE}
\begin{algorithmic}[1]
\State Uniformly sample \( z_i \in Z \)
\State Sample radius \( r > 0 \) from \( f \)
\State Sample \( v \sim \mathcal{N}(0, \Sigma) \), set \( u = \frac{v}{\|v\|} \)
\State \Return  \( z' \gets z_i + r \cdot u \)
\end{algorithmic}
\end{algorithm}
\end{minipage}

However, the \textsc{SimpleKDE} algorithm does not explicitly control the support of the marginals, which is critical for tabular data generation. The copula-transformed representation, however, embeds the data within the unit hypercube, which allows us to control how far a perturbed sampled point \( z' \) can deviate without violating marginal support. We now introduce a more refined rejection-sampling heuristic (Alg. \ref{alg:TabKDE}: \TabKDE) that effectively enforces these boundary constraints.

\begin{minipage}{0.49\linewidth}

\begin{algorithm}[H]
\caption{\textsc{TabKDE}($\TT$)}
\label{alg:TabKDE}
\begin{algorithmic}[1]
\State $Z \in [0,1]^{n\times d} \Leftarrow$ Copula-Transform($E(\TT)$) 
\State $\Sigma \gets \text{Covariance}(Z)$
\State Estimate empirical DCR distribution $f$
\State \textbf{for} $i= 1,\ldots, m$: 
\State\hspace{0.5cm}\( z'_i = \textsc{SampleKDE-iterative}(Z, f, \Sigma) \) 
\State\hspace{0.5cm} \(y_{j} \gets E^{-1}(\textsc{InverseCopula}(z'_{j}))\)
\State \Return \( Y=\{y_1,\ldots, y_m\} \)
\end{algorithmic}
\end{algorithm}
\end{minipage}
\begin{minipage}{0.49\linewidth}
\vspace{-4mm}
\begin{algorithm}[H]
\caption{\textsc{SampleKDE-iterative}($Z,f,\Sigma$)}
\label{alg:SampleKDE-iterative}
\begin{algorithmic}[1]
\State Uniformly sample \( z_i \in Z \)
\State Sample radius \( r > 0 \) from \( f \)
\State Sample \( v \sim \mathcal{N}(0, \Sigma) \), set \( u = \frac{v}{\|v\|} \)
\State \( z' \gets z_i + r \cdot u \)
\State {\bf While} \(\{j : z'_j \notin [0,1] \} \neq \varnothing \): 
    \State \hspace{.5cm} \( J \gets \{j : z'_j \notin [0,1] \} \)
    \State \hspace{.5cm} Sample \( v' \sim \mathcal{N}(0, \Sigma) \), set \( w = \frac{v'}{\|v'\|} \)
    \State \hspace{.5cm} \( s \gets \frac{\| (u_k)_{k \in J} \|}{\| (w_k)_{k \in J} \|} \)
    \State \hspace{.5cm} \( u_j \gets s \cdot w_j \) for each \( j \in J \) 
    \State \hspace{.5cm} \( z' \gets z_i + r \cdot u \)
\State \Return \( z' \)
\end{algorithmic}
\end{algorithm}
\end{minipage}

\TabKDE differs from \textsc{SimpleKDE} only in line 5, where it uses the boundary-aware \textsc{SampleKDE-iterative} (Alg \ref{alg:SampleKDE-iterative}) instead of the simpler \textsc{SampleKDE}. This modified sampler checks for violations of the unit hypercube boundaries and regenerates out-of-bound coordinates. If a valid point cannot be obtained after a fixed number of attempts ($10 \times d$), the sample is discarded, and the process restarts. This mechanism guarantees that all accepted samples lie within the latent space \([0,1]^d\).  
Appendix~\ref{app:correction-stats} shows the iteration is triggered on average less than once per sample, and rarely more than 10 times.  Without this boundary-aware step, resulting data out of distribution; see Figure \ref{app:fig:marginals:simpleVSTabKDE} in Appendix \ref{app:subsec:whyTabKDE}.  

\subsection{Coresets for Generative Tabular Data Modeling}\label{subsec:coreset}

\vspace{-2mm}
A \emph{coreset}~\citep{Phi16} is a compact, weighted set of points that provides a close approximation to the full dataset for a specific downstream task. In the context of KDEs, a coreset serves to approximate the full KDE using significantly fewer, strategically chosen, representative points.

Our proposed \textsc{TabKDE} framework employs the full $\kde_Z$ to generate samples from the Copula latent representation \( Z \subset [0,1]^{n \times d} \) of the tabular data \( \mathcal{T} \). 
To approximate \( \kde_Z(\cdot) \) using a coreset, we define \( \tilde \kde_{\Theta}(\cdot) \) with $\Theta$ comprised of a small set of learnable coreset points \( Q = \{q_1, \ldots, q_m\} \) and their corresponding non-negative weights \( W = \{\omega_1, \ldots, \omega_m\} \), constrained such that \( \sum_{i=1}^{m} \omega_i = 1 \). The approximated density function is
$
\tilde{\kde}_{\Theta}(z) = \sum_{i=1}^{m} \omega_i K(z, q_i),
$
for which a random sample $Q$ with $w_i =1/m$ of size $m=O(1/\eps^2)$ is a coreset so $\|\kde_Z - \tilde \kde_\Theta\|_\infty \leq \eps$~\citet{PT2020}.  
%
%
Moreover, this can be used as a starting point for an optimized coreset, over the  locations $Q$ and their weights $W$ to minimize the empirical $L_2$ via SGD as
$
\mathbb{E}_{z \sim \mathsf{Unif}([0,1]^d)} \left[ \left( \tilde{\kde}_{\Theta}(z) - \kde_Z(z) \right)^2 \right].
$
This optimized version can potentially better preserve key distributional features, such as modes, spread, and overall shape.  Moreover, because we are not replicating the training data, it can reduce the risk of overfitting to the data or leaking training data.  
We call this method \CoresetTabKDE;  and \RandomCoresetTabKDE only samples but does not optimize.  We use coresets of size $m=5000$.

\section{Experimental Results}
\label{sec:experiments}

Our experiments are conducted on the six tabular datasets from UCI Machine Learning Repository\footnote{\url{https://archive.ics.uci.edu/datasets}}  (Adult, Default, Shoppers, Magic, Beijing, News) with between $12$K and $49$K rows and a mixture of $11$ and $48$ dimensions, a mixture of mostly numerical and categorical.  
We also use an IBM  dataset\footnote{\href{https://www.kaggle.com/code/yichenzhang1226/ibm-credit-card-fraud-detection-eda-random-forest/input?select=credit_card_transactions-ibm_v2.csv}{https://www.kaggle.com/code/yichenzhang1226/ibm-credit-card-fraud-detection-eda-random-forest}}
 which is significantly larger; it has about $176$K rows and 14 dimensions, with $5$ ordinal ones, and a total of over $37$K total categories.  See Appendix \ref{app:datasets} for more detail.

\textbf{Baselines. }
We compare our proposed \TabKDE method with several popular baselines, including 
SMOTE~\citep{smote2002},  
GReaT~\citep{borisov2023language}, 
CoDi~\citep{codi2023}, 
TabDDPM~\citep{TabDDPM2023}, 
\TabSYN~\citep{tabsyn2024}, and 
\TabDiff~\citep{tabdiff2025}; some comparisons and other methods are deferred to Appendix \ref{app:baselines} for space.  
%
We also consider several hybrid models that mix elements of \TabKDE with the encoding choices.  Notably, 
in \CopulaDiff we first use our \textsc{CopulaMapping} to embed data into a latent space, then train a diffusion model there, 
and in \VAETabKDE we use \TabSYN's VAE to get latent space $Z$, and then apply the KDE sampler from \TabKDE.  
Broader comparisons with other copula-based methods are in Appendix \ref{app:copula-comapre}.

\begin{table}[t]
\small
\centering
\caption{Runtime comparison of Tabsyn, TabKDE, and SMOTE models across individual datasets on laptop. The IBM dataset is excluded from the average row.}
\begin{tabular}{l|cccc|c|cc}
\toprule
\textbf{Dataset} & \multicolumn{4}{c|}{\TabSYN} & \SMOTE & \multicolumn{2}{c}{\TabKDE}  \\
\midrule
 & VAE Train & Diff. Train & Total Train & Sample & Train+Sample & Train & Sample \\
\midrule
Adult   & 6h 35m 43s & 2h 6m 31s  & 8h 43m 19s & 1m 5s & 4s & 44s & 20s \\
Default & 6h 32m 3s  & 2h 2m 16s  & 8h 34m 59s & 40s   & 2s & 59s & 17s \\
Shoppers & 3h 57m 42s & 0h 55m 32s & 4h 53m 32s & 18s   & 3s & 17s & 5s  \\
Magic   & 3h 51m 7s  & 1h 21m 27s & 5h 13m 0s  & 26s   &  5s & 19s & 7s   \\
Beijing & 5h 31m 57s & 1h 57m 44s & 7h 30m 35s & 54s  & 2s & 35s & 16s  \\
News & 14h 34m 15s & 2h 8m 59s & 16h 44m 11s & 57s   & 4s& 6m 2s & 54s  \\
\midrule
\textbf{Average} & \textbf{6h 50m 27s} & \textbf{1h 45m 24s} & \textbf{8h 36m 36s} & \textbf{43s} & \textbf{3s} & \textbf{1m 29s} & \textbf{19s} \\
\midrule
IBM   & OOM & OOM  & OOM & OOM & OOM & 10m 21s & 6m 4s  \\
\bottomrule
\end{tabular}
\label{tab:runtime_comparison}
\end{table}

\begin{wraptable}{r}{55mm}
\vspace{-12mm}
\centering 
\footnotesize
\caption{Average GPU timing}
\begin{tabular}{lcc}
\hline
\textbf{Method} & \textbf{Train (s)} & \textbf{Sample (s)} \\
\hline
GReaT    & 17112.4 & 251.2 \\
CoDi     &  18487.6 & 11.8 \\
TabDDPM  & 2771.4 & 70.8  \\
\TabSYN   & 1297.8 & 8.4   \\
\hline
\TabKDE   & 39.2   & 39.0  \\
\hline
\end{tabular}
\label{tab:training_sampling_time}
\vspace{-3mm}
\end{wraptable}

\subsection{Scalability and Efficiency}\label{subsec:scalability_efficency}

\vspace{-2mm}
We measure the scalability and efficiency on both the training time, as well as the sample generation time; sample generation measures time for the full synthetic set -- the same size as training set.  
A key motivating factor for generative modeling of tabular data is for making sensitive datasets available for development teams that want to use them to build models without worrying about leaking specific datum.  Since these models vary based on the queried range of the data, regions, and time periods relevant for the analysis, there are frequent retraining on different parts of the full sensitive dataset.  In this setting, each retraining needs to generate one data set that can be used as proxy.

We first compare against \TabSYN and \SMOTE on a laptop using only CPU (2021 Apple 14" MacBook Pro; M1 Pro chip).
Table \ref{tab:runtime_comparison}
shows that the simple \SMOTE algorithm is faster than \TabKDE, but the training time of \TabKDE is orders of magnitude faster than \TabSYN (about 90 seconds to about 8.5 hours).  Appendix \ref{app:scalability} shows other baselines have run times in the ballpark of \TabSYN.  
Moreover, both \TabSYN and \SMOTE run out of memory on the IBM data set since they try to one-hot encode 37K categories, while \TabKDE still completes in under 20 minutes.

We compare GPU runtime (NVIDIA RTX A5000; 24GB memory; max power 230W) in  
Table~\ref{tab:training_sampling_time} over the average training and sampling time on Adult, Default, Shoppers, and Magic datasets.  
\TabKDE is still orders of magnitude faster in training, and while other methods can improve upon \TabKDE (our code is sequential, so not optimized for GPU) in sampling, this cost is dominated by the training time. On the IBM dataset \TabSYN requires 3.5 hours of GPU training and 1 minute for sampling on IBM, whereas \TabKDE trains in 8 seconds and about 7 minutes (433 seconds) to sample.

Among the coreset-based variants that reduce storage (and can improve privacy), \RandomCoresetTabKDE has runtime comparable to \TabKDE.  On the other hand, \CoresetTabKDE has an additional optimization step during training, which we run for on average $55$ minutes measured across Adult, Default, Shoppers, Magic, and Beijing (see Appendix~\ref{app:sec:coreset-eval}), still significantly less than the about $8$ hours required for \TabSYN.  

\subsection{Accuracy}
\label{sec:accuracy}

\vspace{-2mm}
There are many measures of similarity between distributions, and this is complicated by the tabular format of the data.  
We evaluate the quality of the generated synthetic data three ways: (1) marginal distribution alignment, (2) pairwise correlation matching, and (3) finally global alignment between synthetic and hold-out distributions is compared by how well a classifier can separate the distributions.  

\textbf{Marginal distribution alignment.}
When evaluating synthetic tabular data, the marginal distribution alignment score assesses how closely each individual column matches its real‑data distribution represented by training data. Following what was done in the \TabSYN paper, we calculate the Kolmogorov–Smirnov (KS) distance for numerical attributes in $\Num$ and the Total Variation distance for categorical and ordinal attributes in $\Cat$ and $\Ord$.  
Table~\ref{tab:shape} presents, for each dataset, the average marginal alignment errors across all features for each method.
\TabSYN has the best average at $1.42$ while \TabKDE is nearly as good at $1.70$.  
\textsc{SimpleKDE} doesn't preserve marginals in Appendix \ref{app:subsec:whyTabKDE}.

\begin{table}[t!]
\scriptsize
\centering
\caption{Marginal distribution alignment error; lower is better.  In parentheses denotes ratio relative to the smallest value.  
Baseline values, unless stated otherwise, are taken from \citep{tabsyn2024}; the remaining scores (our methods) are obtained using their data split.
}
\begin{tabular}{lcccccc|c}
\toprule
\textbf{Method} & \textbf{Adult} & \textbf{Default} & \textbf{Shoppers} & \textbf{Magic} & \textbf{Beijing} & \textbf{News} & \textbf{Average} \\
\midrule
\SMOTE({\tiny Our reproduced})\hspace{-2mm} & 1.63 ({\tiny 2.55}) & 1.70 ({\tiny 1.49}) & 2.66 ({\tiny 2.16}) & 1.37 ({\tiny 1.93}) & 2.10 ({\tiny 1.62}) & 5.47 ({\tiny 3.18}) & 2.49 ({\tiny 1.75}) \\
GReaT & 12.12 ({\tiny 18.94}) & 19.94 ({\tiny 17.49}) & 14.51 ({\tiny 11.80}) & 16.16 ({\tiny 22.76}) & 8.25 ({\tiny 6.35}) & -- & 14.20 ({\tiny 10.00}) \\
CoDi & 21.38 ({\tiny 33.41}) & 15.77 ({\tiny 13.82}) & 31.84 ({\tiny 25.89}) & 11.56 ({\tiny 16.28}) & 16.94 ({\tiny 13.03}) & 32.27 ({\tiny 18.78}) & 21.63 ({\tiny 15.23}) \\
TabDDPM & 1.75 ({\tiny 2.73}) & 1.57 ({\tiny 1.38}) & 2.72 ({\tiny 2.21}) & 1.01 ({\tiny 1.42}) & 1.30 ({\tiny 1.00}) & 78.75 ({\tiny 45.83}) & 14.52 ({\tiny 10.23}) \\
TabSYN ({\tiny Our reproduced})\hspace{-2mm} &  0.64 ({\tiny 1.00}) &  1.14 ({\tiny 1.00}) & 1.23 ({\tiny 1.00}) &  0.98 ({\tiny 1.38}) &  2.79 ({\tiny 2.15}) &  1.72 ({\tiny 1.00}) & 1.42 ({\tiny 1.00}) \\
\hline
\CopulaDiff &  2.01 ({\tiny 3.14}) &  1.47 ({\tiny 1.29}) &  2.47 ({\tiny 2.01}) &  0.94 ({\tiny 1.32}) &  2.13 ({\tiny 1.64}) &  2.44 ({\tiny 1.42}) &  1.91 ({\tiny 1.35}) \\
\VAETabKDE &  3.80 ({\tiny 5.94}) &  5.84 ({\tiny 5.12}) &  6.31 ({\tiny 5.13}) &  0.71 ({\tiny 1.0}) &  4.94 ({\tiny 3.80}) &  4.45 ({\tiny 2.59}) &  4.34 ({\tiny 3.06}) \\
\RandomCoresetTabKDE &  1.61 ({\tiny 2.52}) &  1.76 ({\tiny 1.54}) &  2.54 ({\tiny 2.07}) &  1.01 ({\tiny 1.42}) &  1.70 ({\tiny 1.31}) &  2.59 ({\tiny 1.51}) & 1.87 ({\tiny 3.48}) \\
\CoresetTabKDE &  3.63 ({\tiny 5.67}) &  3.29 ({\tiny 2.89}) &  3.23 ({\tiny 2.63}) &  1.08 ({\tiny 1.52}) &  3.20 ({\tiny 2.46}) &  2.87 ({\tiny 1.67}) & 2.88 ({\tiny 2.03}) \\
\TabKDE &  1.56 ({\tiny 2.44}) &  1.55 ({\tiny 1.36}) &  2.44 ({\tiny 1.98}) &  0.78 ({\tiny 1.1}) & 1.37 ({\tiny 1.05}) &  2.52 ({\tiny 1.47}) &  1.70 ({\tiny 1.2}) \\
\arrayrulecolor{black}\bottomrule
\end{tabular}
\label{tab:shape}
\end{table}



\begin{figure}[ht]
\centering
\includegraphics[width=0.24\linewidth]{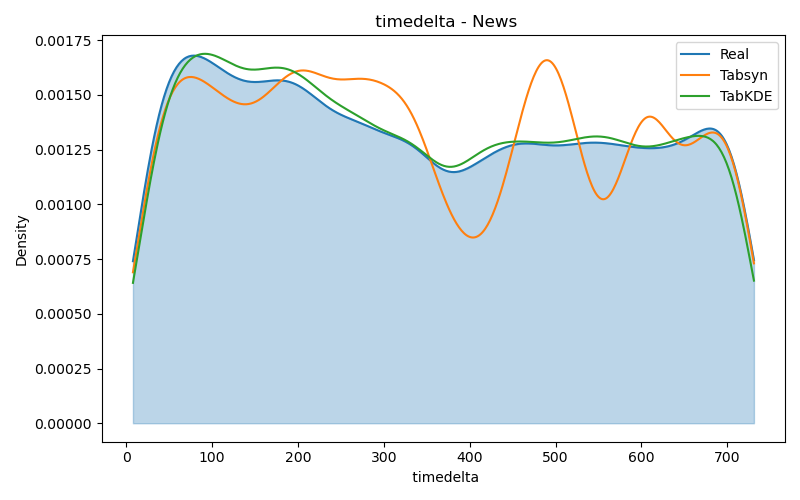}  
\includegraphics[width=0.24\linewidth]{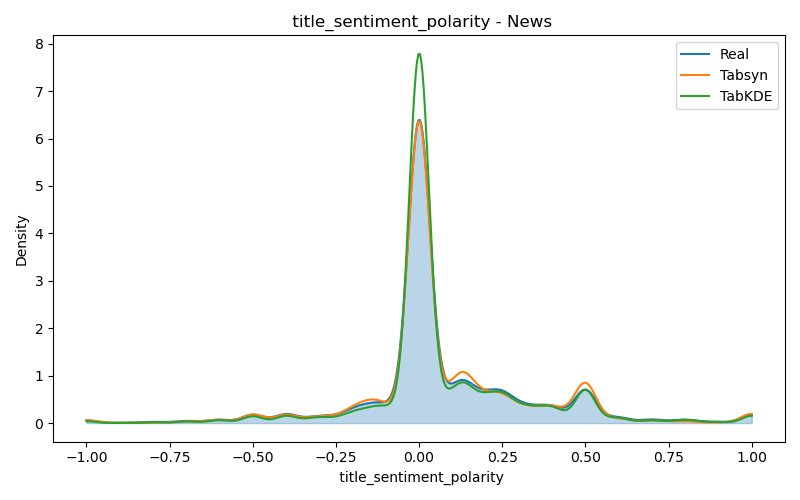} 
\includegraphics[width=0.24\linewidth]{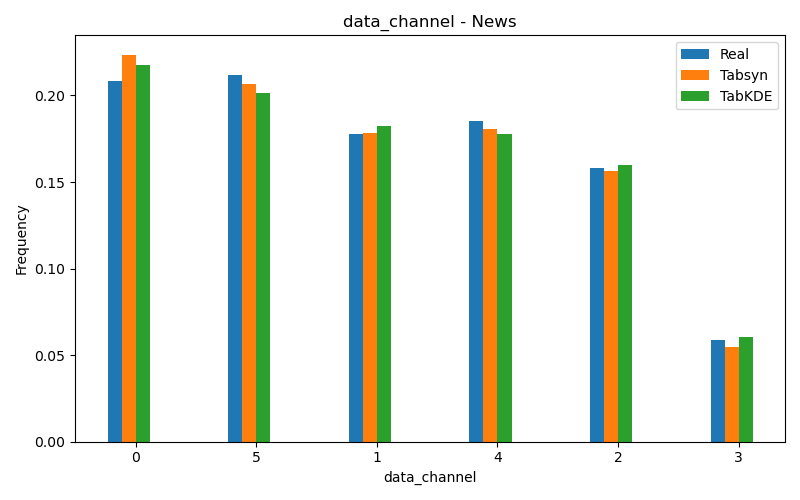}  
\includegraphics[width=0.24\linewidth]{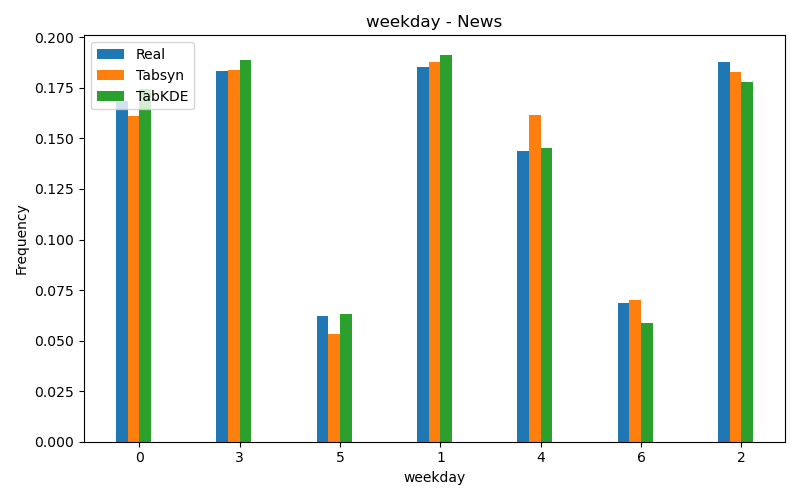} 
\vspace{-3mm}
\caption{Marginals comparison between real data (blue), \TabKDE (green), and \TabSYN (orange).  Representative numerical and categorical data from News dataset. }
\label{fig:marginals:TabsynVSTabKDE}
\vspace{-3mm}
\end{figure}

Figure~\ref{fig:marginals:TabsynVSTabKDE} provides a visual comparison between  some representative selected real marginal distributions and those generated by \TabSYN (orange) and \TabKDE (green) against the real data distributions (blue).  
We observe that \TabKDE and \TabSYN visually match the distributions well, both are about the same.   In particular, on numerical data \TabKDE seems to do better on more uniform distributions whereas \TabSYN does better on spiky ones.

\textbf{Pairwise correlation alignment.} 
We next measure pairwise correlation between columns.  For numerical-numerical pairs, we can use standard Pearson correlations.  For pairs that involve categorical or ordinal feature (as in \citet{tabsyn2024}) we use contingency-table total variation distances.  
In both metrics, smaller error values indicate that the synthetic table is more faithful to the original data. Table~\ref{tab:trend} presents, for each dataset, the average pairwise correlation alignment errors 
across all features for each method.
It also shows a heatmap visualization of the divergence between the pairwise correlations in the real and synthetic data; 
see more in Appendix \ref{app:pairwise}.  
We observe that \TabKDE has better pairwise correlation alignment than all methods except \TabSYN, and is comparable to SMOTE; both \TabKDE and SMOTE have about $2\times$ the correlation discrepancy as \TabSYN.  

We also compare against some variants on the larger IBM dataset (See Appendix~\ref{app:datasets} for more detail).  Recall that on our laptop CPU, neither \TabSYN or \SMOTE can run on this data set -- they both run out of memory; but can run on our larger GPU, although \TabSYN takes about 20 minutes compared to 40 seconds for \TabKDE.  
The average pairwise correlation alignment error for \TabSYN is 40\%, while for \TabKDE and \CopulaDiff it is 27\% and 29\%; see Table \ref{tab:IBM-accuracy-main}.  

\begin{table}[h]
\centering
\small
\caption{Error and time on IBM. }
\begin{tabular}{lcccc}
\toprule
\textbf{Method} & \textbf{Marginal} & \textbf{Pairwise} & \textbf{CPU time} & \textbf{GPU time} \\
\midrule
\TabSYN             &   16.99 &    40.42   & OOM & 3.5 h\\
\CopulaDiff         &   6.81   &    29.42  & 15 h & -\\
\TabKDE    &   4.36   &    27.56  & 7 m & 40 s\\
\bottomrule
\end{tabular}
\label{tab:IBM-accuracy-main}
\end{table}

\begin{table}[h!]
\scriptsize
\caption{Pairwise correlation alignment error; Lower values is better. 
In parentheses is the ratio relative to the smallest value.  Baseline values, unless stated otherwise, are taken from \citep{tabsyn2024}; the remaining scores (our methods) are obtained using their data split.  Below: Representative Pairwise correlation divergence heatmaps for Magic and Beijing datasets.
}
\begin{tabular}{lcccccc|c}
\toprule
\textbf{Method} & \textbf{Adult} & \textbf{Default} & \textbf{Shoppers} & \textbf{Magic} & \textbf{Beijing} & \textbf{News} & \textbf{Average} \\
\midrule
\SMOTE {\tiny (our reproduction)\hspace{-2mm}}  & 4.3 ({\tiny 2.67}) & 11.54 ({\tiny 5.11}) & 3.68 ({\tiny 1.47}) & 1.88 ({\tiny 2.29}) & 3.3 ({\tiny 1.22}) & 1.67 ({\tiny 1.25}) & 4.39 ({\tiny 1.99}) \\
GReaT & 17.59 ({\tiny 10.93}) & 70.02 ({\tiny 30.98}) & 45.16 ({\tiny 18.06}) & 10.23 ({\tiny 12.48}) & 59.6 ({\tiny 21.99}) & -- & 40.52 ({\tiny 18.33}) \\
CoDi & 22.49 ({\tiny 13.98}) & 68.41 ({\tiny 30.27}) & 17.78 ({\tiny 7.11}) & 6.53 ({\tiny 7.97}) & 7.07 ({\tiny 2.61}) & 11.1 ({\tiny 8.28}) & 22.23 ({\tiny 10.06}) \\
TabDDPM & 3.01 ({\tiny 1.87}) & 4.89 ({\tiny 2.16}) & 6.61 ({\tiny 2.64}) & 1.7 ({\tiny 2.07}) & 2.71 ({\tiny 1.00}) & 13.16 ({\tiny 9.82}) & 5.34 ({\tiny 2.42}) \\
\TabSYN{} {\tiny (our reproduction)\hspace{-2mm}} &  1.61 ({\tiny 1.00}) &  2.26 ({\tiny 1.00}) &  2.5 ({\tiny 1.00}) &  0.82 ({\tiny 1.00}) &  4.7 ({\tiny 1.73}) &  1.34 ({\tiny 1.00}) &  2.21 ({\tiny 1.00}) \\
\hline
\CopulaDiff &  4.61 ({\tiny 2.86}) &  3.29 ({\tiny 1.46}) &  5.3 ({\tiny 2.12}) &  1.72 ({\tiny 2.10}) &  4.5 ({\tiny 1.66}) &  2.1 ({\tiny 1.57}) &  3.59 ({\tiny 1.62}) \\
\VAETabKDE &  7.23 ({\tiny 4.49}) &  12.71 ({\tiny 5.62}) &  9.68 ({\tiny 3.87}) &  3.95 ({\tiny 4.82}) &  9.87 ({\tiny 3.64}) &  3.67 ({\tiny 2.74}) &  7.85 ({\tiny 3.55}) \\
\RandomCoresetTabKDE &  3.93 ({\tiny 2.44}) &  13.66 ({\tiny 6.04}) &  4.27 ({\tiny 1.71}) &  4.76 ({\tiny 5.80}) &  4.05 ({\tiny 1.49}) &  2.61 ({\tiny 1.95}) & 5.46 ({\tiny 2.47}) \\
\CoresetTabKDE &  6.3 ({\tiny 3.91}) &  9.91 ({\tiny 4.39}) &  5.77 ({\tiny 2.31}) &  2.18 ({\tiny 2.66}) &  5.86 ({\tiny 2.16}) &  2.82 ({\tiny 2.10}) & 5.47 ({\tiny 2.48}) \\
\TabKDE &  4.51 ({\tiny 2.80}) &  9.93 ({\tiny 4.40}) &  4.31 ({\tiny 1.72}) &  2.72 ({\tiny 3.32}) &  3.74 ({\tiny 1.38}) &  2.83 ({\tiny 2.11}) & 4.67 ({\tiny 2.11}) \\
\bottomrule
\end{tabular}
\label{tab:trend}

\includegraphics[width=\linewidth]{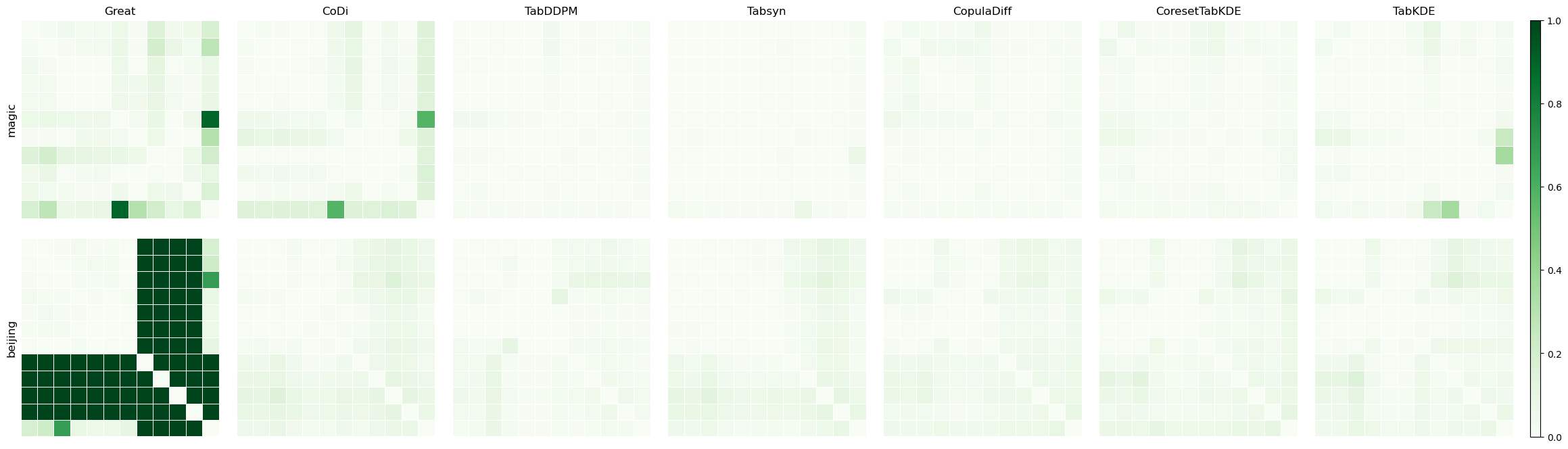}
\end{table}

\paragraph{Global Distribution Alignment.} 
Synthetic data should be able to take the place of real data, letting us train models on it for downstream prediction tasks and have indistinguishable performance.  
We assess by building a classifier to attempt to distinguish between the synthetic data and a split of data held out from the training process.  
We use logistic regression in Table \ref{tab:c2st} and XGBoost in Appendix \ref{app:sec:accuracy}. We quantify this as a classifier two-sample test (C2ST) as provided by SDMetrics; larger values closer to $1$ are better.  
%
We observe that \TabKDE is roughly the same as SMOTE with $0.93$ and only bested by \TabSYN which has about $0.97$.  Other baselines achieve $0.79$ (TabDDPM) or below $0.66$.  

\begin{table}[h!]
\small
\centering
\caption{C2ST Scores; larger is better.  
Baseline values, unless stated otherwise, are taken from \citep{tabsyn2024}; the remaining scores (our methods) are obtained using their data split.
%
%
}
\begin{tabular}{lcccccc|c}
\toprule
\textbf{Method} & \textbf{Adult} & \textbf{Default} & \textbf{Shoppers} & \textbf{Magic} & \textbf{Beijing} & \textbf{News} & \textbf{Average} \\
\midrule
\SMOTE{\tiny (Our reproduction)} & 0.9212& 0.9332& 0.9107& 0.9803 & 0.9972& 0.8633& 0.9334\\
GReaT   & 0.5376 & 0.4710 & 0.4285 & 0.4326 & 0.6893 & ---    & 0.5118 \\
CoDi    & 0.2077 & 0.4595 & 0.2784 & 0.7206 & 0.7177 & 0.0201 & 0.4007 \\
TabDDPM & 0.9755 & 0.9712 & 0.8349 & 0.9998 & 0.9513 & 0.0002 & 0.7888 \\
\TabSYN{\tiny (Our reproduction)}& 0.9949& 0.9804& 0.9699& 0.9893& 0.9268& 0.9584& 0.9699\\
\hline
\CopulaDiff       & 0.8557&	0.9798&	0.8665&	0.9914&	0.9576&	0.9793&  0.9384\\
\VAETabKDE & 0.7483& 0.4828& 0.7242& 0.9984& 0.8022& 0.8075& 0.7606\\
\RandomCoresetTabKDE & 0.9215& 0.9570& 0.8757& 0.9921& 0.9503& 0.8901& 0.9311\\
\CoresetTabKDE & 0.8254& 0.8730& 0.8462& 0.9864& 0.8924& 0.8643& 0.8813\\
\TabKDE & 0.9219& 0.9579& 0.9161& 1.0000& 0.9514& 0.8819& 0.9382\\
\bottomrule
\end{tabular}
\label{tab:c2st}
\end{table}


\subsection{Data Leakage \& Privacy}
\label{sec:privacy}

\vspace{-2mm}
We evaluate prevention of data leakage and preservation of privacy of the training set in synthetic data generation.  Given that our target application is quick one-time use creation of a training data set to mirror protected data in a query, we follow the full black-box model and attack of \cite{chen2020gan}.  This considers the nearest neighbor of synthetic data points to the original data versus a held-out set.  This is just the \emph{DCR score} $p$, as used in \TabDiff~\cite{tabdiff2025}, and shown in Appendix \ref{app:privacy}.    
Ideally these distributions should be indistinguishable, and the score close to $50\%$.  
%
Because there is not a common likelihood function and 
for better interpretability, we reframe this as a membership inference attack~\cite{shi2025comprehensive}, by mapping this to an odds of membership (from train or held-out).  
We map $p$ to \emph{nearest neighbor odds}:  $NNO =\frac{p}{100-p}$, 
which quantifies the strength of this proximity bias (e.g., for \TabKDE $p=58\%$ yields $1.38\times$ odds, while for \SMOTE $p=95\%$ yields $19\times$ odds). 

Table~\ref{tab:dcr_odds} reports DCR as nearest-neighbor odds, where values near $1\times$ indicate that synthetic samples are \emph{not} systematically closer to the training set than to a held-out set (and thus exhibit little geometric ``membership advantage''). Under this lens, \TabKDE\ and its coreset variants remain comparatively benign. In particular, \CoresetTabKDE\ attains near-unity odds across all datasets (average $1.13\times$), indicating only a low preference for training neighborhoods. Even the full \TabKDE  while higher (averages $1.41\times$), stay far below methods that exhibit strong training-set attraction and are orders of magnitude smaller than the extreme nearest-neighbor leakage observed for \SMOTE.

\begin{table}[t]
\caption{DCR expressed as nearest-neighbor odds. If the DCR score is $p$ (in \%), odds are $\frac{p}{100-p}$.}
\centering
\small
\begin{tabular}{lccccc|c}
\toprule
\textbf{Method} & \textbf{Adult} & \textbf{Default} & \textbf{Shoppers} & \textbf{Beijing} & \textbf{News} & \textbf{Average} \\
\midrule
\SMOTE{\tiny (Our reproduction)} & $10.34\times$ & $10.71\times$ & $29.86\times$ & $\infty$ & $99.00\times$ & $22.15\times$ \\
CoDi & $1.00\times$ & $1.08\times$ & $1.04\times$ & $1.04\times$ & $1.03\times$ & $1.04\times$ \\
TabDDPM & $1.05\times$ & $1.09\times$ & $1.72\times$ & $4.03\times$ & $3.83\times$ & $1.87\times$ \\
\TabDiff & $1.00\times$ & $1.05\times$ & $1.01\times$ & $1.02\times$ & $1.04\times$ & $1.02\times$ \\
\TabSYN{\tiny (Our reproduction)} & $1.05\times$ & $1.07\times$ & $1.08\times$ & $1.14\times$ & $1.03\times$ & $1.07\times$ \\
\hline
\CopulaDiff & $1.01\times$ & $1.04\times$ & $1.03\times$ & $1.01\times$ & $1.13\times$ & $1.04\times$ \\
\RandomCoresetTabKDE & $1.65\times$ & $1.71\times$ & $1.43\times$ & $1.74\times$ & $1.25\times$ & $1.54\times$ \\
\CoresetTabKDE & $1.11\times$ & $1.18\times$ & $1.22\times$ & $1.05\times$ & $1.08\times$ & $1.13\times$ \\
\TabKDE & $1.65\times$ & $1.74\times$ & $1.43\times$ & $1.19\times$ & $1.20\times$ & $1.41\times$ \\
\bottomrule
\end{tabular}
\label{tab:dcr_odds}
\end{table}

DCR-based nearest neighbor odds does not account for heldout data nearly as close as the training data to a synthetic point.  In differential privacy, the accepted comparison is the ratio of likelihoods, not the count of which one is maximum.  But there is not a common likelihood across models.  
While there are provably differentially private mechanisms to generate tabular data, we find they have significantly higher error (see Table \ref{tab:dp-adult} in Appendix \ref{app:DP-GC}), even for large privacy tolerances.  

\begin{figure}[htbp]
\small
\phantom{12345678} \SMOTE \hfill \TabSYN \hfill  \CoresetTabKDE \hfill \TabKDE \phantom{123445} 

\includegraphics[width=0.245\linewidth]{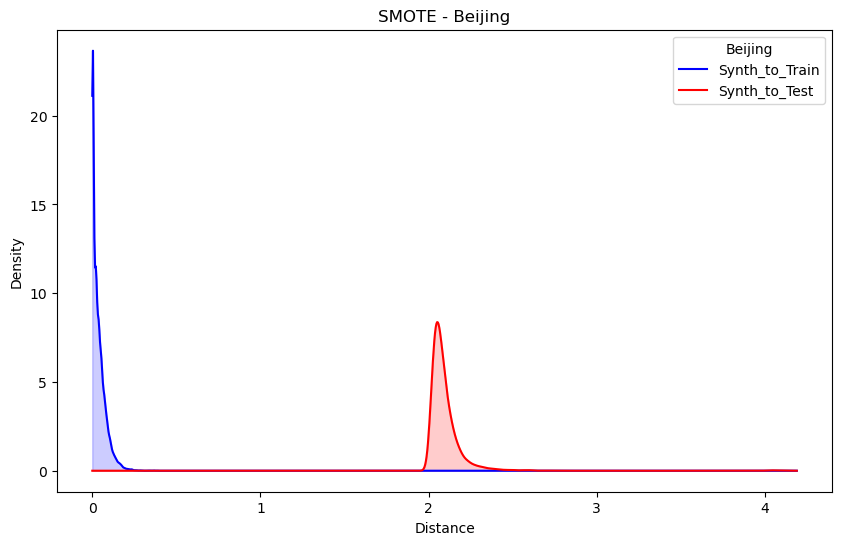}   \includegraphics[width=0.245\linewidth]{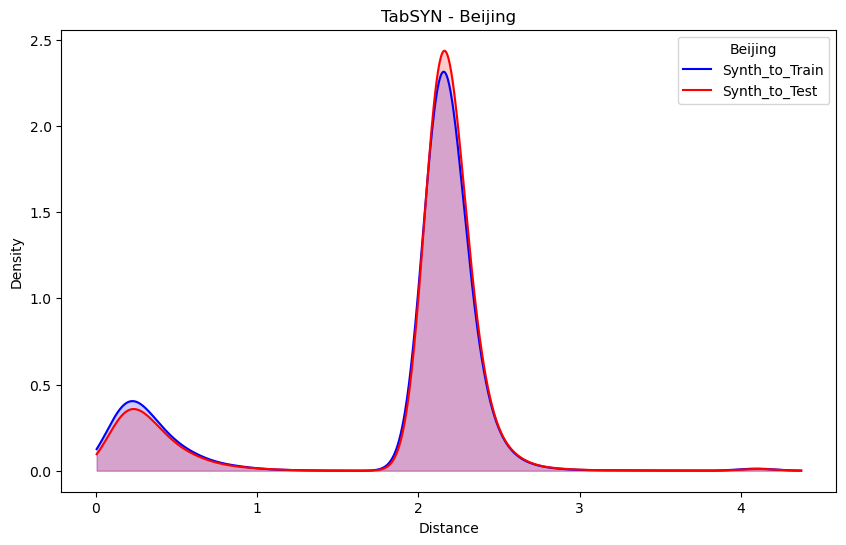} \includegraphics[width=0.245\linewidth]{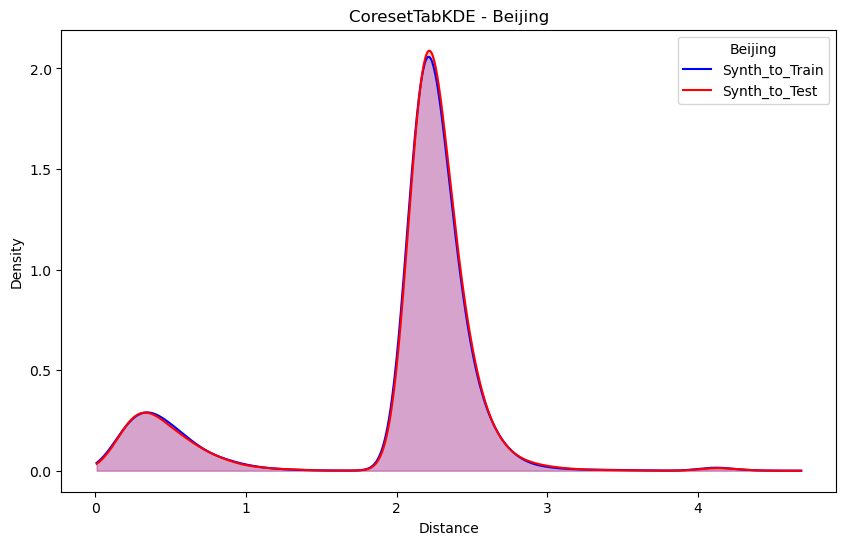} 
\includegraphics[width=0.245\linewidth]{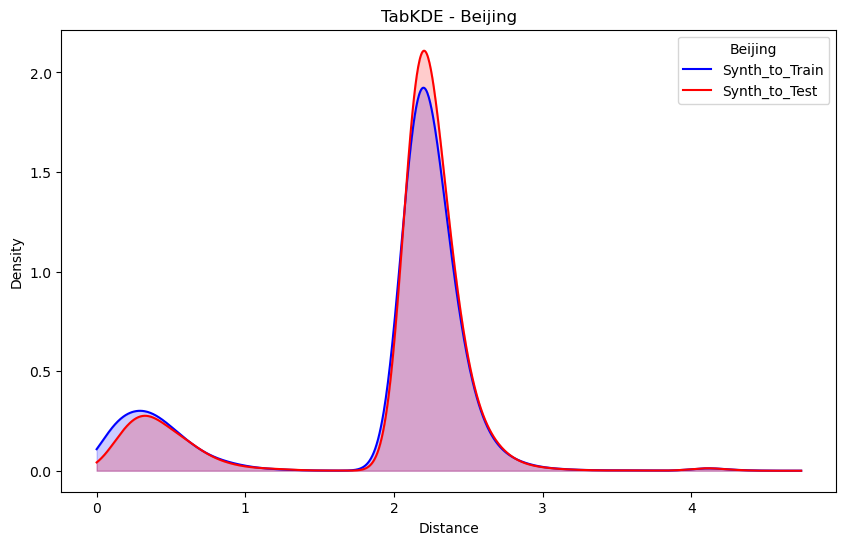} 
\\        
\includegraphics[width=0.245\linewidth]{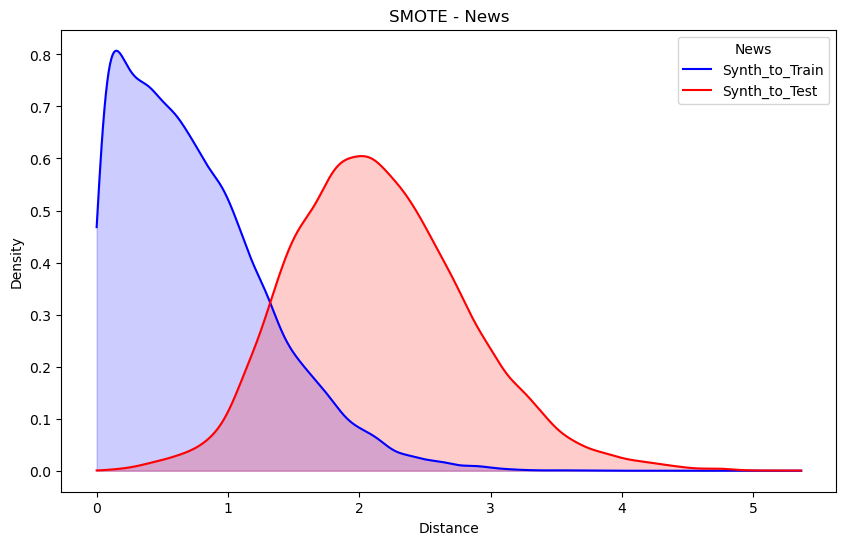} 
\includegraphics[width=0.245\linewidth]{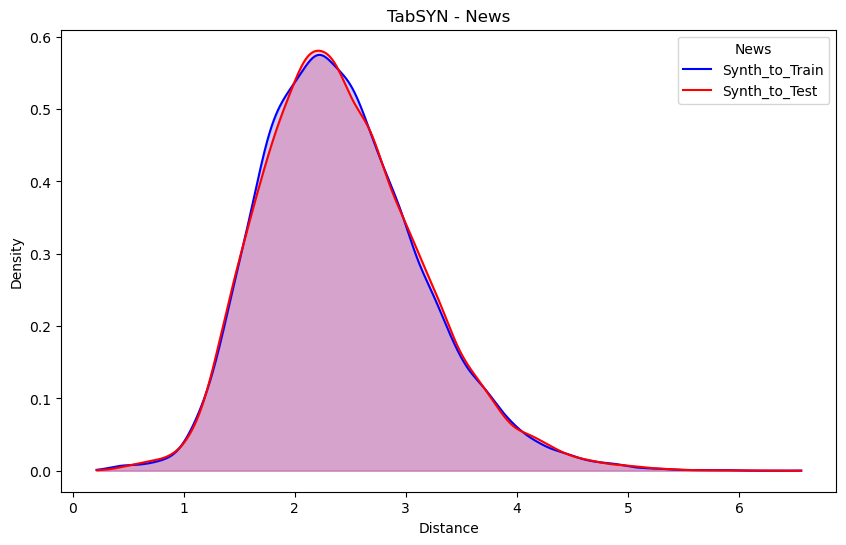} 
\includegraphics[width=0.245\linewidth]{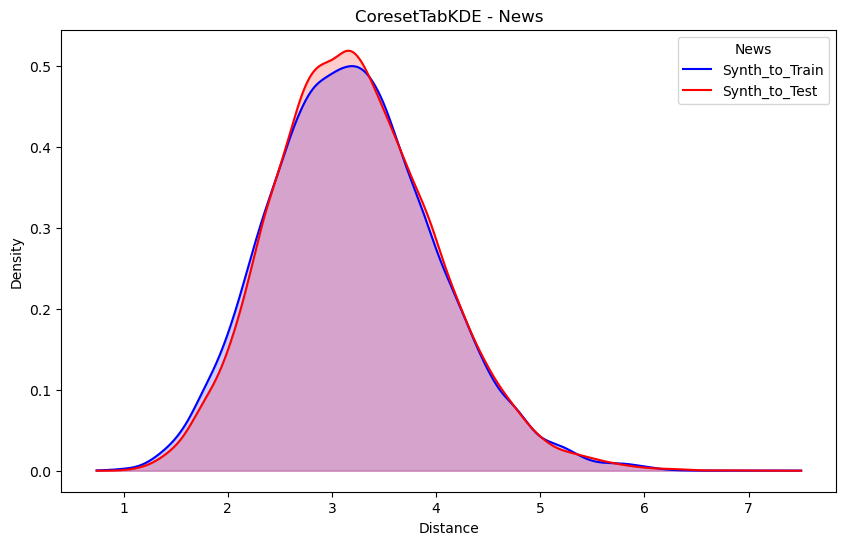} 
\includegraphics[width=0.245\linewidth]{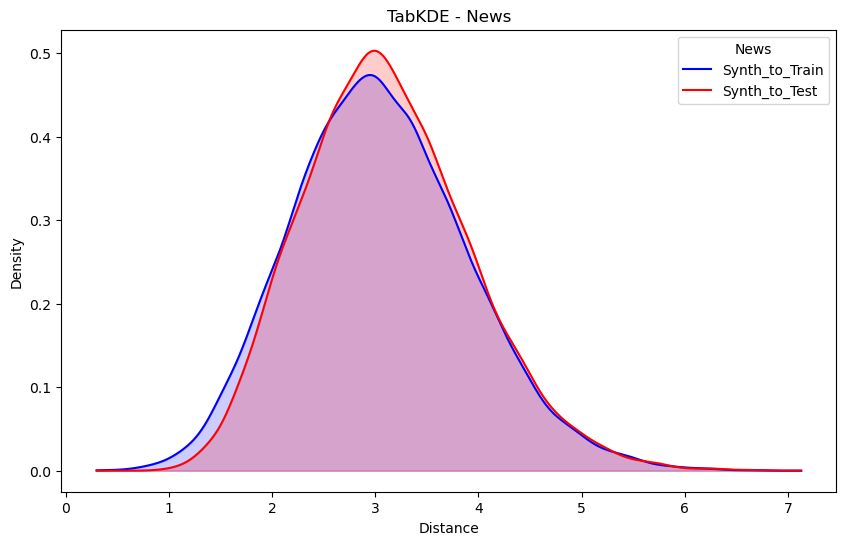} 
\caption{Privacy comparison based on DCR distributions for synthetic to training data (blue) and synthetic to held-out data (red).  First row Beijing and second row News.}
    \label{fig:DCR_comparison}
\end{figure}

So another, albeit less quantitative, evaluation considers the DCR distribution of synthetic data measured to training versus held-out data.  
Figure \ref{fig:DCR_comparison} shows distribution of distances to training (blue) versus to held-out (red) for representative data on Beijing and News.  
If a generator concentrates probability mass near training examples (memorization), the blue distribution
exhibits a systematic left-shift toward zero relative to red, indicating that synthetic samples preferentially
occupy ``training neighborhoods'' across distance scales.
Conversely, strong overlap between the two distributions indicates that synthetic samples lie in neighborhoods
that are similarly populated by unseen held-out records, suggesting little geometry-driven membership advantage
over a range of neighborhood radii.  
We observe that for \TabKDE, \TabSYN, and \CoresetTabKDE these distributions are multi-modal, but still match almost perfectly.  On the other hand \SMOTE has a very different distribution, and the synthetic to train (blue) is always much smaller (typically very close to 0), indicating it may often reproduce the training data.  
Results for all datasets are provided in Figure~\ref{app:fig:DCR_comparison} in Appendix \ref{app:privacy}.

\section{Conclusion}
\TabKDE builds a new paradigm for tabular data generation from simple components, primarily copula transform and KDEs.  It is far more scalable than popular diffusion methods, especially for settings with many categories, while retaining high accuracy and leakage protection.  Moreover, its simple components and retention of one-dimension-per-feature can improve interpretability.

\label{sec:coreset-eval}
We also introduce \emph{coresets for tabular data generation}.  These methods \CoresetTabKDE and \RandomCoresetTabKDE work nearly as well as \TabKDE in terms of accuracy, and use far less space.  Surprisingly, \RandomCoresetTabKDE often has better accuracy than \CoresetTabKDE and is faster since it does not require the optimization step.  However, notably, \CoresetTabKDE has a much improved NNO for privacy (of 1.13$\times$), so provides advanced protection against data leakage within the \TabKDE framework.  
See Appendix \ref{app:coreset} for a longer discussion.

\begin{ack}
Work supported by funding from NSF 2115677 and 2421782, Simons Foundation
MPS-AI-00010515 and Visa Research.
\end{ack}

\bibliographystyle{plainnat}
\bibliography{ref}

\clearpage
\appendix

\section{TabKDE Algorithm}\label{app:sec:tabkde}
 
The \TabKDE algorithm is a simple, scalable, and privacy-aware method for generating high-fidelity synthetic tabular data. It implements the general framework (see Section~\ref{sec:intro}) for tabular generation; the key innovation is in the mapping to latent space and generative modeling sets.  These steps are simple and efficient while satisfying our desiderata.  We overview them here:

\textbf{TabKDE Latent Space Mapping:}  This is accomplished in two parts.  The first step is a copula-based transformation of the tabular dataset $\mathcal{T}$ into a latent representation $Z$ that lies within the continuous space $[0,1]^d$, allowing precise control over the domain of the marginal distributions.  Second is estimating the covariance $\Sigma$ of data in this space, implicitly defining a Mahalanobis distance which captures the similarity within this latent domain.  

\textbf{TabKDE Generative Modeling:}  Here we use a KDE model to allow for a non-parametric complex distributional model.  We simply sample by choosing a training data point in the latent space $z \sim \mathsf{Unif}(Z)$, and then chose a random direction $u$ scaled by the covariance $\Sigma$.  Then we generate a new point 
\[
z' \leftarrow z + u r
\]
where $r$ is a scalar amount controlling the amount of deviation from the training data.  In particular, we select $r$ at random from a learned distribution from the training data estimating the distance to the closest point.  
One more idea is needed, we want to disallow points outside $[0,1]^d$ to respect the original column marginals.  To handle this, we keep coordinates in $[0,1]^d$, and regenerate the others iteratively.    

We next describe all aspects in more detail.

\subsection{Encoding of Tabular Features: $\TT \to E$}
\label{app::subsec:initialencoding}

Let $\TT =\{X_1,\ldots, X_n\}$ be a real data set.
Before transforming tabular data into a latent space, it is essential to convert all feature types into a unified numerical format $\TT \mapsto E \in \R^{n \times d}$, suitable for further processing. Note that the new representation $E$ has $n$ rows (one for each row of the table $\TT$), and more importantly $d$ columns (one for each column of the table $\TT$). This means, do not use one-hot-encoding, and this will be essential for our representation and sampling from the latent space to ensure marginal properties of each data column.  

In this section, we describe how we handle numerical, ordinal, and categorical features through encoding strategies designed to preserve basic structural relationships and facilitate meaningful downstream transformations.

\paragraph{Numerical and Ordinal Mapping.} 
All numerical columns are $z$-score normalized (parameters fit on the training data, saved, and used to de-normalize synthetic outputs back to the original units), so values are reported on the original scale.  
Ordinal features are simply converted to consecutive integers $1,2,\ldots,$ that preserve their inherent order. This ensures that the ordinal relationships between categories are preserved while converting them into numerical representations.


\paragraph{Categorical Mapping.} 

For categorical features, we use a data-driven approach to map each unique category to a continuous space based on the statistical properties of the numerical features in the data. 

Recall that $\Cat = \{d_2+1,\ldots, d\}$ denotes the set of indices corresponding to categorical features. Additionally, assume that for each $j \in \Cat$, the number of unique categories in the $j$-th feature is given by $|C_j| = k_j$. Define $X_\Num \in \mathbb{R}^{n \times d_1}$ as the matrix obtained by selecting only the numerical features from the dataset, and let $u$ represent its first principal component scores (PC1 scores) obtained using Principal Component Analysis (PCA) ($u = X_\Num v$ where $v$ is top principal vector of $X_\Num$).
Let $ j \in \Cat $ be an arbitrary categorical feature with unique category set \( C_j = \{c_1, \ldots, c_{k_j}\} \). For each category \( c \in C_j \), we identify the corresponding row indices in \( D \), defined as  
\[
I_{j,c} = \{ i \in [n] \mid (X_i)_j = c \}.
\]  
We then assign each category \( c \) the value 
\[
v_{j,c} = \frac{1}{|I_{j,c}|} \sum_{i \in I_{j,c}} u_i,
\]  
This value represents the average of the first principal component scores \( u_i \) for all instances where the \( j \)-th categorical feature takes the value \( c \). This process is summarized in the following algorithm.

\begin{algorithm}[H]
\caption{$\textsc{PrincipalGuidedEncoding}(\mathcal{T}, \Cat, \Num)$}
\label{app:alg:categorical_embedding}
\begin{algorithmic}[1]
\State Compute $u$ as the first principal component scores of $X_\Num$; which contains only numerical features in $\mathcal{T}$.
\State \textbf{for} each category $c \in C_j$, in each categorical feature \( j \in \Cat \):
        \[
        v_{j,c} \leftarrow \frac{1}{|I_{j,c}|} \sum_{i \in I_{j,c}} u_i, \quad \text{where } I_c = \{ i \in [n] \mid (X_i)_j = c \}
        \]
\State \textbf{for} each data $(X_i)_j = c \in C_j$, for each categorical feature \( j \in \Cat \):   \;\;\;\;    
        $  (E_i)_j \leftarrow v_{j,c} $

\end{algorithmic}
\end{algorithm}

PGE can be interpreted as using the leading principal component scores of the numerical block as a one-dimensional ``ruler'': each category is embedded as a scalar given by the mean PCA score of the rows in which it appears. When categorical semantics are mediated through numerical covariates, this construction tends to place categories with similar co-occurring numerical profiles nearby along that dominant direction, yielding a continuous embedding that preserves meaningful neighborhood relations. Because the code for each category is learned from the numerical block, PGE helps retain \emph{cat--num} dependence and can also reflect \emph{cat--cat} structure that is induced via shared numerical correlates. In contrast, column-wise encodings such as uniform or frequency mappings depend only on the marginal distribution of the categorical column, and thus cannot incorporate cross-type associations by construction.
See Section~\ref{sec:accuracy} for some related ablation study.

\subsection{Map to Numerical Latent Space: $E \to Z$}
\label{app:EtoZ}
Notably, after the initial encoding step (Subsection~\ref{app::subsec:initialencoding}), all features in the dataset are converted into numerical values, resulting in a representation that lies in a subset of \( \mathbb{R}^d \). Our goal is to further transform this representation into a continuous latent space within the unit hypercube \( [0,1]^d \), where the dependencies between features are preserved, and the data is appropriately normalized to ensure that all numerical features contribute equally, making it well-suited for downstream tasks such as sampling and density estimation.

The algorithm outlined below--\textsc{MapToLatentSpace}--provides a high-level overview of this transformation process. It combines ordinal encoding, a structure-aware encoding of categorical features, and a copula-based normalization of all features. While this procedure is presented here in full, each of its core components will be introduced and discussed in detail in the subsequent sections.

\begin{algorithm}[H]
\caption{$\textsc{MapToLatentSpace}(\mathcal{T}, \Num, \Cat, \Ord)$}
\label{app:alg:mapping_to_latent}
\begin{algorithmic}[1]
\State Mapped each ordinal feature to integers reflecting its natural order. 
\State Encoded categorical features by $\textsc{PrincipalGuidedEncoding}(\mathcal{T}, \Cat, \Num)$
\State Concatenate numerical and the transformed ordinal and encoded categorical features to obtain $E$
\State Convert the encoded data $E$ into \( Z  =\textsc{CopulaMapping}(E)\in [0,1]^{n\times d}\).  
\State  \Return $Z$
\end{algorithmic}
\end{algorithm}


Before exploring the details of this mapping, we first introduce foundational concepts from the copula method—a well-established statistical technique that separates marginal distributions from the dependency structure in multivariate data. 

\paragraph{\bf Introduction to Copula Transformation.}
In many real-world datasets, variables exhibit complex dependencies, making it challenging to model their joint distribution directly. Copula method provides a powerful statistical tool to decouple the dependency structure from the marginal distributions, allowing for more flexible data transformations. The copula method invertibly transforms a dataset \( E \in \R^{n \times d} \), consisting of \( d \)-dimensional features, into a new representation \( Z \in [0,1]^{n \times d} \) and as a result, each individual record in \( Z \) lies in \( [0,1]^d \), and the marginal distributions of \( Z \) are uniform over the interval \([0,1]\). We next examine the underlying mechanism by which this transformation is achieved. 

\begin{enumerate}
    \item \textbf{Copula Forward Transformation (Mapping \( E \) to \( Z \))}:  
    For each dimension \( j \) (where \( j = 1, \dots, d \)), we compute the empirical cumulative distribution function (ECDF) of the $j$-th feature: 
    \begin{align*}
        \hat{F}_j(x) & = \Pr(\text{value of $j$-th coordinate} \leq x)\\
                     & = \frac{1}{n} \sum_{i=1}^{n} \mathbb{I}(x_{ij} \leq x)
    \end{align*}
    where \( \mathbb{I}(x_{ij} \leq x) \) is an indicator function that equals \( 1 \) if \( x_{ij} \leq x\), and \( 0 \) otherwise.
    In summary, \( \hat{F}_j(x) \) represents the proportion of observations in the dataset whose $j$-th coordinates are less than or equal to \( x \). Each coordinate value \( x_{ij} \) is then transformed into a uniform representation:
    \begin{equation}
        z_{ij} = \hat{F}_j(x_{ij})
    \end{equation}
    This ensures that each feature is uniformly mapped into the interval \( [0,1] \), producing a dataset \( Z \) that follows a uniform distribution for each of its marginals while maintaining the dependency structure of \( X \).

    \begin{algorithm}[H]
    \caption{$\textsc{CopulaMapping}(E)$}
    \label{app:alg:copula_representation}
    \begin{algorithmic}[1]
    \State For each feature $j$, compute empirical CDF value $\hat{F}_j(x_{i,j})$ for $D_j = \{x_{1j},\ldots, x_{nj}\}$.
        \State Set $z_i = (z_{i,1}, \ldots, z_{i,d})$, where $z_{i,j} = \hat{F}_j(x_{i,j})$
    \State \Return $Z= \{z_1,\ldots, z_n\}$
    \end{algorithmic}
    \end{algorithm}
Using the Copula transformation, we effectively \emph{standardize} the data set into a unit hypercube, making it more suitable for density estimation, sampling, and synthetic data generation. Furthermore, this method allows for \emph{dependency-preserving transformations}, ensuring that the statistical relationships between variables are retained even when synthetic data is produced.

    \item \textbf{Copula Inverse Transformation (Mapping \( Z \) back to \( E \))}:  It is key that we store the ECDF  $\hat{F}$, because we need to be able to invert it.  Its inverse cumulative distribution function (quantile function) 
    $\hat{F}^{-1}$ is defined as 
\[
\hat{F}^{-1}(q) = \inf \{ x \mid \hat{F}(x) \geq q \} \quad\quad\text{for any value }\quad\quad q\in [0,1].
\]
Given \( z = (z_1, \ldots, z_d) \in [0,1]^d \), each feature \( j \) can be mapped back to its initial numerical representation using the inverse cumulative distribution function $\hat{F}_j^{-1}(\cdot)$.   

\paragraph{Decoding. }  
We can also decode the output $E$ to the structure in the table.  For an ordinal or categorical feature \(j\), we apply probabilistic rounding to the two nearest categories (in the initial numerical embedding), with the probabilities proportional to their distance to \(p_j\). 
Also, for numerical features, the value is reconstructed (up to the appropriate precision) by interpolating between the two closest values (in the original distribution) to \(p_j\), with the interpolation weights determined by their distance from \(p_j\). This step ensures that the generated samples maintain the same marginal distributions as the original dataset.


\begin{algorithm}[H]
\caption{$\textsc{InverseECDF}(z = (p_1,\ldots,p_d), \{\hat{F}_i: i\in [d]\})$}
\label{app:alg:Inverse_Copula}
\begin{algorithmic}[1]
\State {\bf for} each $j\in[d]$:
\State  \hspace{0.5cm} \textbf{if} $\min(\{z_{ij}\}_{i=1}^n) \geq p_j$ or  $\max(\{z_{ij}\}_{i=1}^n) \leq p_j$ \textbf{then}
        \State \hspace{1.0cm} \textbf{Return} $\min(\{x_{ij}\}_{i=1}^n)$ or $\max(\{x_{ij}\}_{i=1}^n)$ respectively
\State \hspace{0.5cm}Let $z_{i_1} < z_{i_2}$ be consecutively ordered points so that $p \in [z_{i_1},  z_{i_2}]$
\State \hspace{0.5cm}\textbf{if} $j \in \Ord$ or $j \in \Cat$, \textbf{then}
\State \hspace{1cm} \textbf{Return} \(x_j = x_{i_1}\) with probability $\frac{|p_j- z_{i_1}|}{|z_{i_1}-z_{i_2}|}$ and otherwise $x_{i_2}$.
\State \hspace{0.5cm}\textbf{elseif} $j \in \Num$, \textbf{then}
\State \hspace{1cm}\ \textbf{Return} 
\begin{equation}
    x = x_{i_2} + \frac{|p_j - z_{i_2}|}{|z_{i_1} - z_{i_2}|} (x_{i_1} - x_{i_2})
\end{equation}
\end{algorithmic}
\end{algorithm}
\end{enumerate}

\subsection{Sensitivity Analysis of the Encoding–Decoding Pipeline}
\PGE represents each category $c$ by scalar $v_{j,c}$: the mean of first principal component scores $u$ (PC1 scores) restricting to value $c$ in column $j$.  It is computed from the $z$-score normalized numerical block, making it insensitive to affine rescaling of the raw numerical columns. \TabKDE then applies a per-coordinate empirical-copula transform (ECDF), which depends only on ranks of these scalars and therefore further reduces sensitivity to monotone re-parameterizations of each coordinate. Note that statistically this is almost impossible to different categories $c\neq c'$ receive same value $v_{j,c}=v_{j,c'}$. During decoding, categorical and ordinal coordinates are recovered via probabilistic rounding to the two nearest embedded codes using the inverse ECDF (which is computed on ranks), so categories become indistinguishable only in the presence of exact ties in their learned codes. Finally, as shown in Section~\ref{sec:accuracy}, \TabKDE is robust to replacing \PGE with alternative single-scalar encodings like frequency or uniform.

\subsection{Learning Distance to Closest Record (DCR)}
\label{app:subsec:dcrdefin}

A central use of synthetic data is as a proxy for private personal data.  So it is paramount to ensure that the synthetic data process is not leaking too much information about the original data. A common measurement of this is  \textbf{Distance to Closest Record (DCR)}~\citep{MateoSanz2004, Steier2025} which evaluates how similar a synthetic record $x_s$ is to a real one from a set $D$. It is formally defined for an appropriate distance metric $\texttt{d}$ as:

\begin{equation}
\text{DCR}(x_s, D) = \min_{x_r \in D} \texttt{d}(x_s, x_r)
\end{equation}

A DCR of  $0$ indicates an identical match, posing a significant privacy risk. Comparing DCR values between synthetic data and both training $D_T$ and holdout $D_H$ datasets helps assess privacy. If synthetic records are much closer to the training data, it suggests the model may be memorizing real data. Ideally, for synthetic data $S$,  DCR distributions $\{\text{DCR}(x_s,D_T)\}_{x_s \in S}$ and $\{\text{DCR}(x_s,D_H)\}_{x_s \in S}$ \textbf{should heavily overlap}, showing that synthetic data reflects general patterns rather than replicating specific records.   

As part of our generative process, we learn this distribution in the copula latent embedding $Z$ using {\it Euclidean distance}. 
Then we can generate synthetic data to mirror this scale of variation.  
We repeatedly randomly split the training data $Z$, and compute the DCR distribution between the two splits (see $\textsc{EmpiricalDCR}$; Alg. \ref{app:alg:DCR_estimation}).  Then it fits a simple mixture of Gaussians model to this distribution; 
using Bayesian Information Criterion (BIC)\footnote{\url{https://scikit-learn.org/stable/modules/generated/sklearn.mixture.GaussianMixture.html##sklearn.mixture.GaussianMixture.bic}}, we select the best model for $k = 1,\ldots, 10$ as the number of components.

\begin{algorithm}[H]
\caption{$\textsc{EmpiricalDCR}(Z)$:
Estimating the Empirical DCR Distribution}
\label{app:alg:DCR_estimation}
\begin{algorithmic}[1]
\State Initialize \( L = [] \)
\State \textbf{for} \( i = 1, \dots, T \) \textbf{do}
    \State \hspace{0.5cm} Partition \( Z \) into two random equal-sized subsets \( Z_1 \) and \( Z_2 \).
    \State \hspace{0.5cm} \textbf{for} each \( z_2 \in Z_2 \) \textbf{do}
        \State \hspace{1cm}  Compute the minimum distance between \( z_2 \) and the records in \( Z_1 \).
        \State \hspace{1cm} Add this distance to \( L \).
\State Fit a mixture of \( k \) Gaussian components to \( L \).
\end{algorithmic}
\end{algorithm}

\subsection{Tabular Kernel Density Estimation: $Z \to$ Sample}

KDE (Kernel Density Estimation) is a non-parametric method used to estimate the probability density function (PDF) of a continuous random variable by smoothing finite data points with a kernel function (typically Gaussian). Its accuracy depends on bandwidth selection and data availability~\citep{Silverman86,Scott2015}.  It can also be used to generate synthetic data by fitting a KDE model to the existing dataset and drawing samples from it. We now formally define KDE.
 
Assuming that \( X = \{ x_1, x_2, \dots, x_n \} \) is a dataset in \( \mathbb{R}^d \), the Kernel Density Estimation (KDE) is given by:

\[
\hat{f}(x) \propto \frac{1}{n} \sum_{i=1}^{n} K \left( \frac{x - x_i}{h} \right)
\]
where:
\begin{itemize}
    \item[-] \( \hat{f}(x) \) is the estimated likelihood at point \( x \),
    \item[-] \( K(\cdot) \) is a centrally-symmetric kernel function (e.g., Gaussian kernel),
    \item[-] \( h > 0 \) is the bandwidth parameter controlling smoothness.
\end{itemize}

\paragraph{\bf Sampling from KDE.} 
To sample from a KDE$_X$, we simply sample a point $x \in X$, and then sample a point nearby proportional to the kernel likelihood.  For example, with a isotropic Gaussian kernel, using this approach, a synthetic data point is generated as \( x \sim \mathcal{N}(x_i, \frac{h}{2}I);\; x_i \sim X\).  
However, we do not use a Gaussian kernel, as this is not adaptive to the DCR distribution -- which may be multi-modal.  So we use our $\textsc{EmpiricalDCR}(Z)$ estimate to define our kernel.  
Still, this approach may raise a couple of concerns:

\begin{itemize}
    \item At first glance, one may worry that this approach does not guarantee preservation of DCR, since it might use our kernel $K(x_i, \cdot)$ to generate a point nearby $x_i$ that lands too close to another $x' \in X$.  However, this is not observed to be an issue, as generation in high-dimensional space makes it highly unlikely to produce points close to other points in the training data (see Subsection~\ref{subsec:dcrdefin}).
    
    \item Second, the sampled point \( x \) may fall outside the convex hull of the dataset \( X \), potentially resulting in unrealistic data generation.  We address this by using the sample covariance $\Sigma$ to guide the perturbation direction.  And moreover, some form of extrapolation in this sense is probably necessary and unavoidable~\citep{balestriero2021learning}, and we believe desirable.   Yet, violating the marginals associated with individual table columns, we find, can distort distributions (see Subsection~\ref{app:subsec:whyTabKDE}). This issue is addressed by generalizing \textsc{SimpleKDE} to a more advanced method, \TabKDE, which controls for this.  

\end{itemize}
To resolve the first issue, we estimate the DCR distribution using Algorithm~\ref{app:alg:DCR_estimation} (\textsc{EmpiricalDCR}) and leverage it to strategically perturb the sampled point \( x_i \). 
We summarize this approach by the following algorithms: Algorithm \ref{app:alg:SimpleKDE} \textsc{SimpleKDE}($\TT$), which iteratively calls Algorithm \ref{app:alg:SampleKDE} \textsc{SampleKDE}($Z,f,\Sigma$) using the copula-transformed data $Z$, its estimated DCR distribution $f$, and its estimated covariance $\Sigma$.

\begin{algorithm}[H]
\caption{\textsc{SimpleKDE}($\TT$)}
\label{app:alg:SimpleKDE}
\begin{algorithmic}[1]
\State Transform table $\TT$ into $Z \in [0,1]^{n\times d}$ as $Z \leftarrow \textsc{CopulaTransform}(\TT, \Num, \Cat, \Ord)$ 
\State $\Sigma \gets \text{Covariance}(Z)$
\State Estimate empirical DCR distribution \( f = \textsc{EmpiricalDCR}(Z) \)
\State \textbf{for} $i= 1,\ldots, m$: 
\State\hspace{0.5cm}\( z'_i = \textsc{SampleKDE}(Z, f, \Sigma) \) 
\State \hspace{0.5cm} \(y_{j} \gets \textsc{InverseECDF}(z'_{j}, \text{\(j\)-th feature type}, F_j)\)

\State \Return \( Y=\{y_1,\ldots, y_m\} \)
\end{algorithmic}
\end{algorithm}

\begin{algorithm}[H]
\caption{$\textsc{SampleKDE}(Z,f,\Sigma)
$}
\label{app:alg:SampleKDE}
\begin{algorithmic}[1]
\State Uniformly sample \( z_i \in Z \)
\State Sample radius \( r > 0 \) from \( f \)
\State Sample \( v \sim \mathcal{N}(0, \Sigma) \), set \( u = \frac{v}{\|v\|} \)
\State \Return  \( z' \gets z_i + r \cdot u \)
\end{algorithmic}
\end{algorithm}

As discussed in Subsection~\ref{app:subsec:whyTabKDE}, the \textsc{SimpleKDE} algorithm does not explicitly control the support of the marginals and, in particular, does not fully address the second challenge outlined earlier. The copula-transformed representation, however, embeds the data within the unit hypercube, which allows us to control how far a perturbed sampled point \( x \) can deviate without violating marginal support. We now introduce a more refined rejection-sampling heuristic (Alg. \ref{app:alg:TabKDE}: \TabKDE) that effectively enforces these boundary constraints.

\begin{algorithm}[H]
\caption{\textsc{TabKDE}(X)}
\label{app:alg:TabKDE}
\begin{algorithmic}[1]
\State Transform table $\TT$ into $Z \in [0,1]^{n\times d}$ as $Z \leftarrow \textsc{CopulaTransform}(\TT, \Num, \Cat, \Ord)$
\State $\Sigma \gets \text{Covariance}(Z)$
\State Estimate empirical DCR distribution \( f = \textsc{EmpiricalDCR}(Z) \)
\State \textbf{for} $i= 1,\ldots, m$: 
\State\hspace{0.5cm}\( z'_i = \textsc{SampleKDE-iterative}(Z, f, \Sigma) \) 
\State\hspace{0.5cm} \( y_i = \textsc{InverseCopula}(z'_i) \)
\State \Return \( Y=\{y_1,\ldots, y_m\} \)
\end{algorithmic}
\end{algorithm}

\TabKDE differs from \textsc{SimpleKDE} only in the sampling step at line 5, where it uses the boundary-aware \textsc{SampleKDE-iterative} instead of the simpler \textsc{SampleKDE}. This modified sampler actively checks for violations of the unit hypercube boundaries and iteratively adjusts any out-of-bound coordinates. If a valid point cannot be obtained after a fixed number of attempts ($10d$), the sample is discarded, and the process restarts. This mechanism guarantees that all accepted samples lie within the latent space \([0,1]^d\).

\begin{algorithm}[H]
\caption{\textsc{SampleKDE-iterative}(Z)}
\label{app:alg:SampleKDE-iterative}
\begin{algorithmic}[1]
\State Uniformly sample \( z_i \in Z \)
\State Sample radius \( r > 0 \) from \( f \)
\State Sample \( v \sim \mathcal{N}(0, \Sigma) \), set \( u = \frac{v}{\|v\|} \)
\State \( z' \gets z_i + r \cdot u \)
\State {\bf While} \(\{j : z'_j \notin [0,1] \} \neq \varnothing \): 
    \State \hspace{.5cm} \( J \gets \{j : z'_j \notin [0,1] \} \)
    \State \hspace{.5cm} Sample \( v' \sim \mathcal{N}(0, \Sigma) \), set \( w = \frac{v'}{\|v'\|} \)
    \State \hspace{.5cm} \( s \gets \frac{\| (u_k)_{k \in J} \|}{\| (w_k)_{k \in J} \|} \)
    \State \hspace{.5cm} \( u_j \gets s \cdot w_j \) for each \( j \in J \) 
    \State \hspace{.5cm} \( z' \gets z_i + r \cdot u \)
\State \Return \( z' \)
\end{algorithmic}
\end{algorithm}

\section{Experimental Setup and Data}\label{app:sec:experimental}

\subsection{Datasets}
\label{app:datasets}
Our experiments are conducted on the six tabular datasets from UCI Machine Learning Repository\footnote{\url{https://archive.ics.uci.edu/datasets}}  (Adult, Default, Shoppers, Magic, Beijing, News )
used in \TabSYN~\citep{tabsyn2024}, along with the IBM  dataset\footnote{\href{https://www.kaggle.com/code/yichenzhang1226/ibm-credit-card-fraud-detection-eda-random-forest/input?select=credit_card_transactions-ibm_v2.csv}{https://www.kaggle.com/code/yichenzhang1226/ibm-credit-card-fraud-detection-eda-random-forest}}
 which is significantly larger.  In all they covering a wide range of domains for tabular data.  
 These datasets include a mix of numerical and categorical features and vary in the number of points, feature types, and task types (classification or regression), making them well-suited for evaluating the generalizability of synthetic data generation methods. A few of the categorical features can be interpreted as ordinal; but outside the IBM dataset, we simply treat them as categorical.  
 We summarize their traits in Table \ref{app:tab:dataset_summary}.

\begin{table}[ht]
\caption{Dataset statistics. {\bf Num} denotes the number of numerical columns, {\bf Cat} the number of categorical columns, {\bf Ord} the number of ordinal features, and {\bf Sum Cat} the total number of unique categories across all categorical and ordinal columns. Ordinal features can be treated as categorical features by disregarding their inherent order; note ($*$) that we do this for  the Adult and Default datasets.  For the IBM dataset, we randomly select two 200k subsamples to serve as the training and testing sets; we ensured that the test set contains no categorical values unseen in the training set. In IBM data, we treat ``\emph{Year}'', ``\emph{Month}'', ``\emph{Day}'', ``\emph{Time}'', ``\emph{Zip}'' features as ordinal. ($\dagger$) In \TabKDE, \simpKDE, \CopulaDiff and \CoresetTabKDE, for Beijing dataset, we treat the features ``Is'', ``Ir'', and ``Iws'' as categorical, and for the Shoppers dataset, we apply the same treatment to the features ``SpecialDay'', ``ProductRelated'', and ``Informational''.
For all of these features, the ratio of unique values in the training set to the total number of data points is very low, with the majority of occurrences concentrated on a single value. }
\label{app:tab:dataset_summary}
\centering
\begin{tabular}{|l|c|c|c|c|c|c|c|l|}
\hline
\textbf{Dataset} & \textbf{Total} & \textbf{Train} & \textbf{Test} & \textbf{Num} & \textbf{Cat} & \textbf{Ord} & \textbf{Sum Cat} & \textbf{Task} \\
\hline
Adult    & 48,842         & 32,561         & 16,281        & 6           & 7           & 2$^*$    & 120             & Classification \\
Default  & 30,000         & 27,000         & 3,000         & 14          & 9           & 1$^*$    & 79              & Classification \\
Shoppers$^\dagger$ & 12,330         & 11,097         & 1,233         & 10          & 8           & 0    & 67              & Classification \\
Magic    & 19,019         & 17,117         & 1,902         & 10          & 1           & 0    & 2               & Classification \\
Beijing$^\dagger$  & 41,757         & 37,581         & 4,176         & 7           & 5           & 0    & 76              & Regression     \\
News     & 39,644         & 35,679         & 3,965         & 46          & 2           & 0    & 13              & Regression     \\
IBM      & 341,675        & 176,221        & 165,454       & 1           & 8          &  5     & 37,721          & Classification \\
\hline
\end{tabular}
\end{table}

{\bf Split of the Data.} We consider two ways to split data into test and train set.  The numbers in Table \ref{app:tab:dataset_summary} reflects the splits done by \TabSYN, which we maintain for direct comparison.  This split was not even in size, partially to ensure there were no categories in the Test split which were not present in the Train split.  
When we do not directly compare to results in the \citep{tabsyn2024}, we use a different random and even split.  

Again, following the \TabSYN paper \citep{tabsyn2024}, the target column is treated as either numerical or categorical based on the task type: it is considered categorical for classification tasks and numerical otherwise. For the Machine Learning Efficiency experiments, each dataset is divided into training, validation, and testing sets. For the Adult dataset, we use its official testing set, while the original training set is further split into training and validation sets with an 8:1 ratio. All other datasets are split into training, validation, and testing sets using an 8:1:1 ratio with a fixed random seed.

To get the IBM dataset to run, we needed to leverage the ordinal representation in variables Year, Month, Day, Time, and Zip.  
We also identify that the Merchant State  and Merchant City  are very strongly correlated with Zip (the zip code), and since these categories are often quite rare, we applied another modeling (called \emph{reduced modeling}) to improve the generation.  We put Zip in the generative model, but not Merchant City or Merchant State.  Then once we generate a Zip, we predict the the Merchant City and Merchant State. We use the same process with the correlated MCC and Merchant Name; we put MCC in the generative model, and use the outcome to predict Merchant Name. For transparency, we however report results both with and without this reduced modeling trick (see Table~\ref{app:tab:IBM-accuracy} and Figure~\ref{app:fig:IBM-correlation}).

\subsection{Baselines}
\label{app:baselines}
We compare our proposed \TabKDE method with several popular baselines, including 
SMOTE~\citep{smote2002},  
CTGAN\citep{Xu2019},
TVAE~\citep{Xu2019}, 
GReaT~\citep{borisov2023language}, 
GOGGLE~\citep{liu2023goggle}, 
CoDi~\citep{codi2023}, 
STaSy~\citep{stasy2023}, 
TabDDPM~\citep{TabDDPM2023}, 
\TabSYN~\citep{tabsyn2024}, and 
\TabDiff~\citep{tabdiff2025}\footnote{We have not yet been able to reproduce all results for \TabDiff, but they report~\citep{tabdiff2025} very similar performance to \TabSYN in terms of accuracy, efficiency, and privacy -- although with small, but noticeable improvement on categorical data.  This follows since they build directly on most of the framework of \TabSYN except for using a discrete diffusion on the categorical parts.  For the comparison to \TabKDE and its relatives, \TabSYN's performance should serve as a good empirical representative.}.  
Through this comparison, we demonstrate that \TabKDE provides a simpler, faster, and more scalable alternative for generating realistic synthetic tabular data, without significantly compromising on quality or privacy.

By abstracting to the general framework (outlined in Section 
\ref{sec:intro}
) we are able to also consider several hybrid models that mix elements of \TabKDE with the encoding choices, the VAE method, or the diffusion-driven generation made popular through \TabSYN and others.  
We specifically consider 
\begin{itemize}
    \item {\bf \CopulaDiff:} We first use our \textsc{CopulaMapping} (Alg.~\ref{app:alg:copula_representation}) to embed data into a latent space, train a Diffusion model there, and then map the generated samples back to the original space.
    \item {\bf \VAETabKDE:} This model trains a VAE to embed data into numerical space (as in \TabSYN), then applies the Copula and KDE methods to generate samples, which are then mapped back to the original tabular format. 
    \item {\bf \VAESimpleKDE:}  \VAESimpleKDE differs from \VAETabKDE only in the sampling step, analogous to how \TabKDE differs from \simpKDE.
    \item {\bf \PGETabsyn:}  In the method we replace one-hot encoding with the encoding outlined in Subsection~\ref{app::subsec:initialencoding} (see the first three steps in $\textsc{MapToLatentSpace}$ Alg.~\ref{app:alg:mapping_to_latent}), which tokenizes the tabular data into the space $E$ before applying VAE and Diffusion models.  
\end{itemize}

We also introduce the coreset variants \CoresetTabKDE and \RandomCoresetTabKDE, which replace $Z$ with a coreset (chosen by gradient descent or random sample, respectively).  These are elaborated on in Section \ref{app:coreset}.  

Moreover, we consider two additional variants to ensure the marginal boundary conditions are satisfied while generating a sample.  The default method in \TabKDE is described in Algorithm \ref{alg:SampleKDE-iterative} \textsc{SampleKDE-iterative}, refining the simpler Algorithm \ref{alg:SampleKDE} \textsc{SampleKDE} which does not prevent marginal range violations.  The alternatives allow for simpler rejection sample.  The \textsf{keep-the-seed} variant on boundary violation keeps the empirical point $z_i \in Z$, but regenerates the direction $u$ and distance $r$ to perturb it in $z' \gets z_i + r \cdot u$ step.  The \textsf{change-the-seed} variant also resamples the $z_i \sim Z$ step.  More details are provided in Section \ref{app:correction-stats}.

\subsection{Evaluation} To evaluate our synthetic data generation method, we focus on three main objectives including {\bf 1) Scalability}, {\bf 2) Accuracy}, and {\bf 3) Privacy}. In terms of Efficiency, we measure and compare the training and sampling time required by each model across various datasets. 
For Accuracy, we assess how well the synthetic data captures (1) the ground truth marginals for each column individually, (2)  correlations for pairs of columns, (3)  the entire distribution using machine learning efficiency, 
 and (4) the balance between fidelity and coverage using $\alpha$-Precision and $\beta$-Recall. 
For Privacy, we use the Distance to Closest Records (DCR) metric to assess privacy protection by measuring how similar the synthetic data is to the training/test data sets.

\section{Scalability and Efficiency}
\label{app:scalability}
We measure the scalability and efficiency on both the training time, as well as the sample generation time; this second part is the generation of a sample the same size as the training set.  As shown in Tables \ref{app:tab:runtime_comparison} and \ref{app:tab:time_tabsyn_baselines}, we see in most existing methods the time is almost entirely dominated by the training aspect.  However, \TabKDE the training and sampling are more comparable because the training time is so much lower.  

All our experiments, unless otherwise specified, were conducted using only the CPU of a 2021 Apple MacBook Pro (14-inch), equipped with an Apple M1 Pro chip. This device features an 8-core CPU (comprising 6 performance cores and 2 efficiency cores) and 16~GB of unified memory. Table~\ref{app:tab:runtime_comparison} presents a comparison of the training times between \TabSYN and \textsc{SMOTE} against our proposed \TabKDE, demonstrating that our method is highly computationally efficient and can be effectively executed on a standard consumer-grade laptop.  
\TabSYN and \textsc{SMOTE} both run out of memory on the IBM data set, this is primarily because it has a huge number of categories, and these methods rely on one-hot encoding, which blows up the dimensionality into a $37{,}733$-dimensional space.  The memory inefficient one-hot encoding is standard in many modern models.  SMOTE requires this dimensionality to identify the $k$ nearest neighbors, which becomes highly inefficient in such a high-dimensional space (see Table~\ref{app:tab:dataset_summary} for dataset details).

In contrast, our proposed tokenization method, \textsc{PrincipalGuidedEncoding} (Algorithm~\ref{app:alg:categorical_embedding}), transforms tabular data into a numerical format with a fixed dimensionality equal to the original number of features (14 in this case, compared to 37722 with one-hot encoding), providing a far more efficient representation.



\begin{table}[ht]
\caption{Runtime comparison of Tabsyn, TabKDE, and SMOTE models across individual datasets on laptop. The IBM dataset is excluded from the average row.}
\footnotesize
\centering
\begin{tabular}{l|cccc|c|cc}
\toprule
\textbf{Dataset} & \multicolumn{4}{c|}{\TabSYN} & \SMOTE & \multicolumn{2}{c}{\TabKDE}  \\
\midrule
 & VAE Train & Diff. Train & Total Train & Sample & Train+Sample & Train & Sample \\
\midrule
Adult   & 6h 35m 43s & 2h 6m 31s  & 8h 43m 19s & 1m 5s & 4s & 44s & 20s \\
Default & 6h 32m 3s  & 2h 2m 16s  & 8h 34m 59s & 40s   & 2s & 59s & 17s \\
Shoppers & 3h 57m 42s & 0h 55m 32s & 4h 53m 32s & 18s   & 3s & 17s & 5s  \\
Magic   & 3h 51m 7s  & 1h 21m 27s & 5h 13m 0s  & 26s   &  5s & 19s & 7s   \\
Beijing & 5h 31m 57s & 1h 57m 44s & 7h 30m 35s & 54s  & 2s & 35s & 16s  \\
News & 14h 34m 15s & 2h 8m 59s & 16h 44m 11s & 57s   & 4s& 6m 2s & 54s  \\
\midrule
\textbf{Average} & \textbf{6h 50m 27s} & \textbf{1h 45m 24s} & \textbf{8h 36m 36s} & \textbf{43s} & \textbf{3s} & \textbf{1m 29s} & \textbf{19s} \\
\midrule
IBM   & OOM & OOM  & OOM & OOM & OOM & 10m 21s & 6m 4s  \\
\bottomrule
\end{tabular}
\label{app:tab:runtime_comparison}
\end{table}

The baseline methods in the \TabSYN~\citep{tabsyn2024} paper were evaluated on the \texttt{Adult} dataset using an NVIDIA RTX 4090 GPU with 24~GB of memory, as shown in Table~\ref{app:tab:time_tabsyn_baselines}. 
In contrast, our experiments—including those for \TabSYN and the proposed \textsc{TabKDE}—were conducted entirely on significantly less powerful hardware. Despite this substantial difference in computational resources, \textsc{TabKDE} demonstrates superior efficiency. As shown in Table~\ref{app:tab:runtime_comparison}, it achieves an average training time of only 1 minute and 29 seconds, and a sampling time of just 19 seconds across six benchmark datasets. This is considerably faster than the \TabSYN model, which requires more than 8.5 hours of training on average. These results highlight that \TabKDE not only nearly matches the performance of more complex models (see Section \ref{app:sec:accuracy}) but also does so at a fraction of the computational cost, making it highly suitable for deployment on standard consumer-grade machines without the need for specialized accelerators.

\begin{table}[ht]
\caption{Training and sampling times for baseline methods on the Adult dataset, evaluated using an NVIDIA RTX 4090 GPU with 24 GB of memory (adapted from the \TabSYN~\citep{tabsyn2024} paper).}
\centering
\begin{tabular}{lcc}
\toprule
\textbf{Method} & \textbf{Training Time} & \textbf{Sampling Time} \\
\midrule
CTGAN     & 17 min 10 s   & 0.86 s \\
TVAE      & 5 min 53 s    & 0.51 s \\
GOGGLE    & 1 h 34 min    & 5.34 s \\
GReaT     & 2 h 27 min    & 2 min 19 s \\
STaSy     & 38 min 3 s    & 8.94 s \\
CoDi      & 2 h 56 min    & 4.62 s \\
TabDDPM   & 17 min 11 s   & 28.92 s \\
\TabSYN & 40 min 22 s & 1.78 s \\
\bottomrule
\end{tabular}
\label{app:tab:time_tabsyn_baselines}
\end{table}

As previously noted, running \TabSYN on the IBM dataset is infeasible given our standard computational resources. This limitation arises mainly due to using one-hot encoding—results in a $37{,}733$-dimensional feature space. However, we can alternatively use  encoding scheme introduced in the preprocessing steps of \TabKDE. Accordingly, we apply both \CopulaDiff and \PGETabsyn to the IBM dataset. Training \CopulaDiff requires 15 hours and 7 minutes, while \PGETabsyn demands over 40 hours in total—25 hours and 56 minutes for the VAE and 14 hours and 39 minutes for the diffusion stage.  As shown in Table \ref{app:tab:runtime_comparison}, \TabKDE only takes about 10 minutes of training time.  
This comparison further highlights \TabKDE’s advantage in scalability over more resource-intensive methods. 
 
\begin{figure}[htbp]
    \centering
    \begin{subfigure}[b]{0.45\linewidth}
        \includegraphics[width=\linewidth]{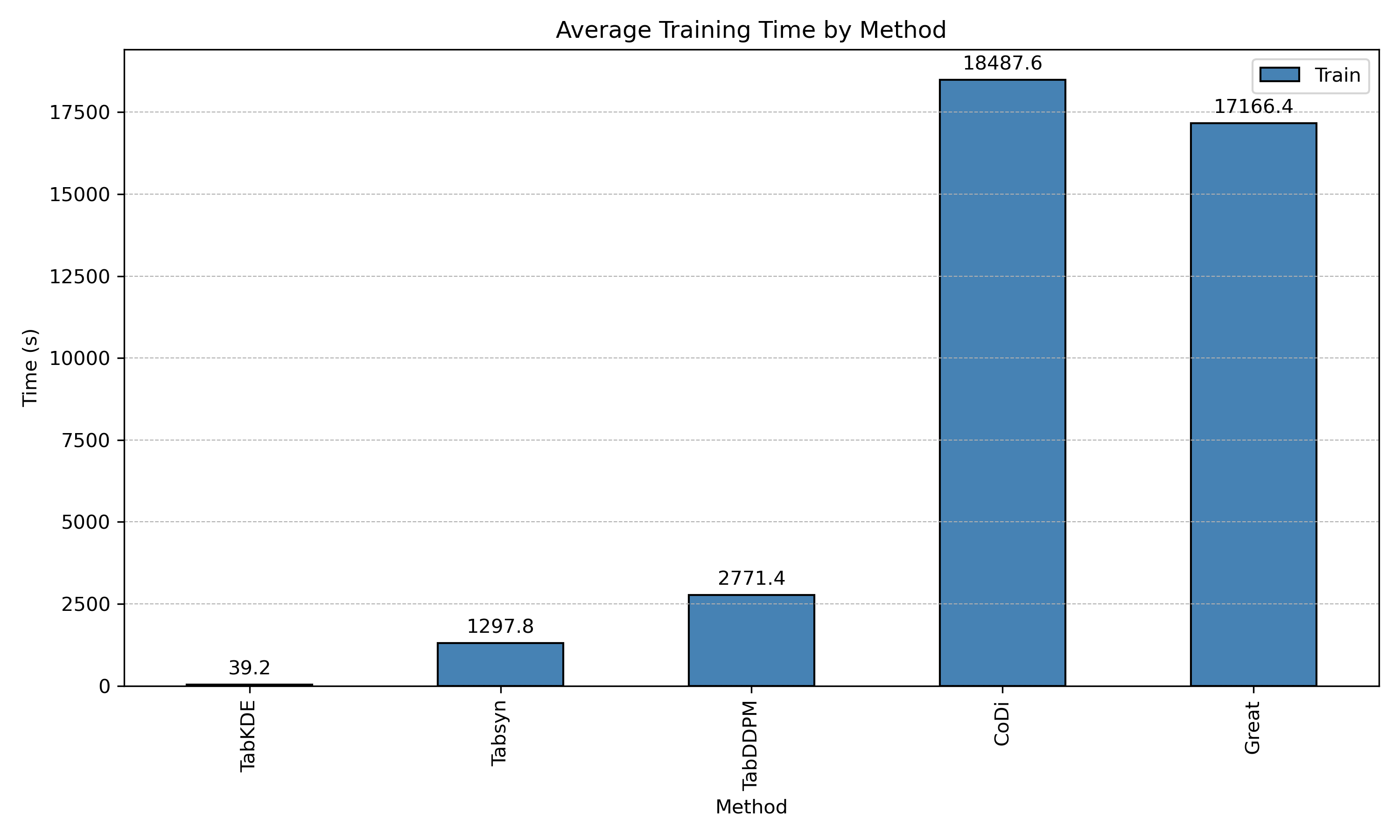}
        \caption{Training time}
        \label{app:fig:training_time}
    \end{subfigure}
    \hfill
    \begin{subfigure}[b]{0.45\linewidth}
        \includegraphics[width=\linewidth]{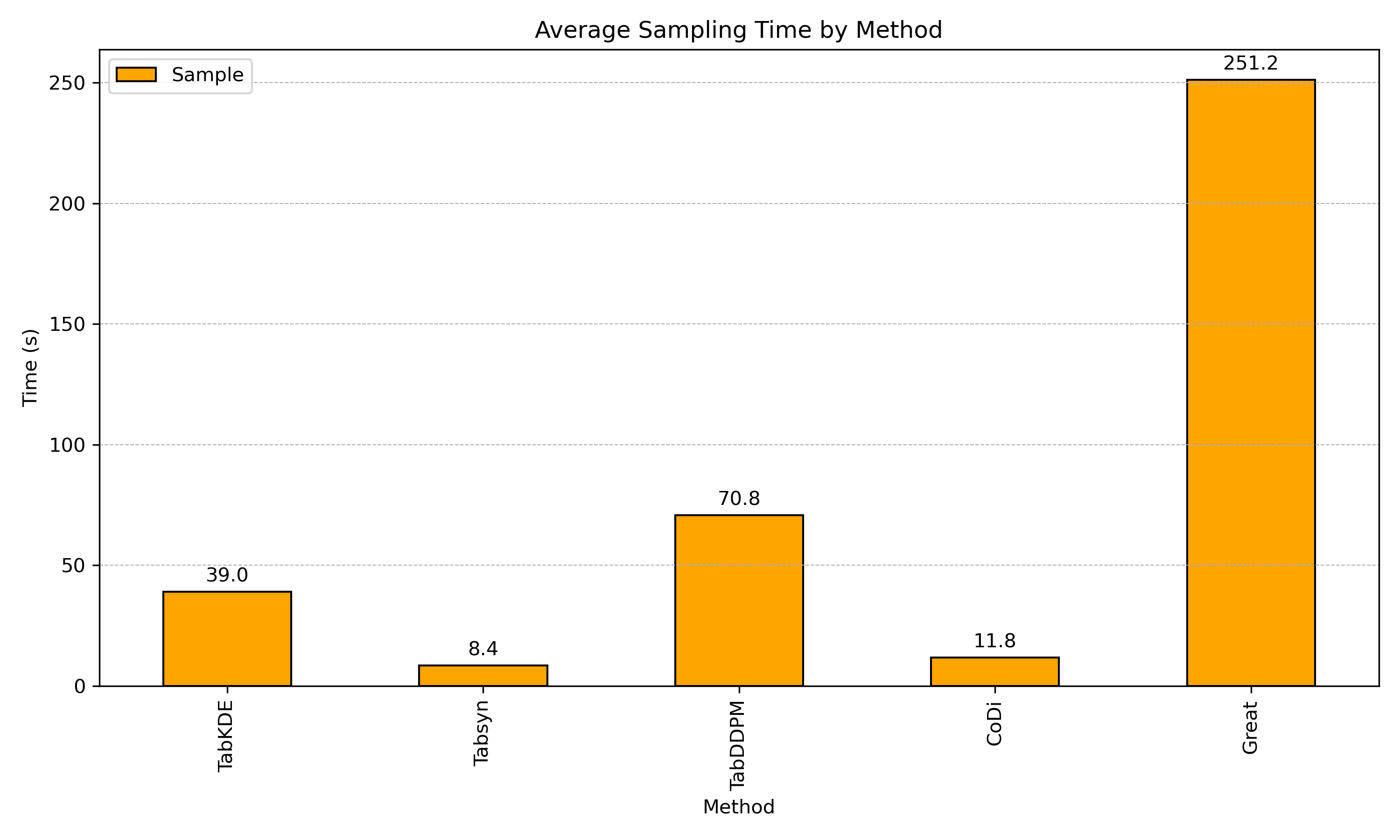}
        \caption{Sampling time}
        \label{app:fig:sampling_time}
    \end{subfigure}
    \caption{Average training and sampling time over Adult, Default, Shoppers, and Magic for different methods.}
    \label{app:fig:training_sampling_time}
\end{figure}

For a direct comparison, we evaluate \TabKDE alongside the baselines \TabSYN, TabDDPM, CoDi, and GReat on the Adult, Default, Shoppers, and Magic datasets, using an NVIDIA RTX A5000 GPU with 24GB of memory and a maximum power draw of 230W, under the same experimental settings as \TabSYN. See Tables~\ref{app:tab:training_sampling_time} and \ref{app:tab:stds_training_sampling_time} and Figure~\ref{app:fig:training_sampling_time} for details.  \TabKDE is orders of magnitude faster in training, but on par with others in sampling time -- we generate samples sequentially, and did not optimize for the GPU.

\begin{table}[h!]
\caption{Average training and sampling time for each method.}
\centering
\begin{tabular}{lcc}
\hline
\textbf{Method} & \textbf{Train Time (s)} & \textbf{Sample Time (s)} \\
\hline
Great    & 17112.4 & 251.2 \\
Codi     &  18487.6 & 11.8 \\
TabDDPM  & 2771.4 & 70.8  \\
\TabSYN   & 1297.8 & 8.4   \\
\hline
\TabKDE   & 39.2   & 39.0  \\
\hline
\end{tabular}
\label{app:tab:training_sampling_time}
\end{table}

\begin{table}[ht]
\centering
\caption{Average training and sampling times (in seconds)$\pm$ standard deviation for each dataset using the \TabKDE model. Values are reported as mean and standard deviation over 10 repeated runs.}
\label{app:tab:stds_training_sampling_time}
\begin{tabular}{lrr}
\toprule
Dataset & Train Time (s) & Sample Time (s) \\
\midrule
Adult    & 51.50\scriptsize{$\pm$1.08} & 50.30\scriptsize{$\pm$0.67} \\
Default  & 45.00\scriptsize{$\pm$0.67} & 55.10\scriptsize{$\pm$0.99} \\
Shoppers & 21.50\scriptsize{$\pm$0.53} & 18.20\scriptsize{$\pm$0.42} \\
Magic    & 33.00\scriptsize{$\pm$0.67} & 21.00\scriptsize{$\pm$0.67} \\
Beijing  & 63.00\scriptsize{$\pm$0.47} & 51.80\scriptsize{$\pm$0.79} \\
News     & 64.30\scriptsize{$\pm$1.16} & 166.80\scriptsize{$\pm$1.32} \\
\midrule
\textbf{Average} & \textbf{46.38\scriptsize{$\pm$0.29}} & \textbf{60.53\scriptsize{$\pm$0.42}} \\
\bottomrule
\end{tabular}
\end{table}

\section{Accuracy Evaluation}
\label{app:sec:accuracy}
In this section, we evaluate the quality of the generated synthetic data using three criteria: (1) marginal distribution alignment, (2) pairwise correlation matching, and (3) finally global alignment between synthetic and hold-out distributions is compared by how well a classifier can separate the distributions.  

\subsection{Marginal distribution alignment}
When evaluating synthetic tabular data, {\bf marginal distribution alignment score} assesses how closely each individual column matches its real‑data distribution represented by train data. Following what was done in the \TabSYN paper, we calculate the Kolmogorov–Smirnov (KS) distance for numerical attributes in $\Num$ and the Total Variation Distance for categorical and ordinal attributes in $\Cat$ and $\Ord$.  
Table~\ref{app:tab:shape} presents, for each dataset, the average marginal alignment errors across all features for each method. Table~\ref{app:std_for_tabkde_marginal} presents the performance of the \TabKDE model in aligning marginal distributions, averaged over 10 runs.

\begin{table}[h!]
\caption{Performance comparison on marginal distribution alignment (Error rate \%). Lower values indicate better performance. The values in parentheses denote the ratio relative to the smallest value.  Baseline values, unless stated otherwise, are taken from \citep{tabsyn2024}; the remaining scores (our methods) are obtained using their data split.
}
\scriptsize
\centering
\begin{tabular}{lcccccc|c}
\toprule
\textbf{Method} & \textbf{Adult} & \textbf{Default} & \textbf{Shoppers} & \textbf{Magic} & \textbf{Beijing} & \textbf{News} & \textbf{Average} \\
\midrule
SMOTE \tiny{(our reproduction)} & 1.63 ({\tiny 2.55}) & 1.70 ({\tiny 1.49}) & 2.66 ({\tiny 2.16}) & 1.37 ({\tiny 1.93}) & 2.10 ({\tiny 1.62}) & 5.47 ({\tiny 3.18}) & 2.49 ({\tiny 1.75}) \\
CTGAN & 16.84 ({\tiny 26.31}) & 16.83 ({\tiny 14.76}) & 21.15 ({\tiny 17.20}) & 9.81 ({\tiny 13.82}) & 21.39 ({\tiny 16.45}) & 16.09 ({\tiny 9.35}) & 17.02 ({\tiny 11.99}) \\
TVAE & 14.22 ({\tiny 22.22}) & 10.17 ({\tiny 8.92}) & 24.51 ({\tiny 19.93}) & 8.25 ({\tiny 11.62}) & 19.16 ({\tiny 14.74}) & 16.62 ({\tiny 9.66}) & 15.49 ({\tiny 10.91}) \\
GOGGLE & 16.97 ({\tiny 26.52}) & 17.02 ({\tiny 14.93}) & 22.33 ({\tiny 18.15}) & 1.90 ({\tiny 2.68}) & 16.93 ({\tiny 13.02}) & 25.32 ({\tiny 14.72}) & 16.74 ({\tiny 11.79}) \\
GReaT & 12.12 ({\tiny 18.94}) & 19.94 ({\tiny 17.49}) & 14.51 ({\tiny 11.80}) & 16.16 ({\tiny 22.76}) & 8.25 ({\tiny 6.35}) & -- & 14.20 ({\tiny 10.00}) \\
STaSy & 11.29 ({\tiny 17.64}) & 5.77 ({\tiny 5.06}) & 9.37 ({\tiny 7.62}) & 6.29 ({\tiny 8.86}) & 6.71 ({\tiny 5.16}) & 6.89 ({\tiny 4.01}) & 7.72 ({\tiny 5.44}) \\
CoDi & 21.38 ({\tiny 33.41}) & 15.77 ({\tiny 13.82}) & 31.84 ({\tiny 25.89}) & 11.56 ({\tiny 16.28}) & 16.94 ({\tiny 13.03}) & 32.27 ({\tiny 18.78}) & 21.63 ({\tiny 15.23}) \\
TabDDPM & 1.75 ({\tiny 2.73}) & 1.57 ({\tiny 1.38}) & 2.72 ({\tiny 2.21}) & 1.01 ({\tiny 1.42}) & 1.30 ({\tiny 1.00}) & 78.75 ({\tiny 45.83}) & 14.52 ({\tiny 10.23}) \\
\TabSYN ({\tiny Our reproduction}) &  0.64 ({\tiny 1.00}) &  1.14 ({\tiny 1.00}) & 1.23 ({\tiny 1.00}) &  0.98 ({\tiny 1.38}) &  2.79 ({\tiny 2.15}) &  1.72 ({\tiny 1.00}) & 1.42 ({\tiny 1.00}) \\
\hline
\PGETabsyn  & 11.21 ({\tiny 17.51}) & 7.66 ({\tiny 6.72}) & 12.98 ({\tiny 10.55}) & 1.36 ({\tiny 1.74}) & 2.65 ({\tiny 2.39}) & 18.65 ({\tiny 10.84}) & 9.08 ({\tiny 6.39})\\
\CopulaDiff &  2.01 ({\tiny 3.14}) &  1.47 ({\tiny 1.29}) &  2.47 ({\tiny 2.01}) &  0.94 ({\tiny 1.32}) &  2.13 ({\tiny 1.64}) &  2.44 ({\tiny 1.42}) &  1.91 ({\tiny 1.35}) \\
\VAESimpleKDE &  3.23 ({\tiny 5.05}) &  7.72 ({\tiny 6.77}) &  6.78 ({\tiny 5.51}) &  3.12 ({\tiny 4.39}) &  7.12 ({\tiny 5.48}) &  10.03 ({\tiny 5.83}) &  6.33 ({\tiny 4.46}) \\
\VAETabKDE &  3.80 ({\tiny 5.94}) &  5.84 ({\tiny 5.12}) &  6.31 ({\tiny 5.13}) &  0.71 ({\tiny 1.0}) &  4.94 ({\tiny 3.80}) &  4.45 ({\tiny 2.59}) &  4.34 ({\tiny 3.06}) \\
\simpKDE &  1.92 ({\tiny 3.00}) &  3.33 ({\tiny 2.92}) &  3.12 ({\tiny 2.54}) &  3.59 ({\tiny 5.06}) &  10.32 ({\tiny 7.94}) &  7.36 ({\tiny 4.28}) &  4.94 ({\tiny 3.48}) \\
\RandomCoresetTabKDE &  1.61 ({\tiny 2.52}) &  1.76 ({\tiny 1.54}) &  2.54 ({\tiny 2.07}) &  1.01 ({\tiny 1.42}) &  1.70 ({\tiny 1.31}) &  2.59 ({\tiny 1.51}) & 1.87 ({\tiny 3.48}) \\
\CoresetTabKDE &  3.63 ({\tiny 5.67}) &  3.29 ({\tiny 2.89}) &  3.23 ({\tiny 2.63}) &  1.08 ({\tiny 1.52}) &  3.20 ({\tiny 2.46}) &  2.87 ({\tiny 1.67}) & 2.88 ({\tiny 2.03}) \\
\TabKDE \tiny{(change the seed)} &   2.61 ({\tiny 4.08}) &   6.05 ({\tiny 5.31}) &  4.59 ({\tiny 3.73}) &  6.41 ({\tiny 9.03}) & 3.30 ({\tiny 2.54}) & 8.18 ({\tiny 4.76}) & 5.19 ({\tiny 3.65}) \\
\TabKDE \tiny{(keep the seed)} & 1.45 ({\tiny 2.27}) & 2.32 ({\tiny 2.04}) & 2.55 ({\tiny 2.07}) &  1.60 ({\tiny 2.25}) & 1.82 ({\tiny 1.40}) &  5.75 ({\tiny 3.34}) & 2.58 ({\tiny 1.82}) \\
\TabKDE &  1.56 ({\tiny 2.44}) &  1.55 ({\tiny 1.36}) &  2.44 ({\tiny 1.98}) &  0.78 ({\tiny 1.1}) & 1.37 ({\tiny 1.05}) &  2.52 ({\tiny 1.47}) &  1.70 ({\tiny 1.2}) \\
\arrayrulecolor{black}\bottomrule
\end{tabular}
\label{app:tab:shape}
\end{table}

\begin{table}[ht]
\centering
\scriptsize
\caption{Performance comparison on marginal distribution alignment (Error rate \%) for each dataset using  \TabKDE model. Values are reported as mean and standard deviation over 10 repeated runs.}
\label{app:std_for_tabkde_marginal}
\begin{tabular}{llllllll}
\toprule
           Metric &                          Adult &                         Default &                       Shoppers &                          Magic &                        Beijing &                           News &                        Average \\
\midrule
Marginal alignment error & 1.54\tiny{\,$\pm$\,0.03} &  1.53\tiny{\,$\pm$\,0.05} & 2.46\tiny{\,$\pm$\,0.09} & 0.80\tiny{\,$\pm$\,0.08} & 1.40\tiny{\,$\pm$\,0.04} & 2.53\tiny{\,$\pm$\,0.04} & 1.71\tiny{\,$\pm$\,0.05} \\
\bottomrule
\end{tabular}
\label{app:std_for_tabkde_marginal}
\end{table}



\begin{figure}[htp]
\centering
\begin{tabular}{cccc}
\includegraphics[width=0.225\linewidth]{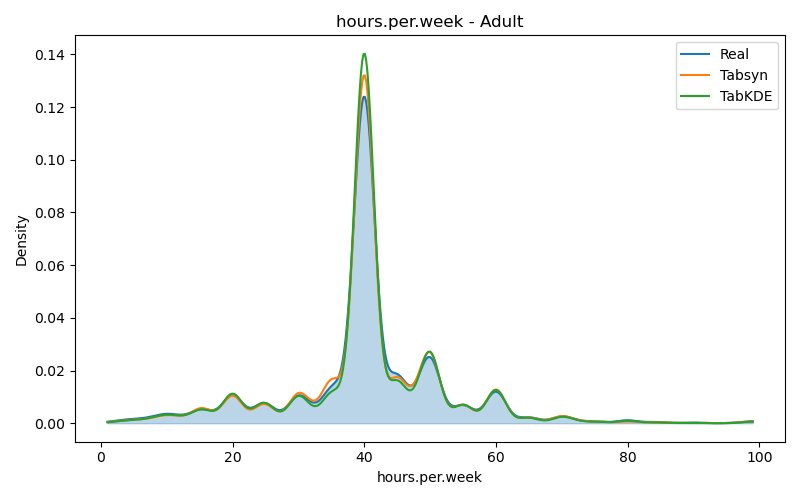} & 
\includegraphics[width=0.225\linewidth]{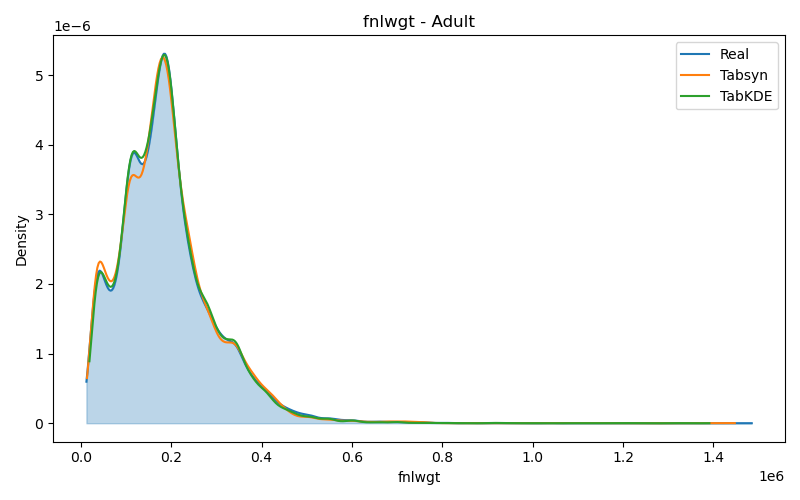} & 
\includegraphics[width=0.225\linewidth]{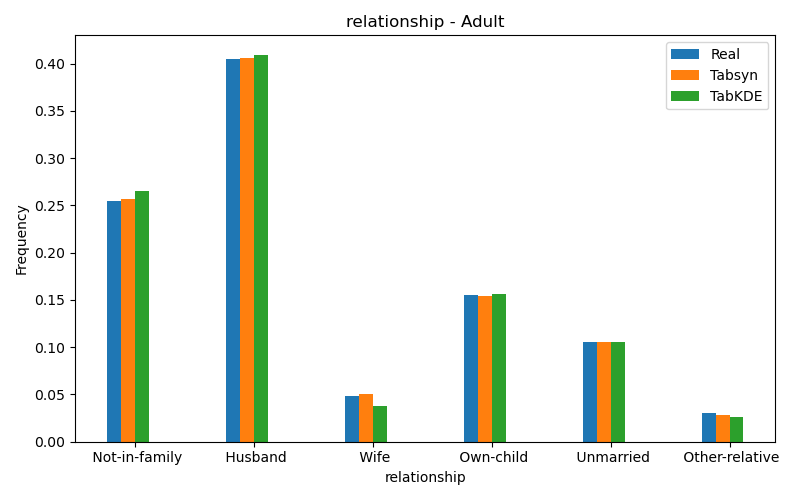} & 
\includegraphics[width=0.225\linewidth]{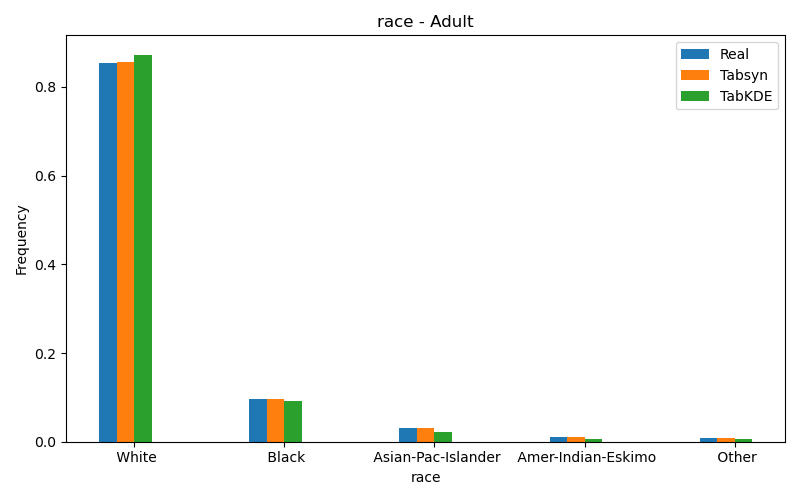} 
\\
\includegraphics[width=0.225\linewidth]{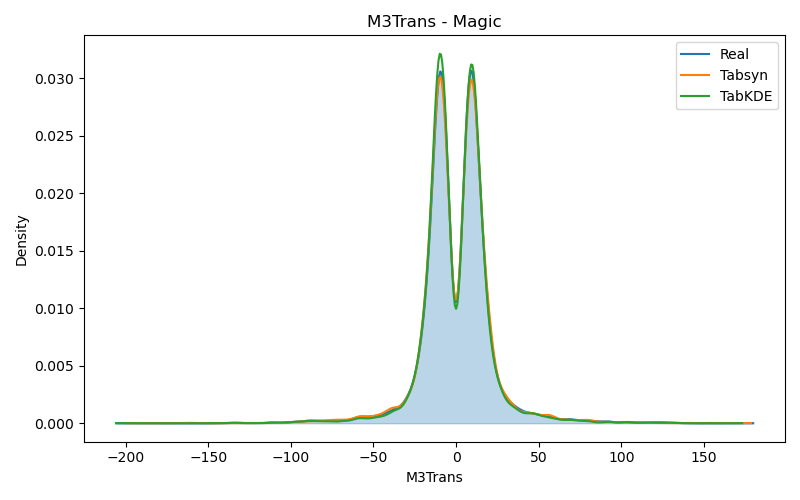} & \includegraphics[width=0.225\linewidth]{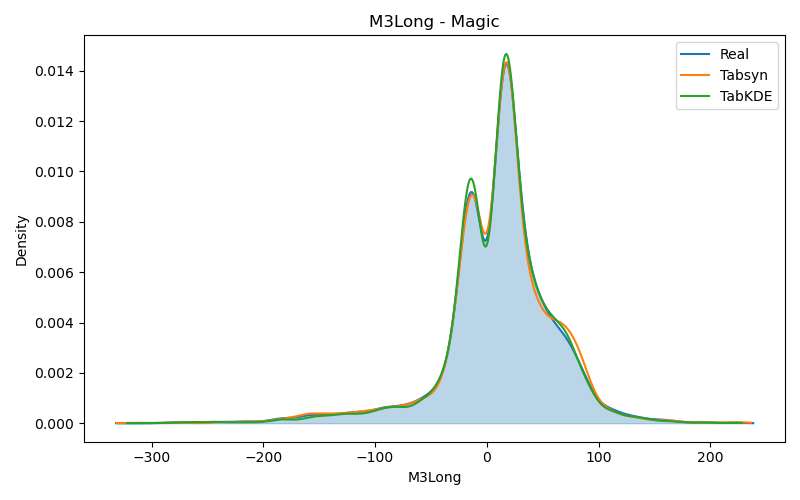}& \includegraphics[width=0.225\linewidth]{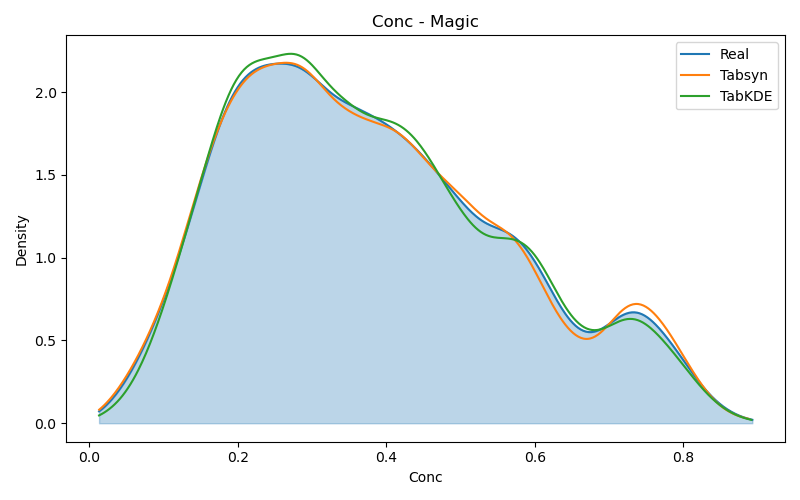} & \includegraphics[width=0.225\linewidth]{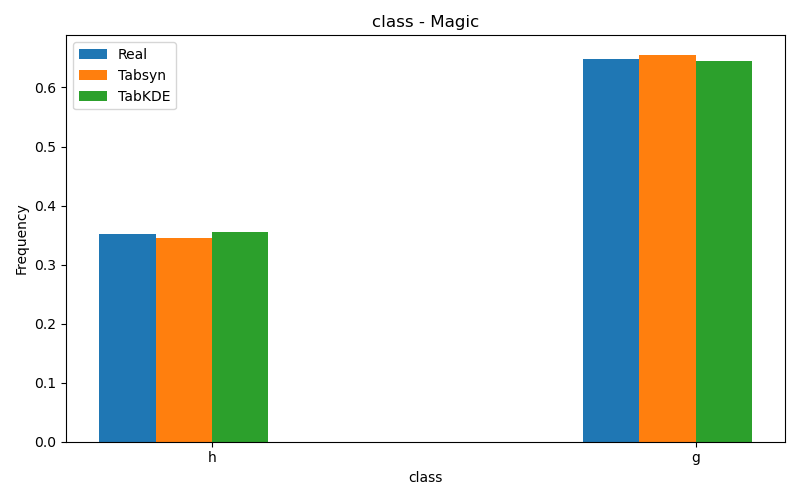} 
\\
\includegraphics[width=0.225\linewidth]{All_Marginals/news/timedelta.png} & \includegraphics[width=0.225\linewidth]{All_Marginals/news/title_sentiment_polarity.png} & 
\includegraphics[width=0.225\linewidth]{All_Marginals/news/data_channel.png} & 
\includegraphics[width=0.225\linewidth]{All_Marginals/news/weekday.png} 
\\
\includegraphics[width=0.225\linewidth]{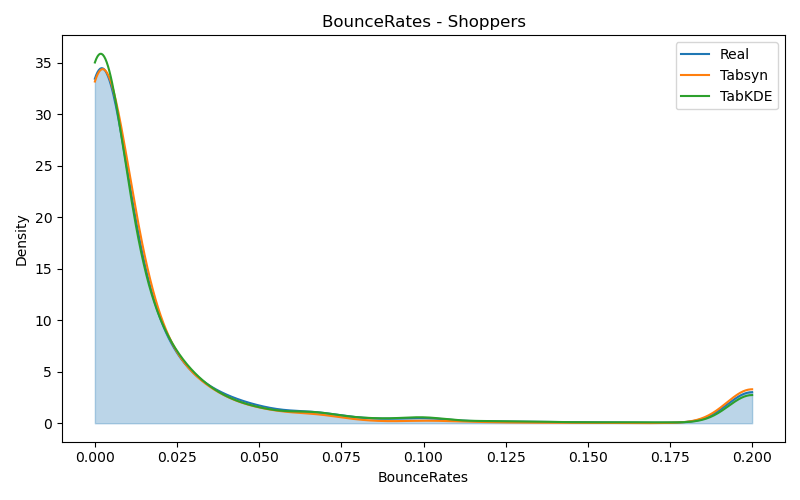} & 
\includegraphics[width=0.225\linewidth]{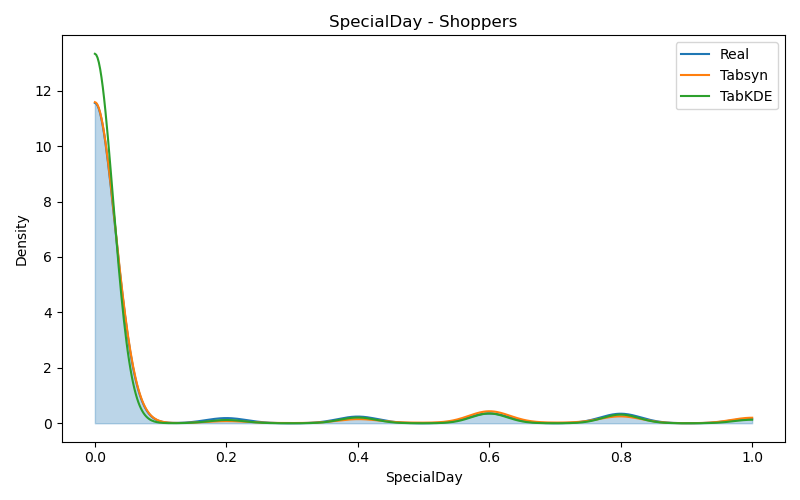}& \includegraphics[width=0.225\linewidth]{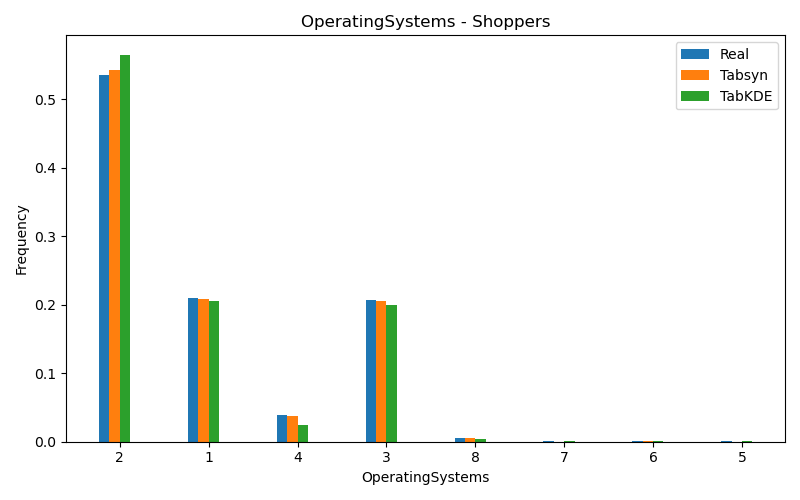} & \includegraphics[width=0.225\linewidth]{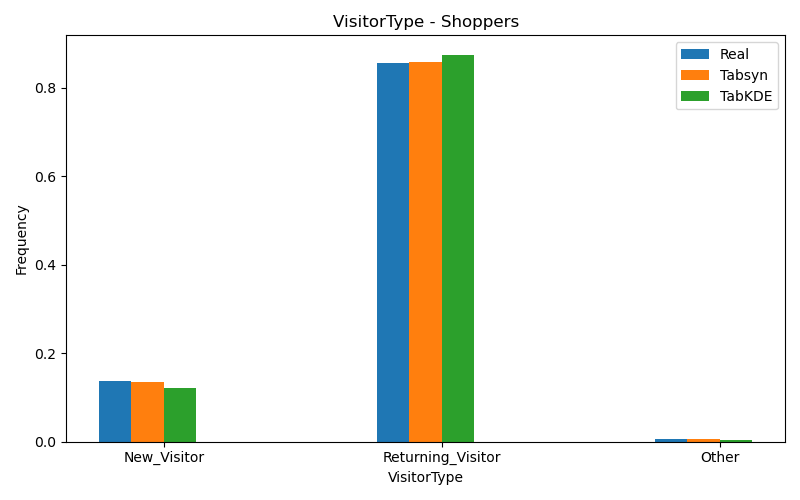} 
\\
\includegraphics[width=0.225\linewidth]{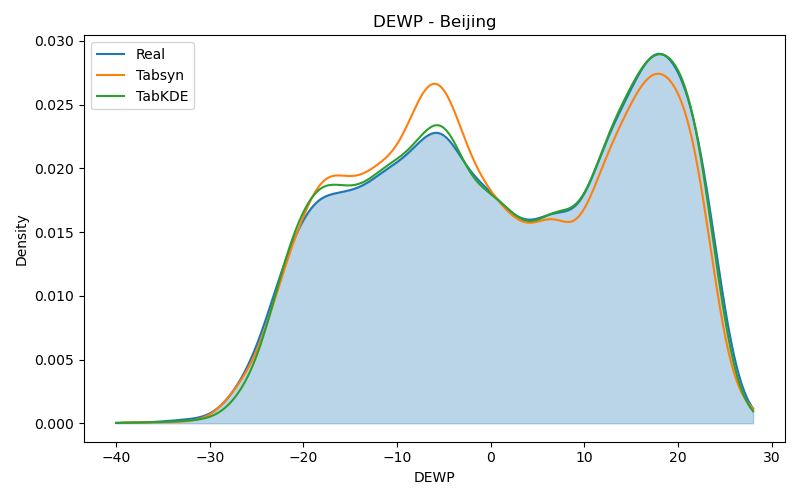} & 
\includegraphics[width=0.225\linewidth]{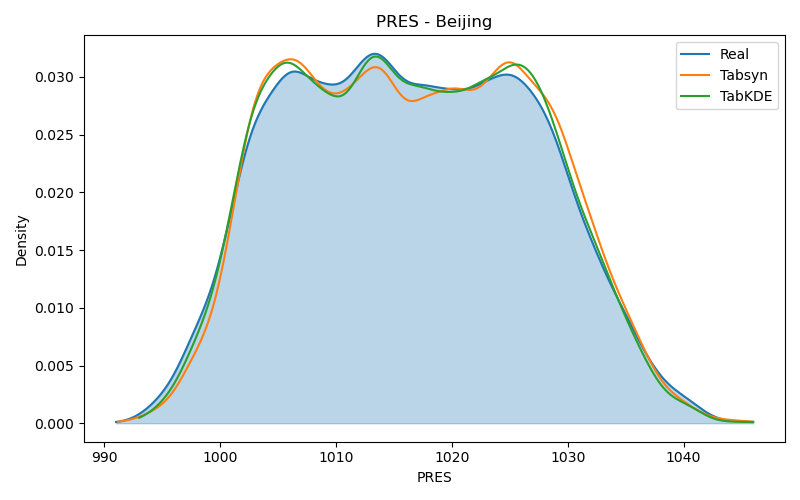} & 
\includegraphics[width=0.225\linewidth]{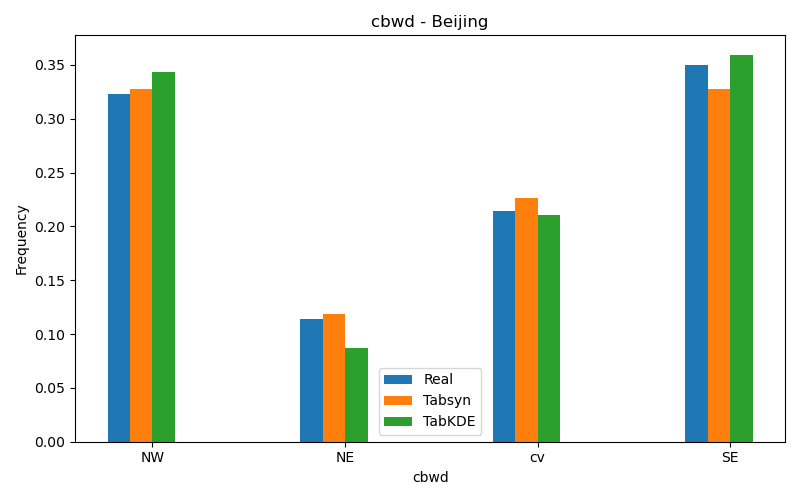} & \includegraphics[width=0.225\linewidth]{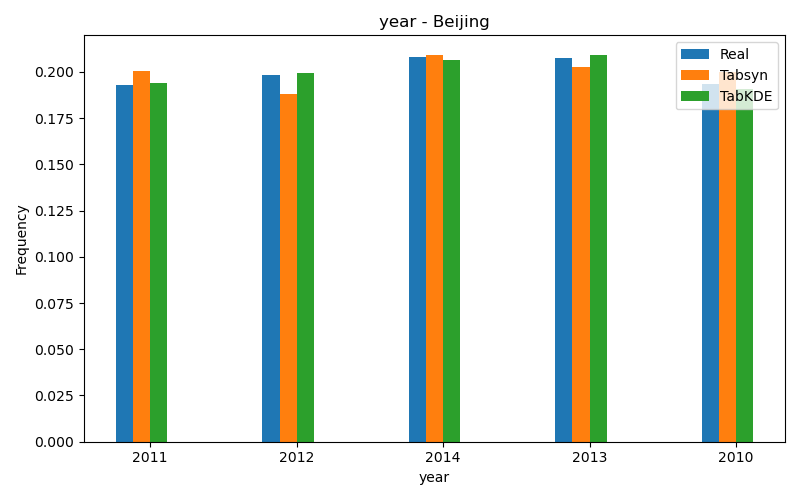} 
\\
\includegraphics[width=0.225\linewidth]{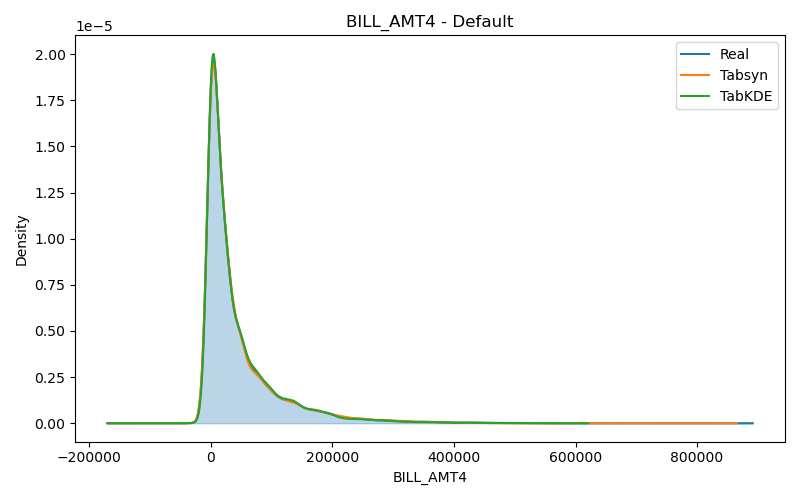} & \includegraphics[width=0.225\linewidth]{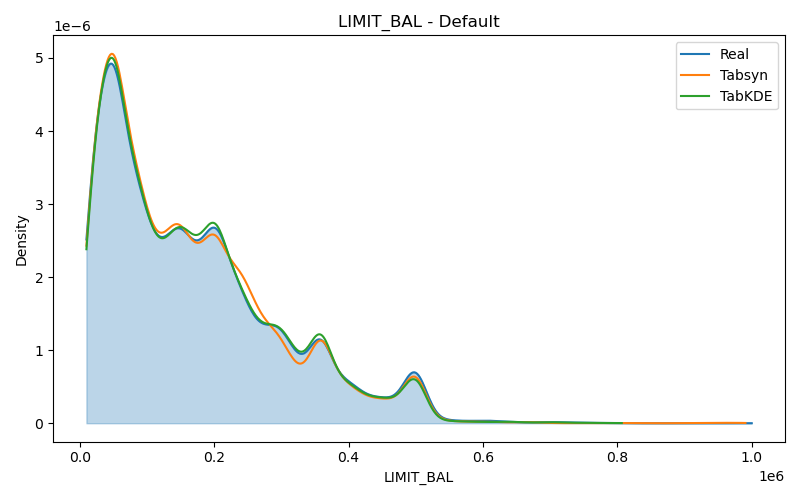} & 
\includegraphics[width=0.225\linewidth]{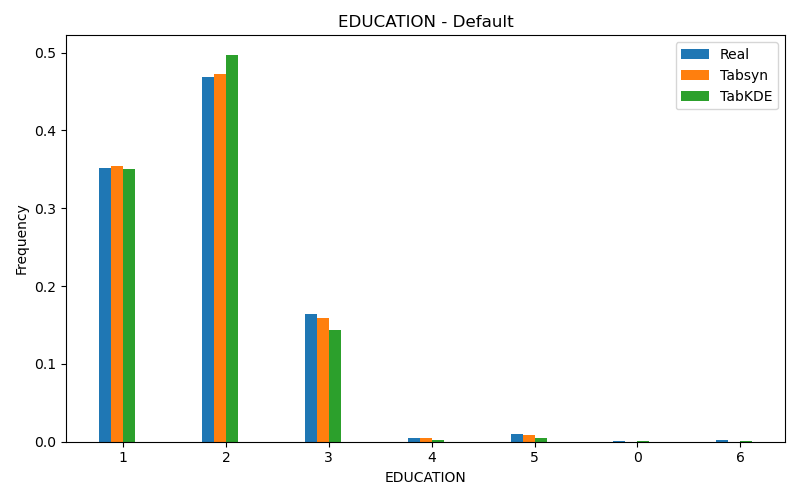} & \includegraphics[width=0.225\linewidth]{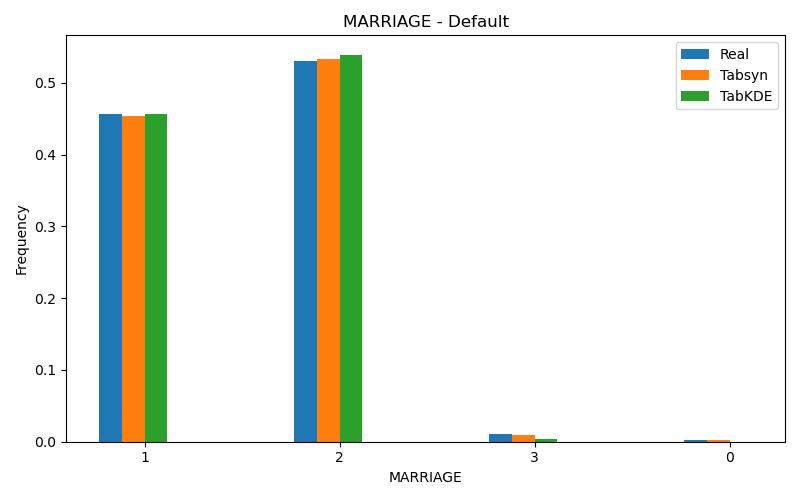} 
\\
\includegraphics[width=0.225\linewidth]{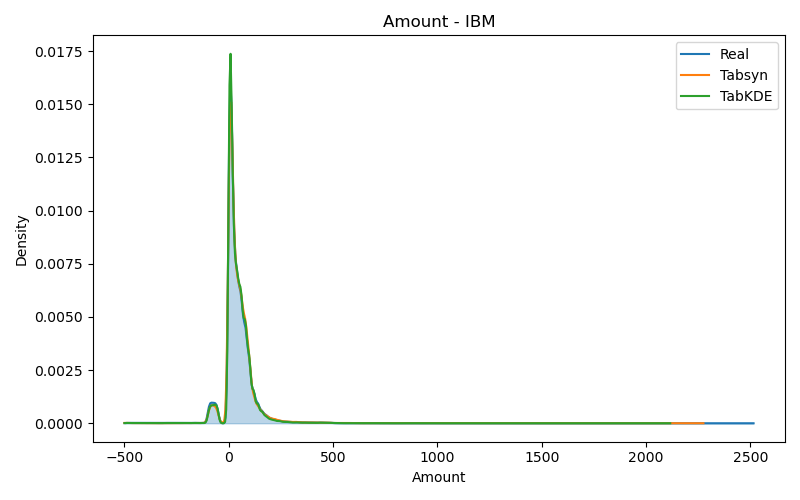} & \includegraphics[width=0.225\linewidth]{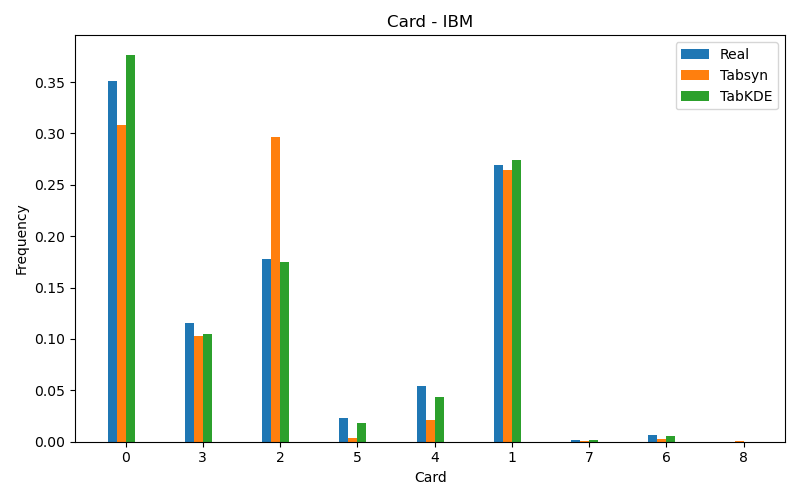}& \includegraphics[width=0.225\linewidth]{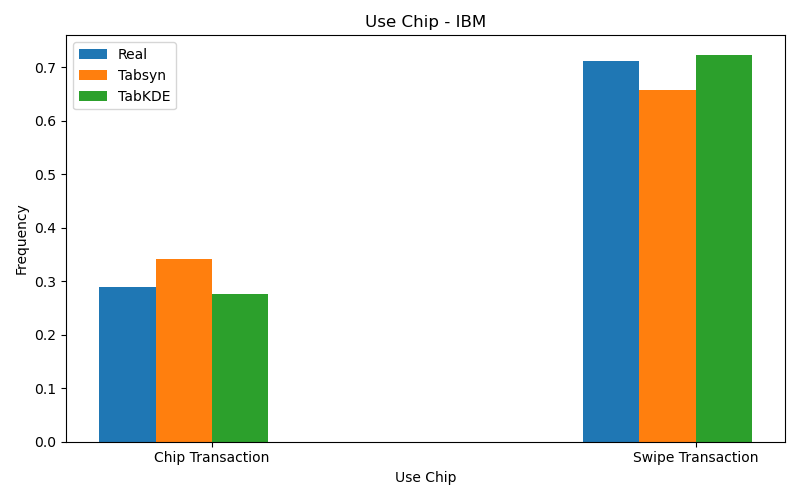} & \includegraphics[width=0.225\linewidth]{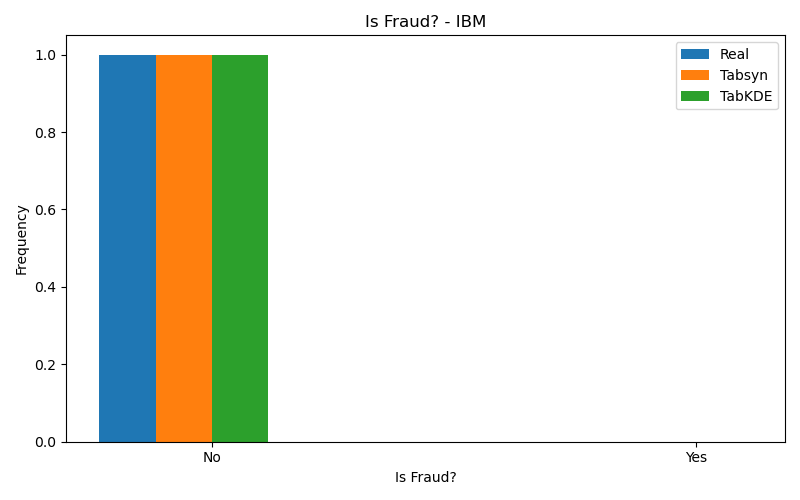} 
\\
\end{tabular}
\caption{Marginals comparison between real data (blue), \TabKDE (green), and \TabSYN (orange).  Each row is a data set, and sample column marginals are shown for each; some categorical and some numerical.  For IBM data set (last row), \TabSYN is replaced with \PGETabsyn since \TabSYN runs out of memory.}
\label{app:fig:marginals:TabsynVSTabKDE}
\end{figure}

Figure~\ref{app:fig:marginals:TabsynVSTabKDE} provides a visual comparison between  some representative selected real marginal distributions and those generated by \TabSYN (orange) and \TabKDE (green) against the real data distributions (blue).  Each row shows 4 columns from a data set.   For IBM data set (bottom row) we use \CopulaDiff (orange) instead of \TabSYN since it cannot scale to this large data set.  It is apparent that \TabKDE usually does as well and better than \TabSYN.  In particular, on numerical data (where continuous distributions are shown), \TabKDE appears to match the real data much closer, but on categorical data, and when there are spikes in numerical data, \TabKDE can have a bit more error.  
Since the KS distance is a worst case, it is very unforgiving for such errors on discrete data, and explains why \TabKDE and \TabSYN appear comparable in these marginal plots, but \TabSYN has consistently smaller scores in Table \ref{app:tab:shape}.  
An average error measure on on numerical data should show an advantage for \TabKDE.  

\subsection{Pairwise correlation alignment} 
\label{app:pairwise}
\begin{figure}
    \centering
    \includegraphics[width=1\linewidth]{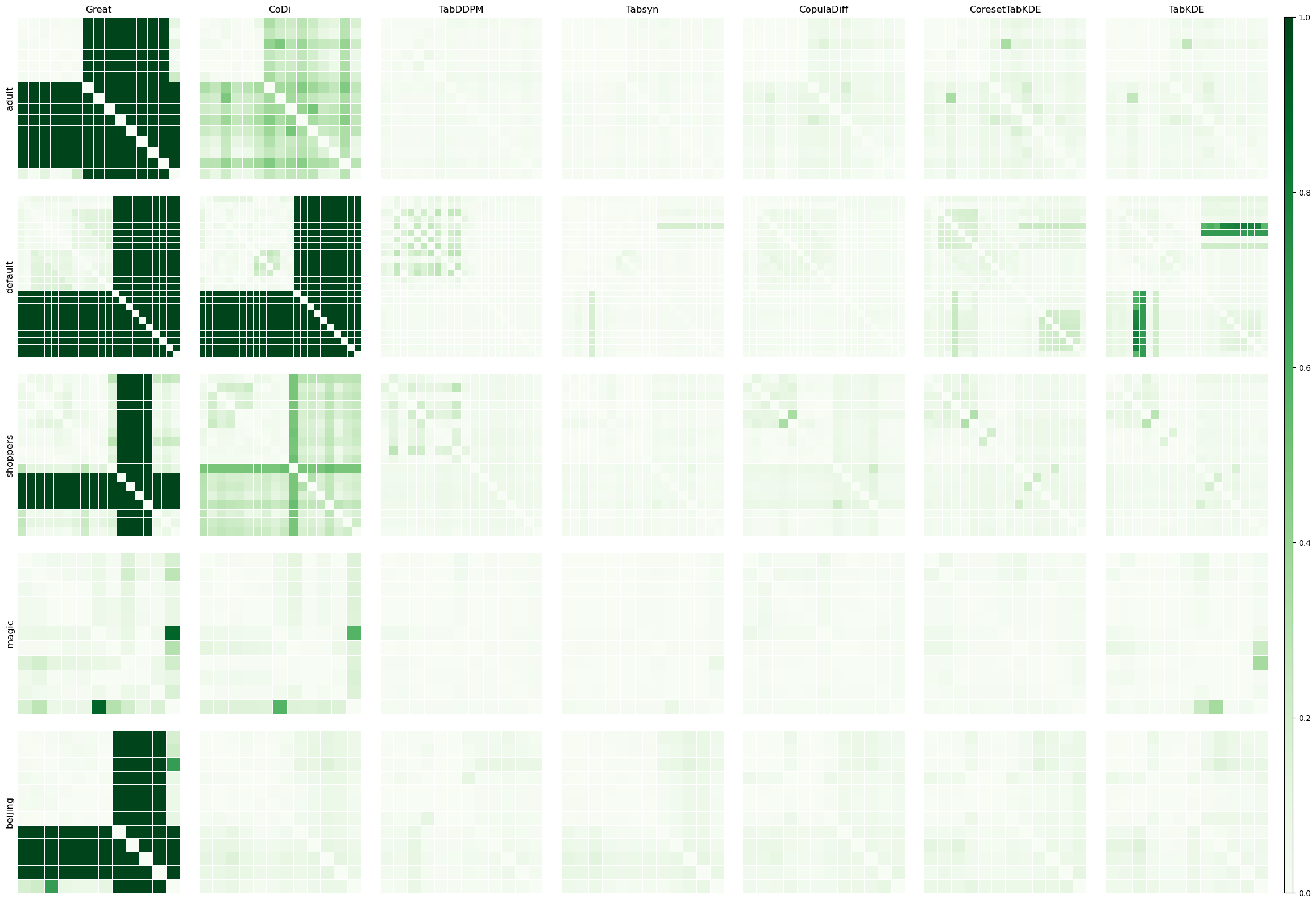}
    \caption{Pairwise correlation Divergence plots for each dataset and methods.  }\label{app:fig:div_correlation_2}
\end{figure}

We next measure pairwise correlation between columns.  For numerical-numerical pairs, we can use standard Pearson correlations.  However for pairs that involve categorical (and to align with measures in the \TabSYN paper, we treat ordinal as categorical), we use contingency-table Total Variation Distances.  
In both metrics, smaller error values indicate that the synthetic table is more faithful to the original data. Table~\ref{app:tab:trend} presents, for each dataset, the average pairwise correlation alignment errors, computed as \((1 - \text{score})\), across all features for each method. 
A heatmap visualization of the divergence between the pairwise correlations in the real and synthetic data is presented in Figure~\ref{app:fig:div_correlation_2}. 
Table~\ref{app:std_for_tabkde_pairwise} presents the performance of the \TabKDE model in Pairwise correlation alignment error  (Error rate \%), averaged over 10 runs.

We observe that \TabKDE has better pairwise correlation alignment than all methods except \TabSYN, and is comparable to SMOTE; both \TabKDE and SMOTE have about $2.5$ the correlation discrepancy as \TabSYN.  The main poor correlation for \TabKDE appears in Default data set, and with `BILL\_AMT3' and `BILL\_AMT4' variables which have similar but less challenges for \TabSYN, as well as GreaT and CoDi.

\begin{table}[h!]
\caption{Performance comparison of tabular data synthesis methods based on Pairwise correlation alignment error  (Error rate \%). Lower values indicate better performance. The values in parentheses denote the ratio relative to the smallest value. 
Baseline values, unless stated otherwise, are taken from \citep{tabsyn2024}; the remaining scores (our methods) are obtained using their data split.
}
\scriptsize
\centering
\begin{tabular}{lcccccc|c}
\toprule
\textbf{Method} & \textbf{Adult} & \textbf{Default} & \textbf{Shoppers} & \textbf{Magic} & \textbf{Beijing} & \textbf{News} & \textbf{Average} \\
\midrule
SMOTE \tiny{(Our reproduction)} & 4.3 ({\tiny 2.67}) & 11.54 ({\tiny 5.11}) & 3.68 ({\tiny 1.47}) & 1.88 ({\tiny 2.29}) & 3.3 ({\tiny 1.22}) & 1.67 ({\tiny 1.25}) & 4.39 ({\tiny 1.99}) \\
CTGAN & 20.23 ({\tiny 12.57}) & 26.95 ({\tiny 11.92}) & 13.08 ({\tiny 5.23}) & 7.0 ({\tiny 8.54}) & 22.95 ({\tiny 8.47}) & 5.37 ({\tiny 4.01}) & 15.93 ({\tiny 7.21}) \\
TVAE & 14.15 ({\tiny 8.79}) & 19.5 ({\tiny 8.63}) & 18.67 ({\tiny 7.47}) & 5.82 ({\tiny 7.10}) & 18.01 ({\tiny 6.65}) & 6.17 ({\tiny 4.60}) & 13.72 ({\tiny 6.21}) \\
GOGGLE & 45.29 ({\tiny 28.13}) & 21.94 ({\tiny 9.71}) & 23.9 ({\tiny 9.56}) & 9.47 ({\tiny 11.55}) & 45.94 ({\tiny 16.95}) & 23.19 ({\tiny 17.31}) & 28.29 ({\tiny 12.8}) \\
GReaT & 17.59 ({\tiny 10.93}) & 70.02 ({\tiny 30.98}) & 45.16 ({\tiny 18.06}) & 10.23 ({\tiny 12.48}) & 59.6 ({\tiny 21.99}) & -- & 40.52 ({\tiny 18.33}) \\
STaSy & 14.51 ({\tiny 9.01}) & 5.96 ({\tiny 2.64}) & 8.49 ({\tiny 3.39}) & 6.61 ({\tiny 8.05}) & 8.0 ({\tiny 2.96}) & 3.07 ({\tiny 2.29}) & 7.77 ({\tiny 3.52}) \\
CoDi & 22.49 ({\tiny 13.98}) & 68.41 ({\tiny 30.27}) & 17.78 ({\tiny 7.11}) & 6.53 ({\tiny 7.97}) & 7.07 ({\tiny 2.61}) & 11.1 ({\tiny 8.28}) & 22.23 ({\tiny 10.06}) \\
TabDDPM & 3.01 ({\tiny 1.87}) & 4.89 ({\tiny 2.16}) & 6.61 ({\tiny 2.64}) & 1.7 ({\tiny 2.07}) & 2.71 ({\tiny 1.00}) & 13.16 ({\tiny 9.82}) & 5.34 ({\tiny 2.42}) \\
\TabSYN{\tiny (Our reproduction)} &  1.61 ({\tiny 1.00}) &  2.26 ({\tiny 1.00}) &  2.5 ({\tiny 1.00}) &  0.82 ({\tiny 1.00}) &  4.7 ({\tiny 1.73}) &  1.34 ({\tiny 1.00}) &  2.21 ({\tiny 1.00}) \\
\hline
\PGETabsyn  & 7.14 ({\tiny 4.43}) & 15.19 ({\tiny 6.72})& 7.56  ({\tiny 3.02})& 2.67 ({\tiny 2.76})& 3.49 ({\tiny 1.29})& 4.60 ({\tiny 3.42})& 6.78 ({\tiny 3.07})\\
\CopulaDiff &  4.61 ({\tiny 2.86}) &  3.29 ({\tiny 1.46}) &  5.3 ({\tiny 2.12}) &  1.72 ({\tiny 2.10}) &  4.5 ({\tiny 1.66}) &  2.1 ({\tiny 1.57}) &  3.59 ({\tiny 1.62}) \\
\VAESimpleKDE &  9.86 ({\tiny 6.12}) &  12.88 ({\tiny 5.70}) &  9.51 ({\tiny 3.80}) &  3.12 ({\tiny 3.80}) &  11.51 ({\tiny 4.25}) &  4.05 ({\tiny 3.02}) &  8.49 ({\tiny 3.84}) \\
\VAETabKDE &  7.23 ({\tiny 4.49}) &  12.71 ({\tiny 5.62}) &  9.68 ({\tiny 3.87}) &  3.95 ({\tiny 4.82}) &  9.87 ({\tiny 3.64}) &  3.67 ({\tiny 2.74}) &  7.85 ({\tiny 3.55}) \\
\simpKDE &  4.64 ({\tiny 2.88}) &  5.16 ({\tiny 2.28}) &  5.26 ({\tiny 2.10}) &  3.3 ({\tiny 4.02}) &  4.72 ({\tiny 1.74}) &  2.96 ({\tiny 2.21}) &  4.34 ({\tiny 1.96}) \\
\RandomCoresetTabKDE &  3.93 ({\tiny 2.44}) &  13.66 ({\tiny 6.04}) &  4.27 ({\tiny 1.71}) &  4.76 ({\tiny 5.80}) &  4.05 ({\tiny 1.49}) &  2.61 ({\tiny 1.95}) & 5.46 ({\tiny 2.47}) \\
\CoresetTabKDE &  6.3 ({\tiny 3.91}) &  9.91 ({\tiny 4.39}) &  5.77 ({\tiny 2.31}) &  2.18 ({\tiny 2.66}) &  5.86 ({\tiny 2.16}) &  2.82 ({\tiny 2.10}) & 5.47 ({\tiny 2.48}) \\

\TabKDE \tiny{(change the seed)} &   5.16 ({\tiny 3.20}) &   13.82 ({\tiny 6.12}) & 6.10 ({\tiny 2.44}) &  6.36 ({\tiny 7.76}) & 5.32 ({\tiny 1.96}) &  3.88 ({\tiny 2.90}) &  6.77 ({\tiny 3.06}) \\
\TabKDE \tiny{(keep the seed)} & 3.51 ({\tiny 2.18}) &  12.93 ({\tiny 5.72}) & 4.20 ({\tiny 1.68}) &  3.48 ({\tiny 4.24}) & 3.85 ({\tiny 1.42}) & 3.58 ({\tiny 2.67}) &  5.26 ({\tiny 2.38}) \\
\TabKDE &  4.51 ({\tiny 2.80}) &  9.93 ({\tiny 4.40}) &  4.31 ({\tiny 1.72}) &  2.72 ({\tiny 3.32}) &  3.74 ({\tiny 1.38}) &  2.83 ({\tiny 2.11}) & 4.67 ({\tiny 2.11}) \\
\arrayrulecolor{black}\bottomrule
\end{tabular}
\label{app:tab:trend}
\end{table}

\begin{table}[ht]
\centering
\scriptsize
\caption{Performance comparison on pairwise correlation alignment (Error rate \%) for each dataset using \TabKDE model. Values are reported as mean and standard deviation over 10 repeated runs. }
\begin{tabular}{llllllll}
\toprule
           Metric &                          Adult &                         Default &                       Shoppers &                          Magic &                        Beijing &                           News &                        Average \\
\midrule
Pairwise Corr. Error & 4.05\tiny{\,$\pm$\,0.27} & 11.33\tiny{\,$\pm$\,1.49} & 4.39\tiny{\,$\pm$\,0.16} & 2.80\tiny{\,$\pm$\,0.68} & 3.80\tiny{\,$\pm$\,0.22} & 2.95\tiny{\,$\pm$\,0.17} & 4.89\tiny{\,$\pm$\,0.47} \\
\bottomrule
\end{tabular}
\label{app:std_for_tabkde_pairwise}
\end{table}

\begin{figure}
    \centering
    \includegraphics[width=0.9\linewidth]{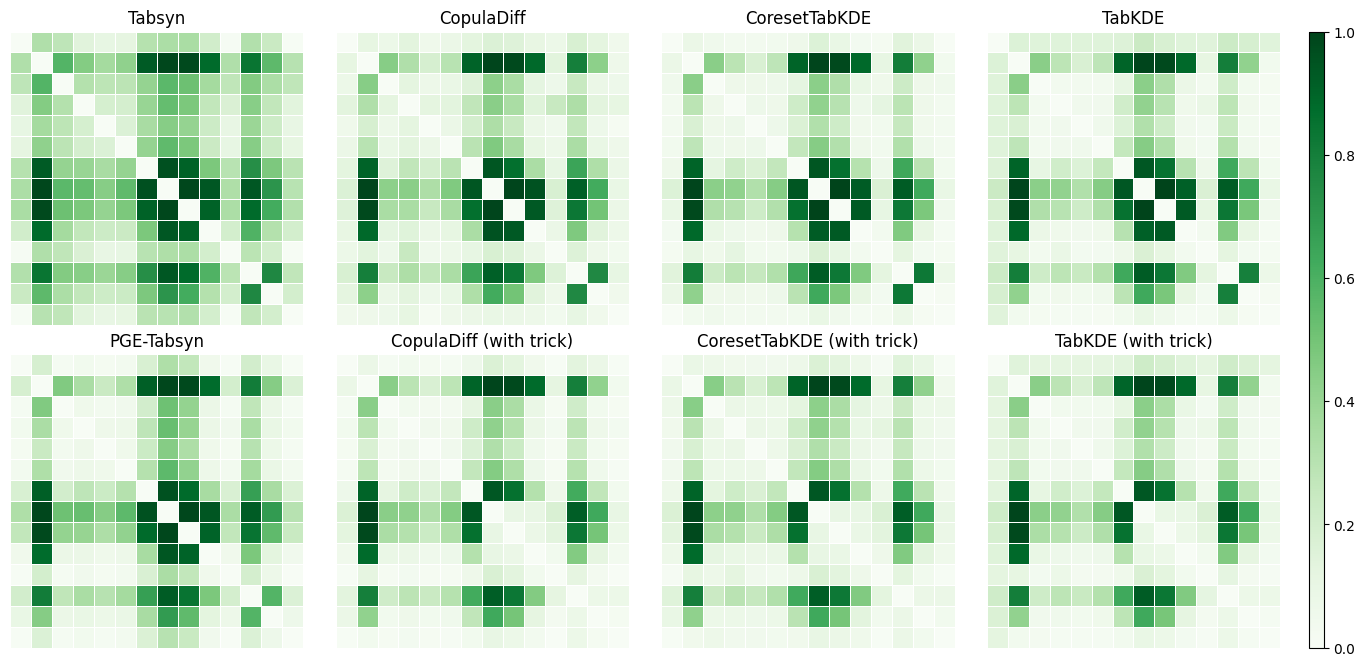}
\caption{Pairwise correlation Divergence plots for IBM dataset and methods:  \TabSYN (without modeling trick), \PGETabsyn (without modeling trick), \CopulaDiff (with and without modeling trick), \CoresetTabKDE, \TabKDE (with and without modeling trick). For \CoresetTabKDE, the size of coreset and the bandwidth are {5},000 and $.15$, respectively. Models that use the reduced modeling trick are indicated by “with trick” in parentheses.}

\label{app:fig:IBM-correlation}
\end{figure}

We also compare against some variants on the larger IBM dataset.  Recall that on our laptop CPU, neither \TabSYN or SMOTE can run on this data set -- they both run out of memory.  
Instead we compare \TabKDE against our baselines including \CopulaDiff and \PGETabsyn.  
Also, note that we apply a modeling trick with Zip / Merchant State / Merchant City and with MCC / Merchant Name with \TabKDE but not \PGETabsyn and \TabSYN (on GPU). 
Recall also that \TabKDE (about 10 minutes training time on CPU; 40 seconds on GPU) was much faster than either of \CopulaDiff (about 15 hours on CPU), \PGETabsyn (over 40 hours on CPU) or \TabSYN (OOM on CPU, about 20 minutes on GPU).  The average pairwise correlation alignment error for \TabSYN and \PGETabsyn (without modeling trick) are \%40.42  and \%30.53, respectively, while for \TabKDE and \CopulaDiff (with modeling trick), are is \%25.29 and \%22.59; see Table \ref{app:tab:IBM-accuracy}.  Indeed as shown in Figure \ref{app:fig:IBM-correlation} of pairwise correlation plots, the methods work largely similar except on the highly correlated pairs where we employ the modeling trick.

\begin{table}[h!]
\caption{Marginal and Pairwise accuracy on IBM dataset.
See the last paragraph of Subsection~\ref{app:datasets} for reduced modeling explanation.}
\centering
\begin{tabular}{lccc}
\toprule
\textbf{Method} & \textbf{Marginal Alig. Error} & \textbf{Pairwise Corr. Error} & \textbf{Reduced Modeling} \\
\hline
\TabSYN         &   16.99 &    40.42   & no\\
\PGETabsyn      &   9.29   &    30.53  & no\\
\CopulaDiff     &   6.81   &    29.42  & no\\
\hline
\TabKDE \tiny{(change the seed)}  &  7.26    &   34.42   & no\\
\TabKDE \tiny{(keep the seed)}    &  4.63    &  38.54    & no\\
\TabKDE \tiny{(change the seed)}  &  6.08    &  24.69    & yes\\
\TabKDE \tiny{(keep the seed)}    &  4.63    &  36.19  & yes\\
\hline
\TabKDE    &   5.36   &    27.56  & no\\
\CopulaDiff     &   3.59   &    22.59  & yes\\
\CoresetTabKDE     &   5.25   &    23.93  & yes\\
\TabKDE         &   3.58   &    25.29  & yes\\
\bottomrule
\end{tabular}
\label{app:tab:IBM-accuracy}
\end{table}


\subsection{Global Distribution Alignment} 
\label{app:global-align}
High‑quality synthetic data should be able to take the place of real data, letting us train models on it for downstream tasks like classification or regression and have indistinguishable performance (and without revealing private information). 
We assess by building a classifier to attempt to distinguish between the synthetic data and a split of data holdout from the training process.  We consider two standard classifiers: XGBoost and logistic regression.  

First, following standard practice, we use XGBoost to build a classifier on the synthetic data, and measure the AUC for classification tasks, and RMSE for regression tasks.   This is referred to as \emph{machine learning efficiency}.  The results in Table \ref{app:tab:mle} on baselines from the \TabSYN paper, as well as SMOTE, and other baselines in our model.  
Note that \TabSYN appears twice in this table (as well as in Tables~\ref{app:tab:c2st} and~\ref{app:tab:dcr}), to include the value recorded in \citet{tabsyn2024} and then our reproduction of the result.   
The Real row in Table~\ref{app:tab:mle} shows the ideal performance, using true training data to train the classifier or regression models. 
The Average Gap column indicates the percentage drop in performance when synthetic data is used to train the classifier or regression models. Table~\ref{app:tab:mle_mean_std} presents the mean and standard deviation of machine learning efficiency (MLE) over 10 runs, indicating that the \TabKDE{} model demonstrates strong robustness with respect to this metric.

Again \TabSYN provides the best error, and \TabKDE is nearly as good -- almost within the natural variation in \TabSYN's performance.  Perhaps surprisingly, the simple method \SMOTE also performs on par with \TabSYN and \TabKDE.


\begin{table}[h!]
\caption{Machine learning efficiency performance comparison across datasets. Unless noted otherwise, scores above the line are taken from \citet{tabsyn2024}; the remaining scores (our methods) are obtained using their data split. To compute the Average Gap, we take the average across all datasets of the relative difference between the performance of a model trained on synthetic data ($s_i$) and the performance of the same model trained on real data ($r_i$);
$\text{Average Gap} = \frac{1}{N} \sum_{i=1}^{N} \left( \frac{\left|s_i - r_i \right|}{r_i} \right) \times 100.$
}
\scriptsize
\centering
\begin{tabular}{lcccccc|c}
\toprule
\textbf{Methods} & \textbf{Adult} & \textbf{Default} & \textbf{Shoppers} & \textbf{Magic} & \textbf{Beijing} & \textbf{News} & \textbf{Average Gap} \\
 & (AUC↑) & (AUC↑) & (AUC↑) & (AUC↑) & (RMSE↓) & (RMSE↓) & (\%) \\
\midrule
Real       & 0.927 & 0.770 & 0.926 & 0.946 & 0.423 & 0.842 & 0\% \\
SMOTE      & 0.899 & 0.741 & 0.911 & 0.934 & 0.593 & 0.897 & 9.39\% \\
CTGAN      & 0.886 & 0.696 & 0.875 & 0.855 & 0.902 & 0.880 & 24.5\% \\
TVAE       & 0.878 & 0.724 & 0.871 & 0.887 & .770 & 1.01 & 20.9\% \\
GOGGLE     & 0.778 & 0.584 & 0.658 & 0.654 & 1.09 & 0.877 & 43.6\% \\
GReaT      & 0.913 & 0.755 & 0.902 & 0.888 & 0.653 & OOM  & 13.3\% \\
STaSy      & 0.906 & 0.752 & 0.914 & 0.934 & 0.656 & 0.871 & 10.9\% \\
CoDi       & 0.871 & 0.525 & 0.865 & 0.932 & 0.818 & 1.21 & 30.5\% \\
TabDDPM     & 0.907 & 0.758 & 0.918 & 0.935 & 0.592 & 4.86 & 9.14\% \\

\TabSYN {\tiny (Our reproduction)}      
                     &  0.911 &  0.760 &  0.913 &  0.942 &  0.663 &  0.820 &  10.70\%\\ 
\midrule
\PGETabsyn           &  0.910 & 0.740  &  0.913 & 0.939  & 0.642  & 2.240  & 37.61\%\\
\CopulaDiff          &  0.901 &  0.763 &  0.912 &  0.939 &  0.667 &   0.921 &  12.18\%\\
\VAESimpleKDE        &  0.896 &  0.733 &  0.874 &  0.912 &  0.777 &  1.037 &  20.70\%\\
\VAETabKDE           &  0.890 &  0.747 &  0.871 &  0.913 &  0.649 &  0.860 &  11.20\%\\
\simpKDE             &  0.901 &  0.730 &  0.913 &  0.931 &  0.756 &  1.167 &  21.39\%\\
\RandomCoresetTabKDE &  0.883 &  0.730 &  0.911 &  0.929 &  0.713 &  0.881 &  14.42\%\\
\CoresetTabKDE       &  0.881 &  0.712 &  0.919 &  0.928 &  0.744 &  0.877 &  15.86\%\\
\TabKDE \tiny{(change the seed)} 
& 0.898  & 0.733  & 0.901  & 0.932  & 0.683  &  0.865  & 12.72\%\\
\TabKDE \tiny{(keep the seed)} 
& 0.890  & 0.742  & 0.913  & 0.924  & 0.695  & 0.878  & 13.32\%\\
\TabKDE              &  0.906 &  0.745 &  0.917 &  0.934 &  0.675 &  0.869 & 11.76\%\\
\bottomrule
\end{tabular}
\label{app:tab:mle}
\end{table}

\begin{table}[ht]
\centering
\caption{Machine learning efficiency comparison across datasets for the \TabKDE{} model, reporting accuracy as mean~$\pm$~standard deviation over 10 runs.}
\label{app:tab:mle_mean_std}
\begin{tabular}{lllllll}
\toprule
{} & \textbf{Adult} & \textbf{Default} & \textbf{Shoppers} & \textbf{Magic} & \textbf{Beijing} & \textbf{News} \\
\midrule
MLE & 0.904\scriptsize{\,$\pm$\,0.003} & 0.744\scriptsize{\,$\pm$\,0.012} & 0.916\scriptsize{\,$\pm$\,0.006} & 0.931\scriptsize{\,$\pm$\,0.004} & 0.678\scriptsize{\,$\pm$\,0.010} & 0.852\scriptsize{\,$\pm$\,0.021} \\
\bottomrule
\end{tabular}
\end{table}

Second, we use a logistic regression classifier.  Follow standard practice we now use this to try to separate the synthetic data from the heldout data.  We quantify this as a classifier two-sample test (C2ST) as provided by SDMetrics; larger values closer to $1$ are better.  
Table \ref{app:tab:c2st} shows both a comparison drawn directly from the \TabSYN paper against a variety of recent baselines; below the line we reproduce results on \TabSYN, show results for \SMOTE, and variants of our method \TabKDE.  

As before, \TabKDE is roughly the same as SMOTE with $0.93$ and only bested by \TabSYN which has about $0.97$.  Other baselines achieve $0.79$ (TabDDPM) or below $0.66$.  


\begin{table}[h!]
\caption{C2ST Scores of generative models on tabular datasets.  
Unless noted otherwise, scores above the line are taken from \citet{tabsyn2024}; the remaining scores (our methods) are obtained using their data split.
}
\scriptsize
\centering
\begin{tabular}{lcccccc|c}
\toprule
\textbf{Method} & \textbf{Adult} & \textbf{Default} & \textbf{Shoppers} & \textbf{Magic} & \textbf{Beijing} & \textbf{News} & \textbf{Average} \\
\midrule
\SMOTE{\tiny (Our reproduction)} & 0.9212& 0.9332& 0.9107& 0.9803 & 0.9972& 0.8633& 0.9334\\
CTGAN   & 0.5949 & 0.4875 & 0.7488 & 0.6728 & 0.7531 & 0.6947 & 0.6586 \\
TVAE    & 0.6315 & 0.6547 & 0.2962 & 0.7706 & 0.8659 & 0.4076 & 0.6044 \\
GOGGLE  & 0.1114 & 0.5163 & 0.1418 & 0.3262 & 0.4779 & 0.0745 & 0.2747 \\
GReaT   & 0.5376 & 0.4710 & 0.4285 & 0.4326 & 0.6893 & ---    & 0.5118 \\
STaSy   & 0.4054 & 0.6814 & 0.5482 & 0.6939 & 0.7922 & 0.5287 & 0.6083 \\
CoDi    & 0.2077 & 0.4595 & 0.2784 & 0.7206 & 0.7177 & 0.0201 & 0.4007 \\
TabDDPM & 0.9755 & 0.9712 & 0.8349 & 0.9998 & 0.9513 & 0.0002 & 0.7888 \\
\TabSYN{\tiny (Our reproduction)}& 0.9949& 0.9804& 0.9699& 0.9893& 0.9268& 0.9584& 0.9699\\
\midrule
\PGETabsyn & 0.9240 & 0.9039  & 0.7679 & 0.9859 & 0.8629 & 0.8060 & 0.8751\\
\CopulaDiff       & 0.8557&	0.9798&	0.8665&	0.9914&	0.9576&	0.9793&  0.9384\\
\VAESimpleKDE  & 0.7199& 0.4082& 0.6736& 0.9665& 0.7392& 0.3782&  0.6476\\
\VAETabKDE & 0.7483& 0.4828& 0.7242& 0.9984& 0.8022& 0.8075& 0.7606\\
\simpKDE & 0.9196& 0.8716& 0.8110& 0.9711& 0.9497& 0.4975& 0.8368\\
\RandomCoresetTabKDE & 0.9215& 0.9570& 0.8757& 0.9921& 0.9503& 0.8901& 0.9311\\
\CoresetTabKDE & 0.8254& 0.873& 0.8462& 0.9864& 0.8924& 0.8643& 0.8813\\
\TabKDE \tiny{(change the seed)} & 0.8772 & 0.8612 & 0.7875 & 0.8272 & 0.9266 & 0.7220 & 0.8336\\
\TabKDE \tiny{(keep the seed)} & 0.9082 & 0.9260 & 0.8789 & 0.9672 &0.9548 & 0.7849 & 0.9033\\
\TabKDE & 0.9219& 0.9579& 0.9161& 1.0000& 0.9514& 0.8819& 0.9382\\
\bottomrule
\end{tabular}
\label{app:tab:c2st}
\end{table}

\subsection{Precision and Recall}
$\alpha$-Precision and $\beta$-Recall are two complementary metrics used to assess the quality of synthetic tabular data, as used in the \TabSYN paper \citep{tabsyn2024}. 
$\alpha$-Precision measures the fidelity of the synthetic data to the real data, indicating how well the synthetic samples preserve fine-grained details and local structures. A higher $\alpha$-Precision score reflects greater similarity to the original data. In contrast, $\beta$-Recall evaluates the extent to which the synthetic data covers the real data distribution, with higher scores indicating broader and more diverse coverage of the feature space. An ideal generative model should balance both metrics—achieving high $\alpha$-Precision while also maintaining strong $\beta$-Recall—thus producing synthetic data that is both accurate and representative of the true distribution. Tables~\ref{app:tab:alpha_precision} and~\ref{app:tab:beta_recal} summarize  $\alpha$-Precision and $\beta$-Recall scores. 

\TabSYN does the best on $\alpha$-precision, but \SMOTE does better on $\beta$-recall.  On $\alpha$-precision \TabKDE (95\%) nearly matches \TabSYN (98.7\%), and is better than any other method (including \SMOTE), except our variant \CopulaDiff which reaches (96\%).  
On $\beta$-precision, \TabKDE (42\%) almost matches \TabSYN (48\%) as is almost as good as any other method with GReaT and STaSy slightly better (43\%); other than \SMOTE (78\%).  But as we discuss next, this likely because \SMOTE generates data mirroring some of the training data.  

\begin{table}[htbp]
\caption{$\alpha$-Precision scores for various methods on the 6 standard data sets.  The last column shows the average score and rank. 
Unless noted otherwise, scores above the line are taken from \citet{tabsyn2024}; the remaining scores (our methods) are obtained using their data split.
}
\centering
\setlength{\tabcolsep}{2.8pt}
\begin{tabular}{lcccccc|cc}
\toprule
\textbf{Method} & \textbf{Adult} & \textbf{Default} & \textbf{Shoppers} & \textbf{Magic} & \textbf{Beijing} & \textbf{News} & \textbf{Average} & \textbf{Ranking}  \\
\midrule
\SMOTE{\tiny (Our reproduction)} & 92.83 & 98.40 & 92.60 &  96.76 & 98.64 & 87.93 & 94.52 & 6   \\
CTGAN & 77.74 & 62.08 & 76.97 & 86.90 & 96.27 & 96.96 & 82.82 & 12\\
TVAE & 98.17 & 85.57 & 58.19 & 86.19 & 97.20 & 86.41 & 85.29 & 10\\
GOGGLE & 50.68 & 68.89 & 86.95 & 90.88 & 88.81 & 86.41 & 78.77 & 15 \\
GReaT & 55.79 & 85.90 & 78.88 & 85.46 & 98.32 & - & 80.87 & 13\\
STaSy & 82.87 & 90.48 & 89.65 & 86.56 & 89.16 & 94.76 & 88.91 & 8 \\
CoDi & 77.58 & 82.38 & 94.95 & 85.01 & 98.13 & 87.15 & 87.03 & 9\\
TabDDPM & 96.36 & 97.59 & 88.55 & 98.59 & 97.93 & 0.00 & 79.83 & 14 \\
\TabSYN{\tiny (Our reproduction)}  & 99.19 &  98.79 &  97.57 & 99.69 & 98.85 &  97.04 & 98.52 & 1 \\
\midrule
\CopulaDiff & 98.09 & 98.99 & 95.43 & 98.43 & 97.33 & 93.98 & 97.04 & 2  \\
\VAESimpleKDE & 88.21 & 80.90 & 82.46& 7.03 & 75.56 & 19.29 & 72.24 & 16 \\
\VAETabKDE & 98.39 & 91.71 & 97.36 & 98.50 & 93.29 & 84.86 & 94.02 & 7   \\
\simpKDE & 98.10 & 93.88 & 98.84 & 90.13 & 96.41 & 29.25 & 84.44 & 11   \\
\RandomCoresetTabKDE & 95.67 & 4.62 & 91.64 & 98.68 & 98.11 & 96.68 & 95.90 & 3  \\
\CoresetTabKDE & 98.01 & 89.44 & 90.27 & 99.12 & 95.70 & 94.96& 94.58 & 5   \\
\TabKDE & 94.46& 94.45 & 92.18 & 98.98 & 97.47 & 97.48 & 95.83 & 4   \\
\bottomrule
\end{tabular}
\label{app:tab:alpha_precision}
\end{table}

\begin{table}[htbp]
\caption{$\beta$-Recall scores for various methods on the 6 standard data sets.  The last column shows the average score and rank.  Unless noted otherwise, scores above the line are taken from \citet{tabsyn2024}; the remaining scores (our methods) are obtained using their data split.
}
\centering
\setlength{\tabcolsep}{2.5pt}
\begin{tabular}{lcccccc|cc}
\toprule
\textbf{Method} & \textbf{Adult} & \textbf{Default} & \textbf{Shoppers} & \textbf{Magic} & \textbf{Beijing} & \textbf{News} & \textbf{Average} & \textbf{Ranking}  \\
\midrule
\SMOTE{\tiny (Our reproduction)} & 76.88 & 76.00 & 77.09 & 82.45 & 79.22 & 80.00 & 78.60 & 1 \\
CTGAN & 30.80 & 18.22 & 31.80 & 11.75 & 34.80 & 24.97 & 25.39 & 16 \\
TVAE & 38.87 & 23.13 & 19.78 & 32.44 & 28.45 & 29.66 & 28.72 & 15 \\
GOGGLE & 8.80 & 14.38 & 9.79 & 9.88 & 19.87 & 2.03 & 10.79 & 17 \\
GReaT & 49.12 & 42.04 & 44.90 & 34.91 & 43.34 & OOM & 42.86 & 6 \\
STaSy & 29.21 & 39.31 & 37.24 & 53.97 & 54.79 & 39.42 & 42.99 & 5 \\
CoDi & 9.20 & 19.94 & 20.82 & 50.56 & 52.19 & 34.40 & 31.19 & 13\\
TabDDPM & 47.05 & 47.83 & 47.79 & 48.46 & 56.92 & 0.00 & 41.34 & 7 \\
\TabSYN {\tiny (Our reproduction)} &  47.53 & 46.74 &  48.85 & 47.64 & 50.56 & 45.10 & 47.74 & 2 \\
\midrule
\CopulaDiff & 41,29 &  46.21 & 43.21 & 46.38 & 51.65 & 43.86 & 45.43 & 3 \\
\VAESimpleKDE & 38.78 & 20.71 & 38.01 & 40.13 & 45.62 & 2.00 & 30.87 & 14 \\
\VAETabKDE & 43.68 & 27.32 & 48.81 & 45.56 & 51.12 & 12.64 & 38.18 & 9 \\
\simpKDE & 45.90 & 36.60 & 43.12 & 44.78 & 52.68 & 3.90 & 37.83 & 11 \\
\RandomCoresetTabKDE & 37.67 & 35.81 & 44.77 & 44.82 & 51.56 & 17.54 & 38.69 &   8 \\
\CoresetTabKDE &  27.46 & 20.86 &  39.40 & 40.35 & 48.30 & 13.01 & 31.56 & 12 \\
\TabKDE & 48.54 &  43.05 &  47.22 & 48.80 &  54.39 &  17.82 & 43.30 & 4 \\
\bottomrule
\end{tabular}
\label{app:tab:beta_recal}
\end{table}

\section{Privacy Preservation}
\label{app:privacy}
Finally, we evaluate how well we can preserve the privacy of the training data in the synthetic data generation process.  We use the distance to closest record (DCR) function in the latent space to evaluate this.  That is for each synthetic data point generated, we both look at the distribution of distances to training or held-out data, and also whether the closest record was from the held-out or training set.  An ideal synthetic distribution would match the distance distribution of the training data to the heldout data, and would be roughly equally likely to be close to the heldout and training data.  

First Table \ref{app:tab:dcr} we calculate the "DCR score" which is the percentage of synthetic data closer to training data than held-out; ideally we would like this to be close to 50\%.  This was proposed by the recent \TabDiff paper \citep{tabdiff2025}, and we reproduce their results in the top half of the table, and show our methods (and \SMOTE and \TabSYN) below the line on an equal split.  We see most diffusion methods (including our \CopulaDiff) can consistently achieve below 52\%.  Our main method \TabKDE obtains an average DCR score of about 58\%, which is servicable.  

On the other hand, \SMOTE has an average DCR score of 95\%.  This indicates that it often nearly replicates the training data.  Its method chooses a training record, finds the $k$ nearest neighbor, and selects a new point in the convex combination of these points, then de-tokenizes back to the tabular format.  Because it works with a one-hot encoding, probably most records map back to the same discrete values as the first record, and it often failures to generate substantially new data, hence leaking the training data.

\begin{table}[ht]
\caption{The DCR score indicates the likelihood that a generated data sample resembles the training set more than the test set. 
A value nearer to 50\% is considered ideal. Ratios recomputed using the true minimum in each column. Values above the line, unless stated otherwise, are taken from \citep{tabdiff2025}; the remaining scores (our methods) are obtained using their data split.} 
\centering
\small
\begin{tabular}{lccccc|c}
\toprule
\textbf{Method} & \textbf{Adult} & \textbf{Default} & \textbf{Shoppers} & \textbf{Beijing} & \textbf{News} & \textbf{Average} \\
\midrule
\SMOTE{\tiny (Our reproduction)} &
91.18 {\tiny (1.83)} &
91.46 {\tiny (1.82)} &
96.76 {\tiny (1.93)} &
100.00 {\tiny (1.99)} &
99.00 {\tiny (1.96)} &
95.68 {\tiny (1.89)} \\
STaSy & 
50.33 {\tiny (1.01)} & 
50.23 {\tiny (1.00)} & 
51.53 {\tiny (1.03)} & 
50.59 {\tiny (1.01)} & 
50.59 {\tiny (1.00)} & 
50.65 {\tiny (1.00)} \\
CoDi & 
49.92 {\tiny (1.00)} & 
51.82 {\tiny (1.03)} & 
51.06 {\tiny (1.02)} & 
50.87 {\tiny (1.01)} & 
50.79 {\tiny (1.00)} & 
50.89 {\tiny (1.01)} \\
TabDDPM & 
51.14 {\tiny (1.02)} & 
52.15 {\tiny (1.04)} & 
63.23 {\tiny (1.26)} & 
80.11 {\tiny (1.59)} & 
79.31 {\tiny (1.57)} & 
65.19 {\tiny (1.29)} \\
\TabSYN{\tiny (Our reproduction)} &
51.33 {\tiny (1.03)} &
51.61 {\tiny (1.03)} &
51.99 {\tiny (1.03)} &
53.20 {\tiny (1.06)} &
50.76 {\tiny (1.00)} &
51.78 {\tiny (1.02)} \\
\TabDiff &
50.10 {\tiny (1.00)} &
51.11 {\tiny (1.02)} &
50.24 {\tiny (1.00)} &
50.50 {\tiny (1.00)} &
51.04 {\tiny (1.01)} &
50.60 {\tiny (1.00)} \\
\midrule
\PGETabsyn & 
50.38 {\tiny (1.01)} & 
51.33 {\tiny (1.01)} & 
51.81 {\tiny (1.03)} & 
50.67 {\tiny (1.01)} & 
50.48 {\tiny (1.00)} & 
50.94 {\tiny (1.01)} \\
\CopulaDiff &
50.34 {\tiny (1.01)} &
50.96 {\tiny (1.01)} &
50.72 {\tiny (1.01)} &
50.29 {\tiny (1.00)} &
53.00 {\tiny (1.05)} &
51.06 {\tiny (1.01)} \\
\VAESimpleKDE &
61.33 {\tiny (1.23)} &
58.08 {\tiny (1.16)} &
58.83 {\tiny (1.17)} &
60.73 {\tiny (1.21)} &
59.00 {\tiny (1.17)} &
59.59 {\tiny (1.18)} \\
\VAETabKDE &
61.28 {\tiny (1.23)} &
57.87 {\tiny (1.15)} &
57.70 {\tiny (1.15)} &
60.42 {\tiny (1.20)} &
58.00 {\tiny (1.15)} &
59.45 {\tiny (1.17)} \\
\simpKDE &
63.32 {\tiny (1.27)} &
63.49 {\tiny (1.26)} &
58.18 {\tiny (1.16)} &
55.42 {\tiny (1.10)} &
56.12 {\tiny (1.11)} &
59.71 {\tiny (1.18)} \\
\RandomCoresetTabKDE &
62.30 {\tiny (1.25)} &
63.09 {\tiny (1.26)} &
58.91 {\tiny (1.17)} &
63.50 {\tiny (1.26)} &
55.59 {\tiny (1.10)} &
60.68 {\tiny (1.20)} \\
\CoresetTabKDE &
52.59 {\tiny (1.05)} &
54.11 {\tiny (1.08)} &
55.04 {\tiny (1.10)} &
51.17 {\tiny (1.02)} &
52.00 {\tiny (1.03)} &
52.98 {\tiny (1.05)} \\
\TabKDE \tiny{(change the seed)} &
64.41 {\tiny (1.29)}     &
68.22 {\tiny (1.36)}     &
59.98 {\tiny (1.19)}     &
55.15 {\tiny (1.10)}     &
58.37 {\tiny (1.16)}     &
61.23 {\tiny (1.21)}     \\
\TabKDE \tiny{(keep the seed)} &
64.24 {\tiny (1.29)} &
67.78 {\tiny (1.35)}     &
60.28 {\tiny (1.20)}     &
54.60 {\tiny (1.09)}     &
56.57 {\tiny (1.12)}     &
60.69 {\tiny (1.20)}     \\
\TabKDE &
62.23 {\tiny (1.25)} &
63.46 {\tiny (1.26)} &
58.80 {\tiny (1.17)} &
54.24 {\tiny (1.08)} &
54.54 {\tiny (1.08)} &
58.55 {\tiny (1.16)} \\
\bottomrule
\end{tabular}
\label{app:tab:dcr}
\end{table}

\begin{figure}[htbp]
    \centering
    \setlength{\tabcolsep}{2pt}
    \renewcommand{\arraystretch}{1.2}
    \begin{tabular}{cccc}
        \textbf{\SMOTE} & \TabSYN & \CoresetTabKDE & \TabKDE \\
        \includegraphics[width=0.25\linewidth]{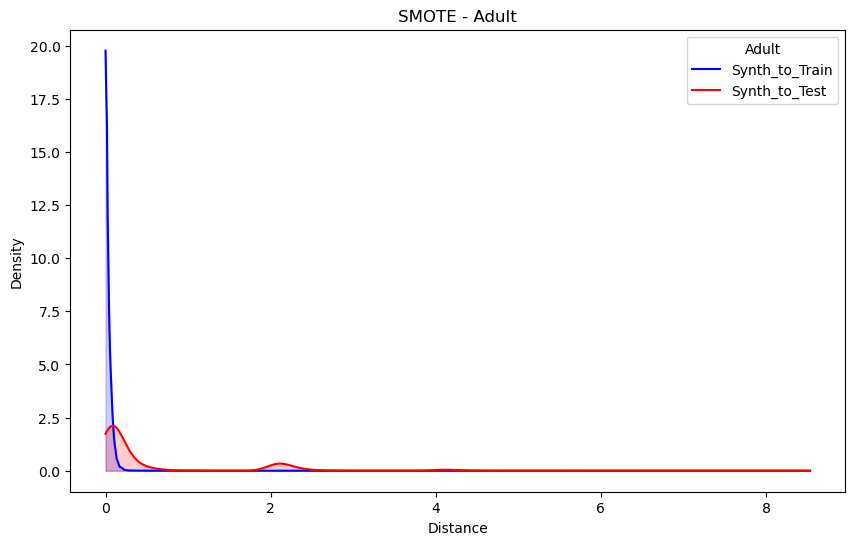} &
        \includegraphics[width=0.25\linewidth]{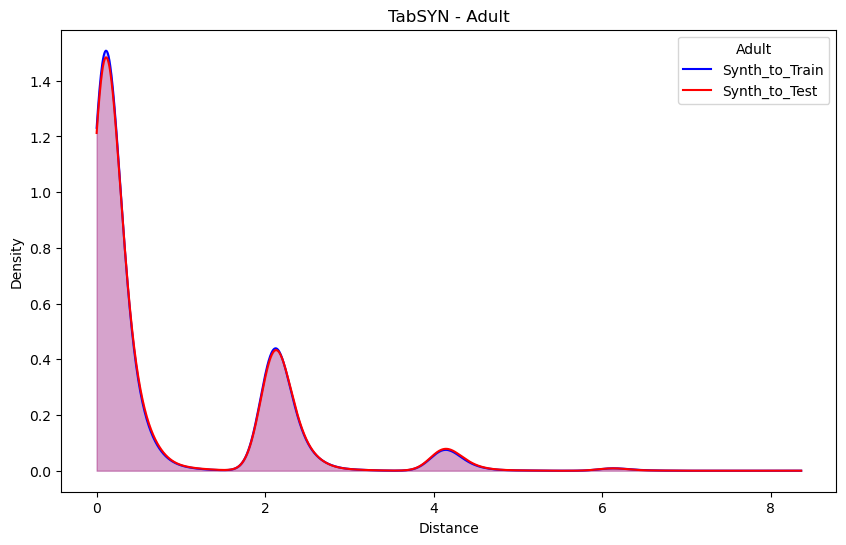} &
        \includegraphics[width=0.25\linewidth]{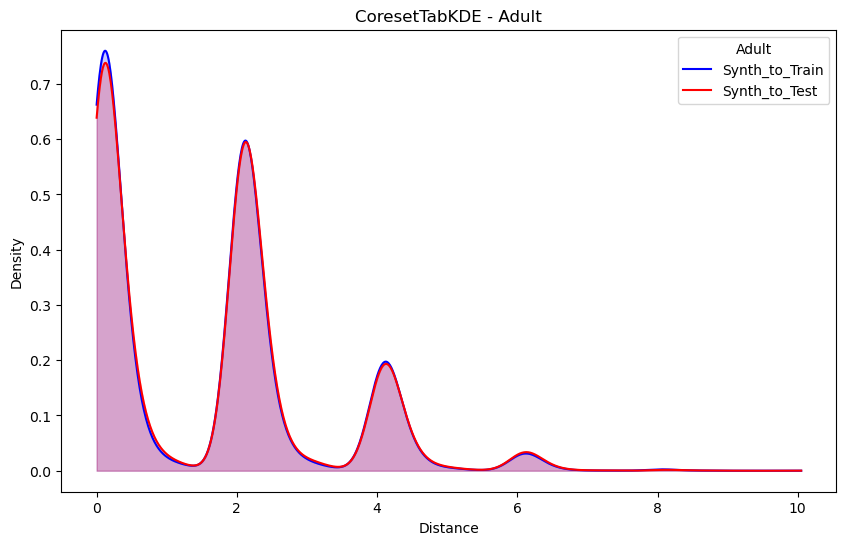} &
        \includegraphics[width=0.25\linewidth]{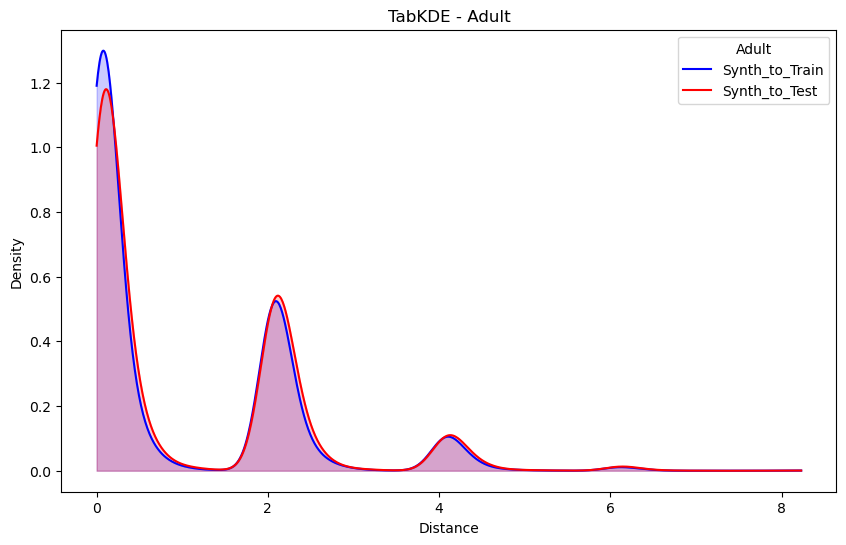} \\
        
        \includegraphics[width=0.25\linewidth]{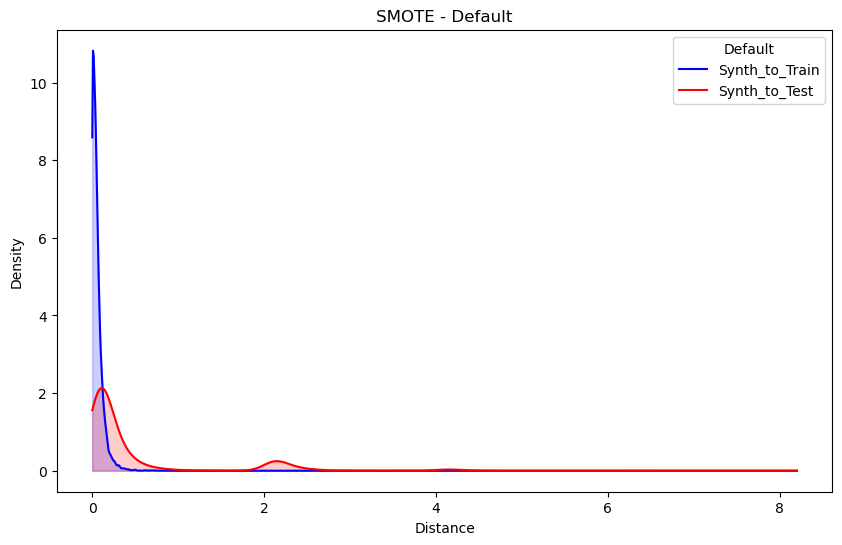} &
        \includegraphics[width=0.25\linewidth]{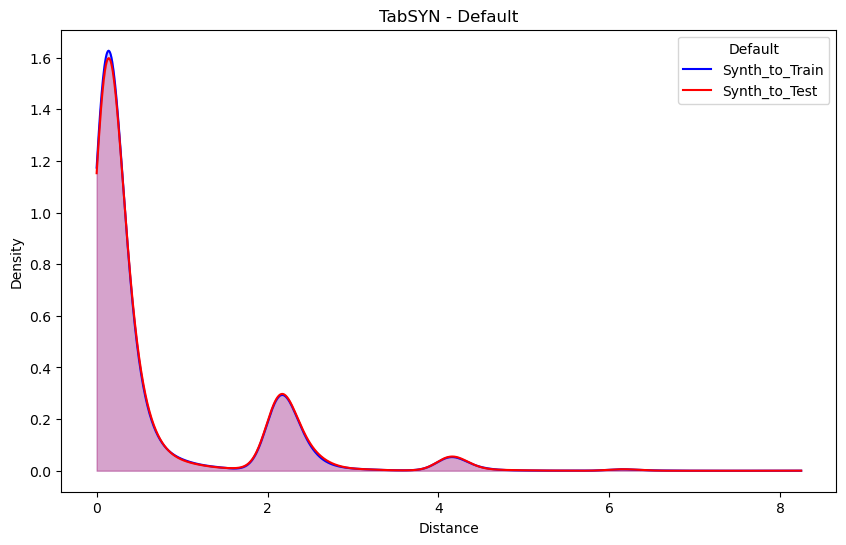} &
        \includegraphics[width=0.25\linewidth]{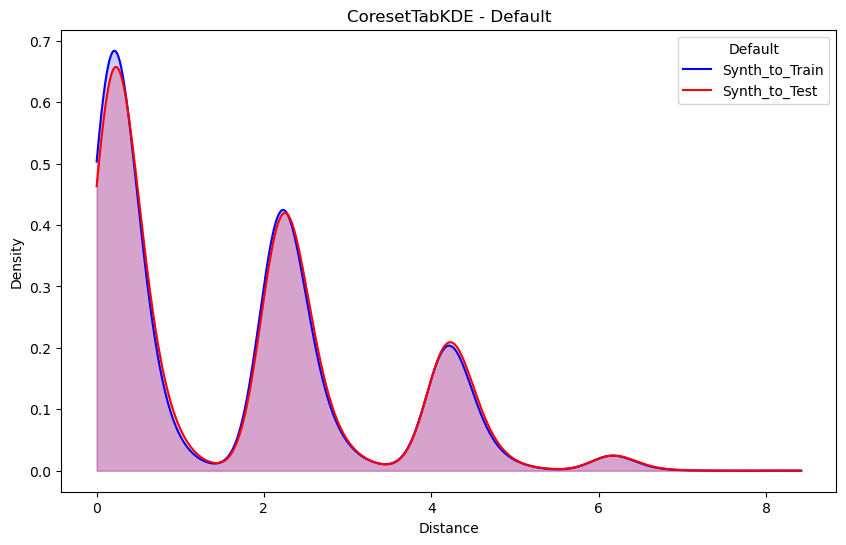} &
        \includegraphics[width=0.25\linewidth]{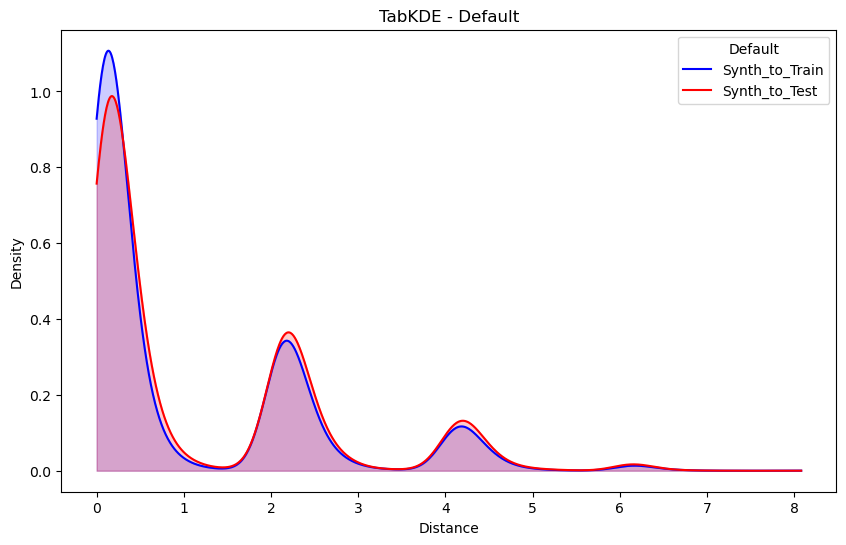} \\
        
        \includegraphics[width=0.25\linewidth]{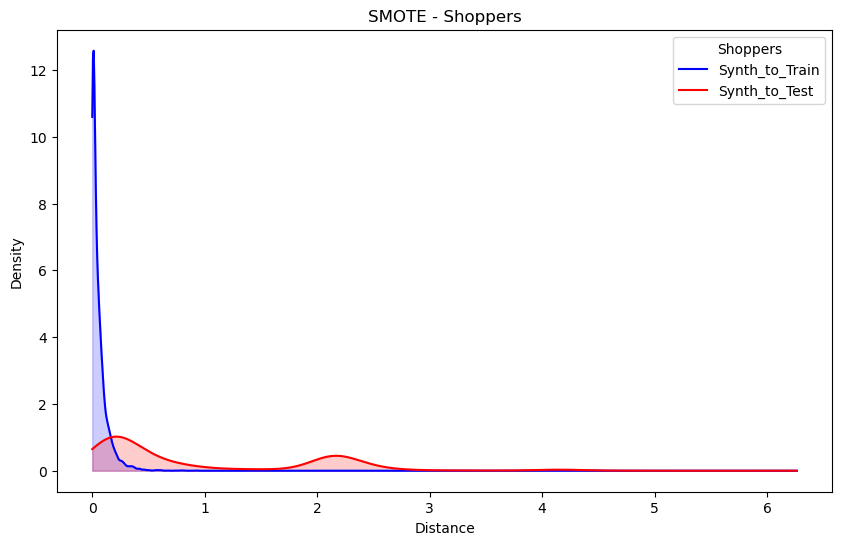} &
        \includegraphics[width=0.25\linewidth]{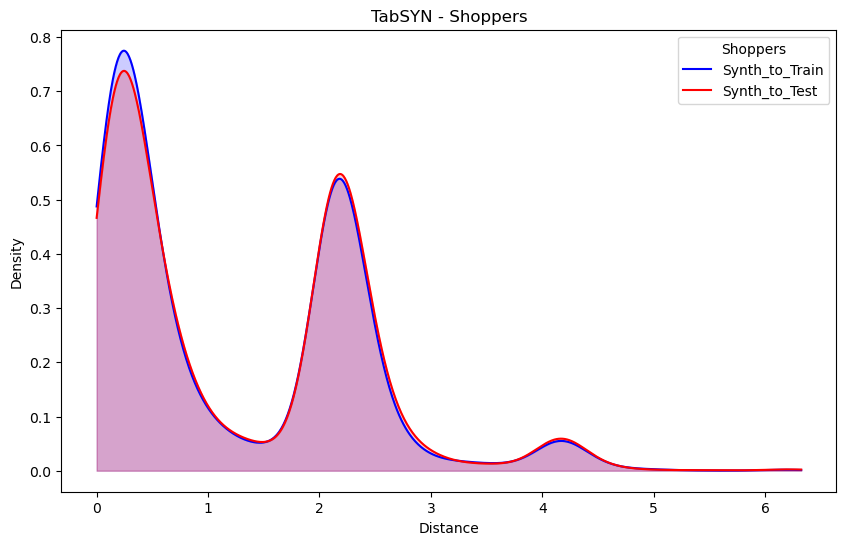} &
        \includegraphics[width=0.25\linewidth]{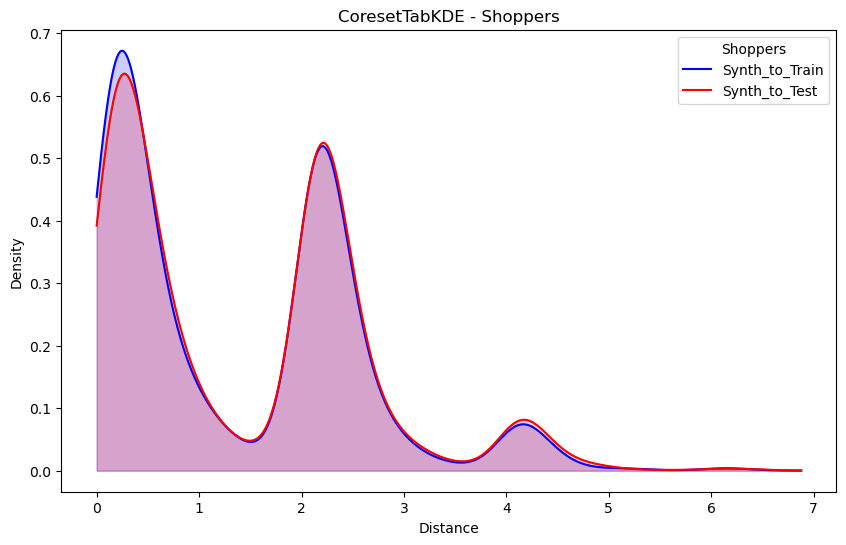} &
        \includegraphics[width=0.25\linewidth]{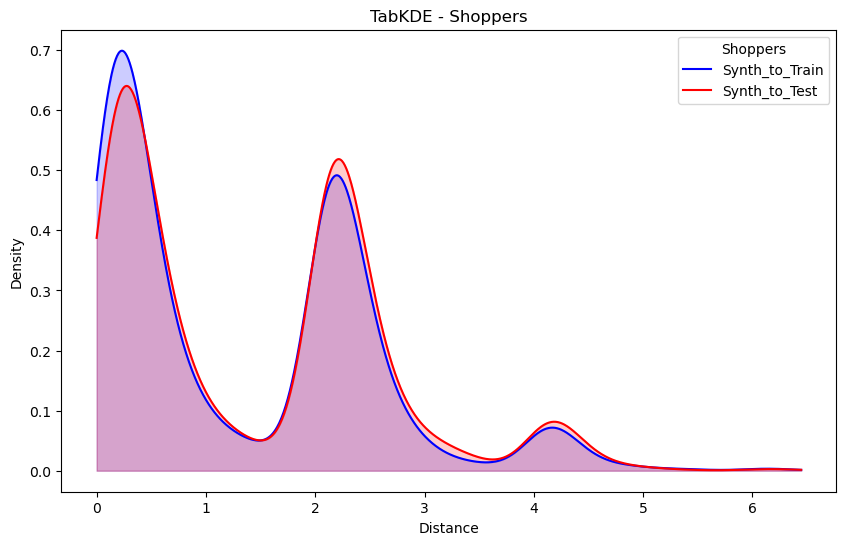} \\
        
        \includegraphics[width=0.25\linewidth]{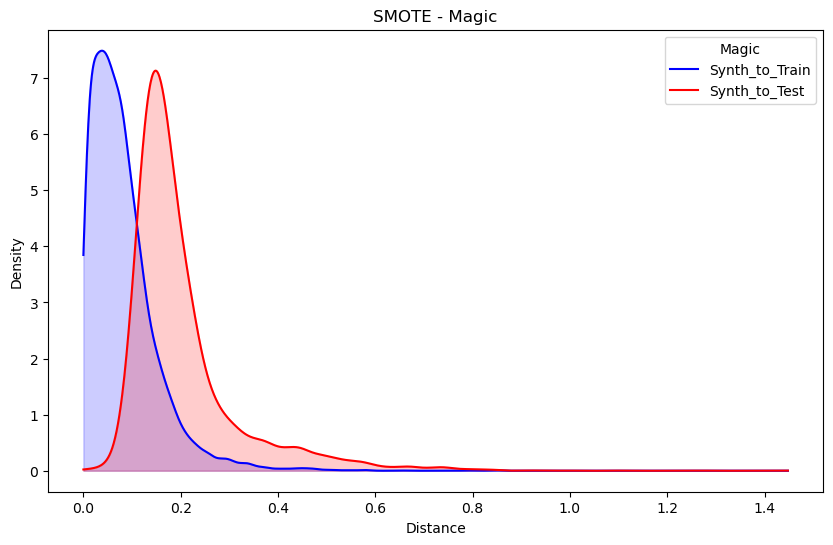} &
        \includegraphics[width=0.25\linewidth]{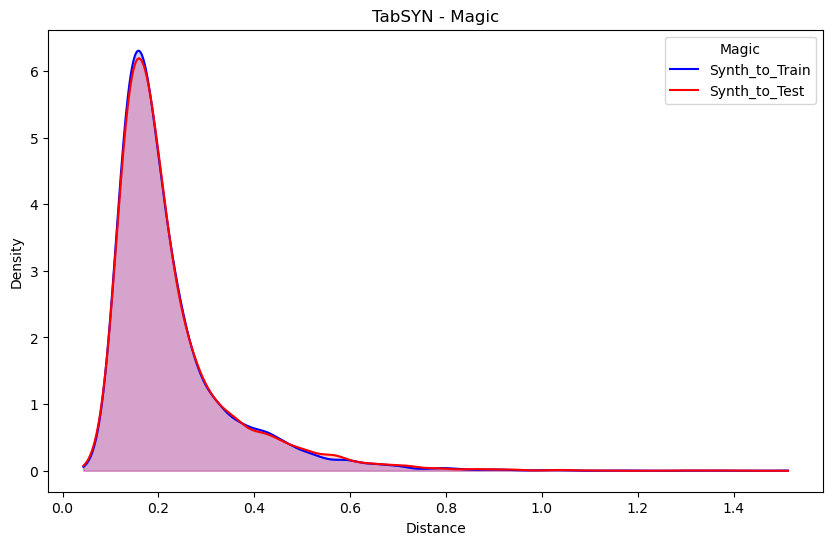} &
        \includegraphics[width=0.25\linewidth]{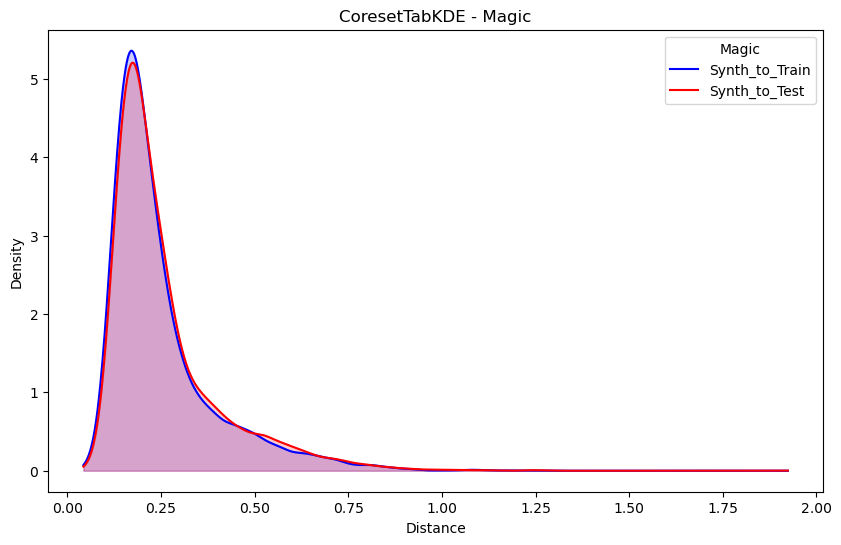} &
        \includegraphics[width=0.25\linewidth]{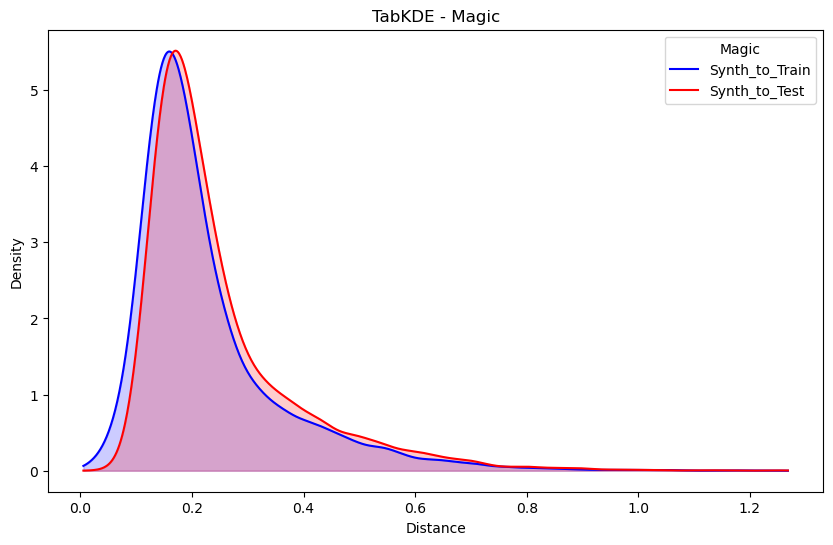} \\
        
        \includegraphics[width=0.25\linewidth]{DCR_plots_/beijing_equal/Smote.png} &
        \includegraphics[width=0.25\linewidth]{DCR_plots_/beijing_equal/Tabsyn.png} &
        \includegraphics[width=0.25\linewidth]{DCR_plots_/beijing_equal/CoresetTabKDE.png} &
        \includegraphics[width=0.25\linewidth]{DCR_plots_/beijing_equal/TabKDE.png} \\
        
        \includegraphics[width=0.25\linewidth]{DCR_plots_/news_equal/Smote.png} &
        \includegraphics[width=0.25\linewidth]{DCR_plots_/news_equal/Tabsyn.png} &
        \includegraphics[width=0.25\linewidth]{DCR_plots_/news_equal/CoresetTabKDE.png} &
        \includegraphics[width=0.25\linewidth]{DCR_plots_/news_equal/TabKDE.png} \\
    \end{tabular}
    \caption{Privacy comparison based on DCR distributions for synthetic to training data (blue) and synthetic to held-out data (red).  Each row is a data set, the columns show results for \SMOTE, \TabSYN, \CoresetTabKDE, and \TabKDE.}
    \label{app:fig:DCR_comparison}
\end{figure}

If the closest record is from the training or heldout data is an imperfect measure of privacy, since there may be heldout data nearly as close.  One way to evaluate this is to consider the distribution of how close the synthetic data to the heldout (red) matches the distribution of the synthetic data to the train (blue).  We show this in Figure \ref{app:fig:DCR_comparison} for \TabKDE, \TabSYN, and \SMOTE for the 6 standard datasets.  
We observe that for \TabKDE and \TabSYN these distributions are multi-modal, but still match almost perfectly for each data set.  On the other hand \SMOTE has a very different distribution, and the synthetic to train (blue) is always much smaller (almost always close to 0 for Adult, Default, Shopping, and Beijing), indicating it is too closely just reproducing the training data.

We note a practical limitation in computing DCR for datasets with high-cardinality categorical features. In the IBM dataset, $13$ categorical columns each have thousands of distinct values; under standard one-hot encoding, this yields $37,722$ dimensions. At this scale, both memory usage and computational cost become prohibitive and exceeded our available hardware, leading to out-of-memory failures during evaluation (even with aggressive use of FAISS). While dimensionality reduction could mitigate the resource burden, it would also alter the meaning of DCR by assessing proximity in a compressed space rather than the original feature space. Accordingly, we report that DCR evaluation was not feasible for our highest-dimensional dataset under our computational constraints.

\section{Coresets for Generative Tabular Data Modeling}\label{app:coreset}
A \emph{coreset}~\citep{Phi16} is a compact, weighted set of points that provides a close approximation to the full dataset for a specific downstream task. In the context of Kernel Density Estimation (KDE), a coreset serves to approximate the full KDE using significantly fewer, yet strategically chosen, representative points.
A \emph{weak coreset} is a coreset that the set is not necessarily a subset of the original point set.  In this setting, the ``weak'' aspect will turn out to be strategically advantageous.  

Our proposed \textsc{TabKDE} framework employs the full KDE to generate samples from the Copula latent representation \( Z \subset [0,1]^{n \times d} \) of the tabular data \( \mathcal{T} \). The full KDE over \( Z \) is given by:
\[
f_Z(z) = \frac{1}{|Z|} \sum_{z_i \in Z} K \left( \frac{z - z_i}{h} \right),
\]
which we here consider it as the ground-truth likelihood function over \( z \in [0,1]^d \).
To approximate \( f_Z(\cdot) \) using a weak coreset, we define a parameterized KDE \( \tilde{f}_{\Theta}(\cdot) \) based on a small set of learnable coreset (support) points \( Q = \{q_1, \ldots, q_m\} \) and their corresponding non-negative weights \( W = \{\omega_1, \ldots, \omega_m\} \), constrained such that \( \sum_{i=1}^{m} \omega_i = 1 \), where \( \Theta = \{Q, W\} \). The approximated density function is:
\[
\tilde{f}_{\Theta}(z) = \sum_{i=1}^{m} \omega_i K \left( \frac{z - q_i}{h} \right),
\]
with \( m \ll n \).
It is known \citep{PT2020} that a sample $Q \sim Z$ of $m = O((1/\eps^2)\log(1/\delta))$ points already ensures a strong $L_\infty$ coresets approximation that $\|f_Z - f_Q\|_\infty \leq \eps$ with probability at least $1-\delta$.  Note that for a fixed kernel $K$, this bound is independent of the dimension $d$.  

We note that this $f_Z$ is not precisely our sampling distribution, since using \textsc{SampleKDE-Iterative} we do a coordinate-wise rejection of points for violating the $[0,1]^d$ boundary condition.  Hence, this accuracy guarantee does not exactly hold for our \TabKDE algorithm.  We do however consider a more common form of rejection sampling, where if for a base point $z \in Z$ we generate a sample outside of $[0,1]^d$, then resample from $K(z,\cdot)$.  This means each generating kernel $K(z,\cdot)$ is the intersection of a centrally symmetric kernel $K$ and a box $[0,1]^d$, each with super-level sets with VC-dimension $O(d)$, so their intersection is also VC-dimension $O(d)$~\cite{anthony2009neural}.  This means we can employ another coreset analysis of KDEs by \cite{JKPV2011} which for such a setting ensures that samples of size $O((1/\eps)(d+\log(1/\delta))$ is an $\eps$-approximate coreset with probability at least $1-\delta$.  However, we found (see Section \ref{app:correction-stats}  ) that this rejection sampling method is less effective in terms of accuracy, so do not install it as the sampling procedure for \TabKDE.  

Also, while we use this random sample as a starting point, we seek to improve it with a weak coreset.  
In particular, the parameters \( \Theta \) are optimized by minimizing the expected squared \( L_2 \) deviation between the full KDE and its coreset approximation, evaluated over samples drawn from the uniform distribution on \( [0,1]^d \):
\[
\mathbb{E}_{z \sim \mathsf{Unif}([0,1]^d)} \left[ \left( \tilde{f}_{\Theta}(z) - f_Z(z) \right)^2 \right].
\]
By optimizing the positions and weights of the coreset points to minimize the discrepancy between the coreset-based KDE and the full KDE, the method preserves key distributional features,  such as modes, spread, and overall shape.  Moreover, because we use a weak coreset, we are not replicating the training data, and this minimizes the risk of overfitting to the data or leaking its private attributes.

This objective can be optimized via stochastic gradient descent (SGD) with \textsc{TrainCoresetKDE} (Algorithm \ref{app:alg:TrainCoresetKDE}).

\begin{algorithm}[H]
\caption{\textsc{TrainCoresetKDE}$(Z, T)$}
\label{app:alg:TrainCoresetKDE}
\begin{algorithmic}[1]
\State Define $f_Z(z) = \frac{1}{|Z|} \sum_{z_i \in Z} K \left( \frac{z - z_i}{h} \right)$
\State Initialize randomly the coreset points $Q = \{q_1, \dots, q_m\} \subset Z$  
\State Initialize the weights $W = \{\omega_1, \dots, \omega_m\}$ with  $\omega_i = 1/m$

\State Define $\tilde{f}_{\Theta}(z) = \sum_{i=1}^m \omega_i K\left( \frac{z - q_i}{h} \right)$

\State {\bf For} $t = 1$ to $T$:
    \State \hspace{0.5cm} Sample $z \sim \mathsf{Unif}([0,1]^d)$
    \State \hspace{0.5cm} Compute loss $\mathcal{L}(z) = \left(\tilde{f}_{\Theta}(z)-f(z)\right)^2$
    \State \hspace{0.5cm} Update $\Theta$ via gradient descent to minimize $\mathcal{L}(z)$
\State \Return $\Theta = \{Q, W\}$
\end{algorithmic}
\end{algorithm}

 We employ the Gaussian kernel as $K(v) = \exp(-v^2)$ in our formulation.  
Once trained, the learned weak coreset \( \{(q_i, \omega_i)\}_{i=1}^m \) replaces the full KDE sampling step in Algorithm~\ref{app:alg:SampleKDE}. This modification constitutes the only difference between \textsc{TabKDE} and \textsc{CoresetTabKDE}, and is detailed in the sampling procedure below.

\begin{algorithm}[H]
\caption{$\textsc{SampleCoresetKDE-iterative}(Q, W)$}
\label{app:alg:SampleKDE_coreset}
\begin{algorithmic}[1]
\State $Q, W = \textsc{TrainCoresetKDE}(Z, T)$
\State Sample \( q_i \in Q \) with probability \( \omega_i \in W\)
\State Sample radius \( r > 0 \) from \( f \)
\State Sample \( v \sim \mathcal{N}(0, \Sigma) \), set \( u = \frac{v}{\|v\|} \)
\State \( z' \gets z_i + r \cdot u \)
\State {\bf While} \(\{j : z'_j \notin [0,1] \} \neq \varnothing \): 
    \State \hspace{.5cm} \( J \gets \{j : z'_j \notin [0,1] \} \)
    \State \hspace{.5cm} Sample \( v' \sim \mathcal{N}(0, \Sigma) \), set \( w = \frac{v'}{\|v'\|} \)
    \State \hspace{.5cm} \( s \gets \frac{\| (u_k)_{k \in J} \|}{\| (w_k)_{k \in J} \|} \)
    \State \hspace{.5cm} \( u_j \gets s \cdot w_j \) for each \( j \in J \) 
    \State \hspace{.5cm} \( z' \gets z_i + r \cdot u \)
\State \Return \( z' \)
\end{algorithmic}
\end{algorithm}

We may also define a \textsc{RandomCoresetTabKDE} variant, in which the optimization step in Alg.~\ref{app:alg:TrainCoresetKDE} (Step 1) is omitted.   Instead, a subset \( Q \subset Z \) of size \( m \) is sampled uniformly at random; then we simply invoke \textsc{SampleKDE-iterative}(Q) (Alg \ref{app:alg:SampleKDE}) with $Q$ instead of $Z$.  
As noted above, this simple approach of taking a random sample $Q \sim Z$ has strong $L_\infty$ approximation guarantees on how well it approximates the KDE of $Z$; and this does not depend on either the dimension $d$ or the size $n$ of $Z$ -- it only depends on the size $|Q| = m$ of the sample.

\subsection{Empirical Evaluation of Coreset Methods}
\label{app:sec:coreset-eval}
\begin{figure}[h]
    \centering
    \begin{minipage}{0.48\linewidth}
        \centering
        \includegraphics[width=\linewidth]{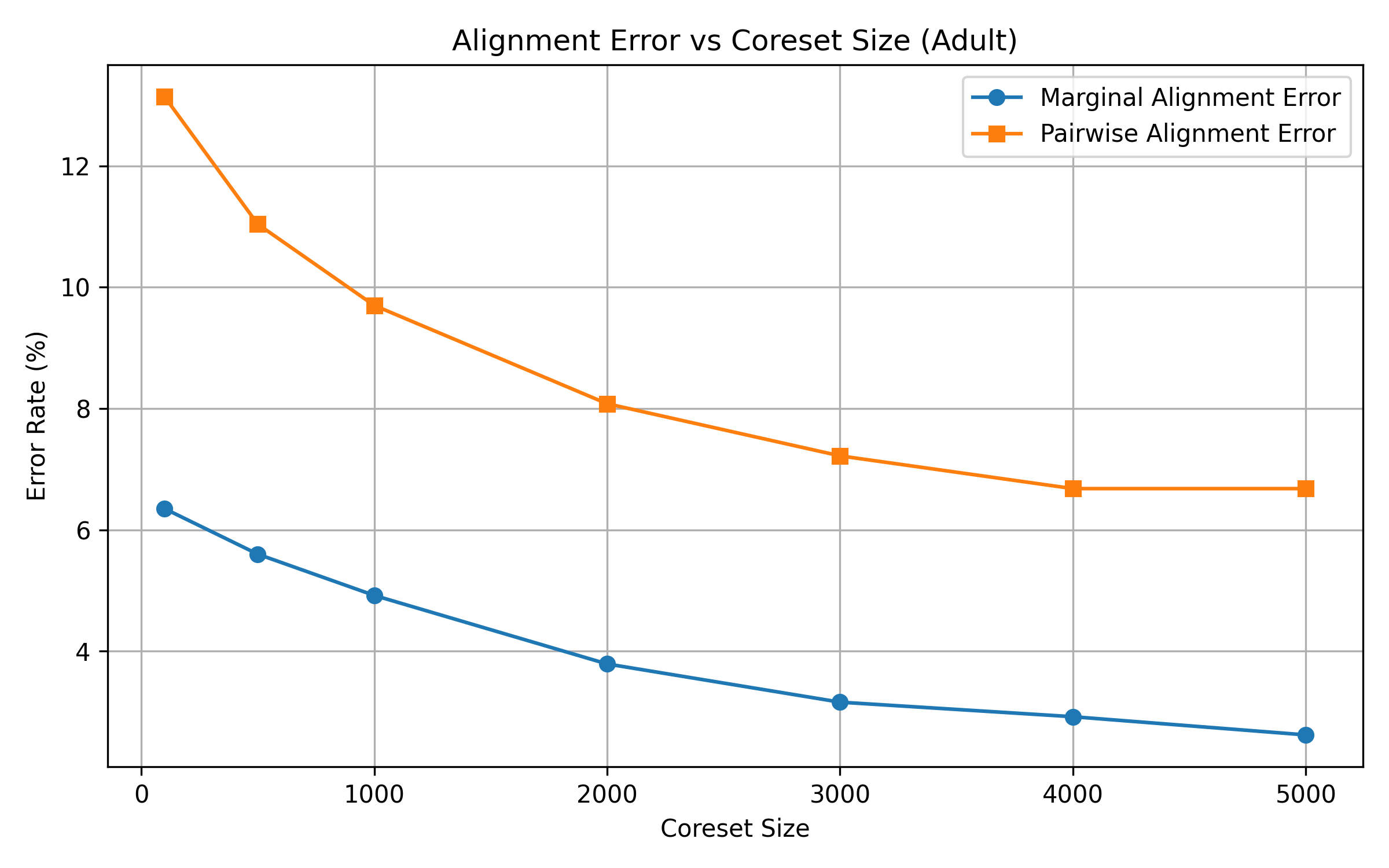}
    \end{minipage}
    \hfill
    \begin{minipage}{0.48\linewidth}
        \centering
        \includegraphics[width=\linewidth]{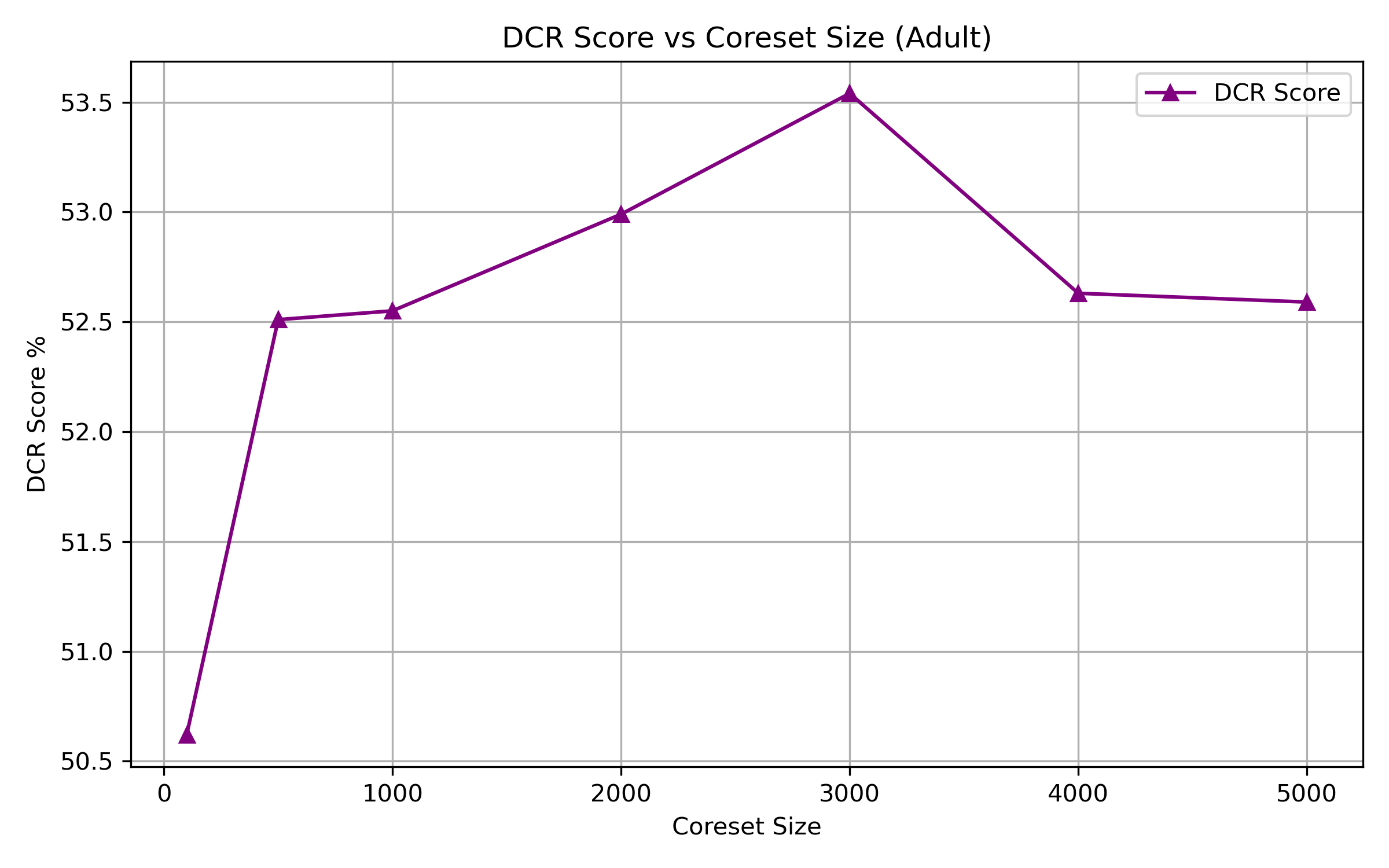}
    \end{minipage}
    \caption{Left: Marginal distribution alignment error (\%) and pairwise correlation alignment error (\%) as coreset size increases. Right: DCR score as coreset size increases.}
        \label{app:fig:coreset-size}
\end{figure}

In this section, we empirically examine the advantages and limitations of the coreset approaches \CoresetTabKDE and \RandomCoresetTabKDE introduced in Subsection~\ref{app:coreset}. Across all the data sets,
we set $m=5{,}000$; as shown in Figure \ref{app:fig:coreset-size}, the marginal and pairwise correlation alignment scores look to plateau around that value. 
By ablation study, we set $h=0.2$ for data sets Adult,  Default, Shoppers,
Magic, 
Beijing, and News, we set $h= 0.2, 0.4, 0.2, 0.2, 0.2, 0.5$, respectively.  
We consider bandwidths $h \in \{0.1, 0.2, \ldots, 1.0\}$ at $0.1$ intervals, and examine the loss function of the \textsc{TrainCoresetKDE} procedure run for $T=30$ epochs.
We select the bandwidth $h$ where the corresponding loss has the steepest descent towards zero.  
High values of $h$ result in negligible loss updates, while overly small values lead to premature convergence at suboptimal loss values.  
If multiple consecutive bandwidths yield similar behavior, we take the smaller as the chosen value.

\begin{figure}[h]
    \centering
    \begin{minipage}{0.48\linewidth}
        \centering
        \includegraphics[width=\linewidth]{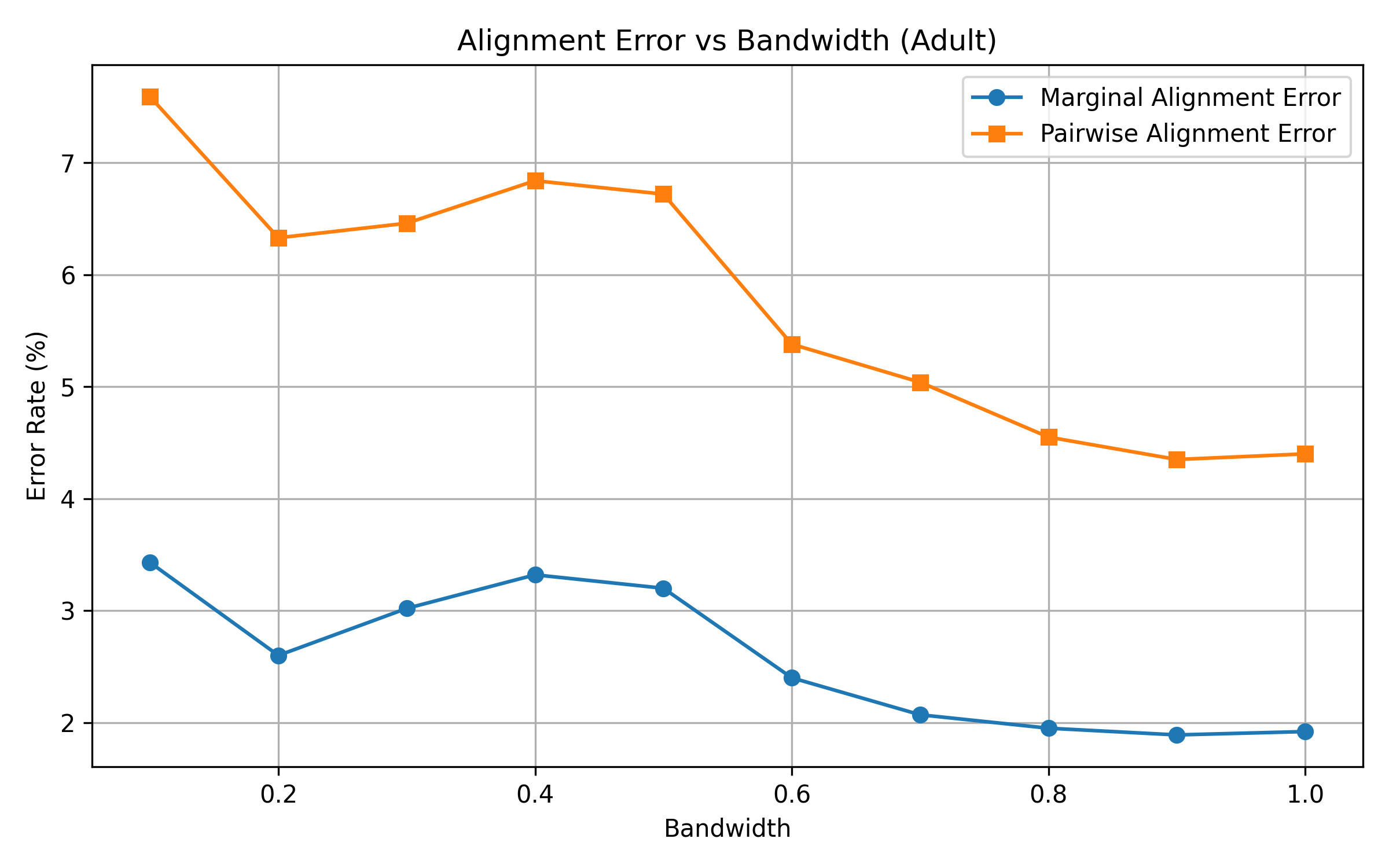}
    \end{minipage}
    \hfill
    \begin{minipage}{0.48\linewidth}
        \centering
        \includegraphics[width=\linewidth]{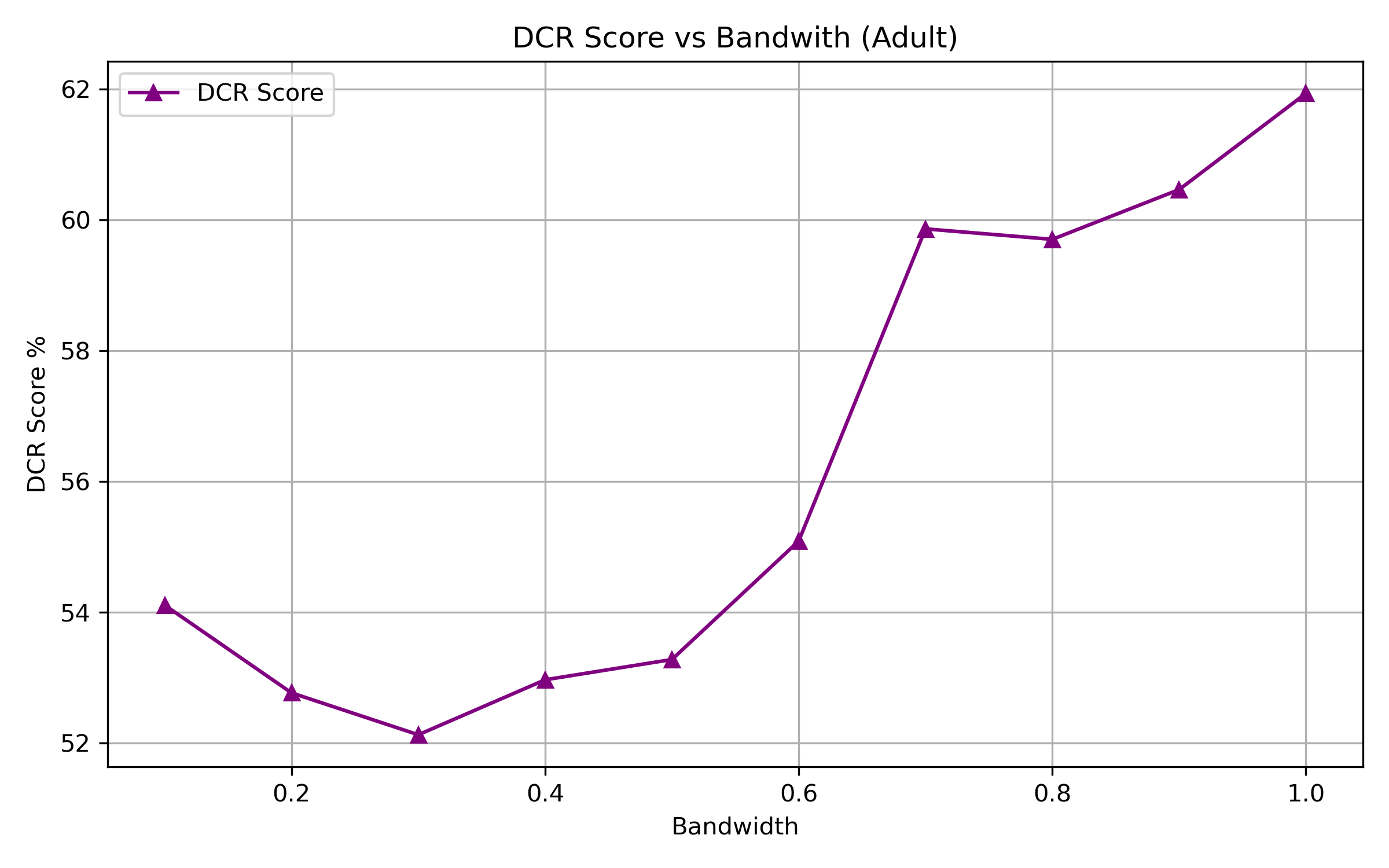}
    \end{minipage}
    \caption{Left: Marginal distribution alignment error (\%) and pairwise correlation alignment error (\%) as bandwidth increases. Right: DCR score as bandwidth increases.}
        \label{app:fig:coreset-h}
\end{figure}

While the bandwidth selection is chosen based only on the loss function in the training data, we also validate our selection on the test data.   As shown on the Adult data set with $m=5000$ in Figure \ref{app:fig:coreset-h}, the alignment error (both marginal and pairwise correlation; higher better) is fairly stable across choices of $h$ in $0.1$ to $1.0$, but has a local peak around $h=0.2$.  
Moreover, the DCR score has a more noticeable drop for $h \leq 0.6$; hence $h=0.2$ is confirmed as a good choice.  
While this KDE is built in $[0,1]^d$ copula space for all data sets, the dimension changes from $6$ (for Adult) to $46$ (for News), and the higher dimensional setting has more room to spread points out, and prefers a larger bandwidth.  

The accuracy scores are reported in the numerous tables above, and one can observe this method achieves accuracy nearly as good as \TabKDE, but often a bit worse; see for instance Tables~\ref{app:tab:shape} and~\ref{app:tab:trend}.  
The scalability of \RandomCoresetTabKDE is also about the same on the data sets we consider as \TabKDE (which is already very efficient).  However, now we require much less space to store the model; and the based on KDE-coreset results~\citep{PT2020,JKPV2011}, the accuracy for a fixed size coreset should not decrease as the training data grows.  However, the \CoresetTabKDE requires some optimization training time: approximately 55 minutes (on the laptop CPU) average across the five datasets: Adult, Default, Shoppers, Magic, and Beijing.  

What is more interesting is the effect on privacy in using \CoresetTabKDE.  As shown in Table~\ref{app:tab:dcr}, \CoresetTabKDE offers a notable improvement in privacy (under the unstable DCR score), with an average DCR score of about 53\%.   This is because the coreset no longer precisely stores the training data; rather it is storing a distribution of data points $Q$ which have a similar KDE as does the training data $Z$.  Hence, when it generates synthetic data, it is not using some training data point $z \in Z$ as a starting point.

\section{Ablation Study on Variants of TabKDE}
\label{app:app:ablation}

In this section we do a small ablation study to investigate the differences among variants of \TabKDE.  Most data is presented in earlier tables in the main paper.  The main take-a-ways are as follows.  

The methods that use Diffusion or VAE are significantly slower, and less scalable.  As discussed above some of this can be ameliorated by avoiding one-hot encoding, but still the difference in runtime is very large on the IBM data set.  
Second, none of the methods that take elements from \TabSYN directly match it in terms of accuracy, although \CopulaDiff sometimes does nearly as well, and on average, they all also do not outperform \TabKDE.  
Third, on privacy, \CopulaDiff achieves an average DCR score of 51\%, so it does quite well, improving upon \TabKDE and \CoresetTabKDE.

We next provide analysis comparing to SimpleKDE.  It should already have been apparent from the accuracy evaluation where it performs a bit worse than \TabKDE that it is not the preferred method.  But next we provide a more in depth discussion on the marginals, where the issue becomes even more clear why the iterative element is required.

\subsection{Motivation for TabKDE: Boundary Control Challenges in SimpleKDE}\label{app:subsec:whyTabKDE}

\begin{figure}[htp]
    \centering
    \begin{tabular}{cccc}
        \includegraphics[width=0.225\linewidth]{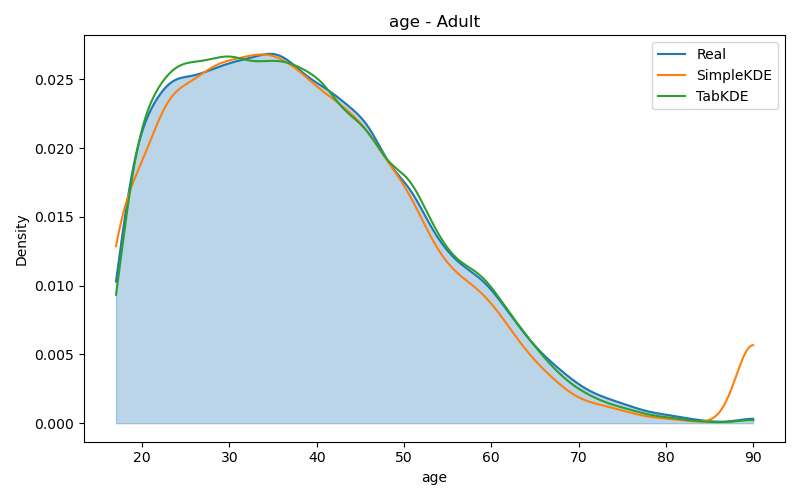} &
        \includegraphics[width=0.225\linewidth]{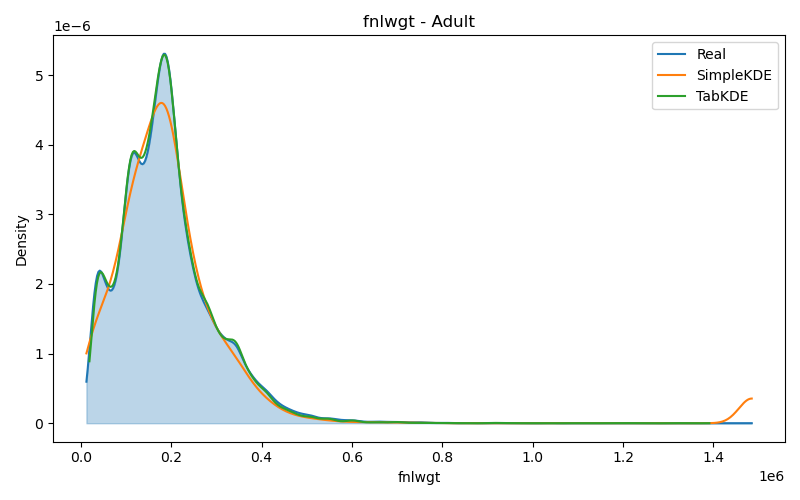} &
        \includegraphics[width=0.225\linewidth]{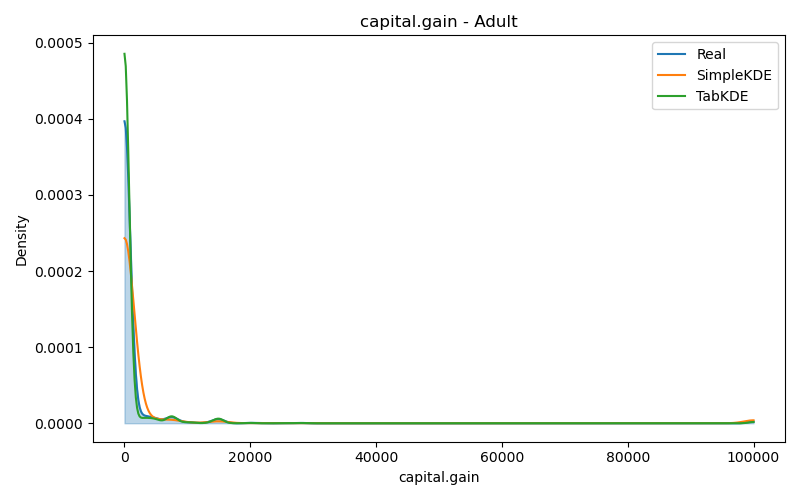} &
        \includegraphics[width=0.225\linewidth]{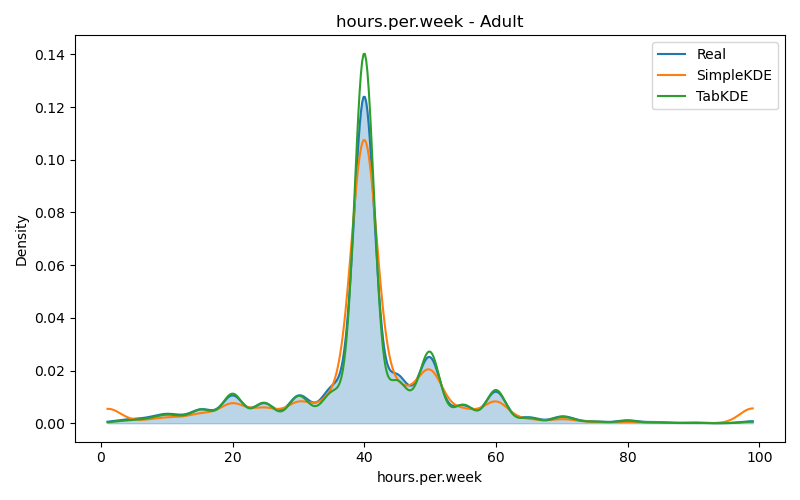} \\
        
        \includegraphics[width=0.225\linewidth]{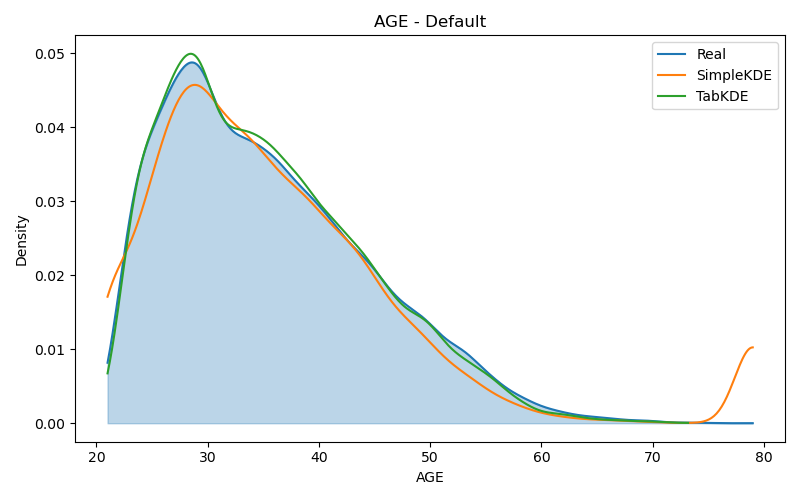} &
        \includegraphics[width=0.225\linewidth]{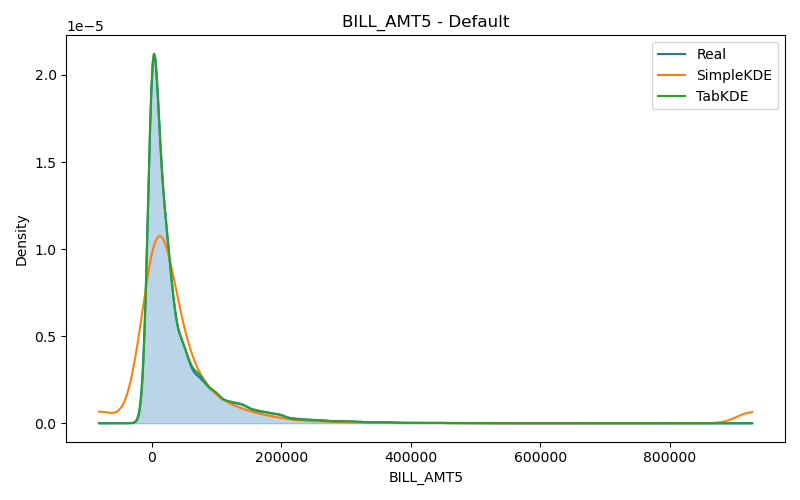} &
        \includegraphics[width=0.225\linewidth]{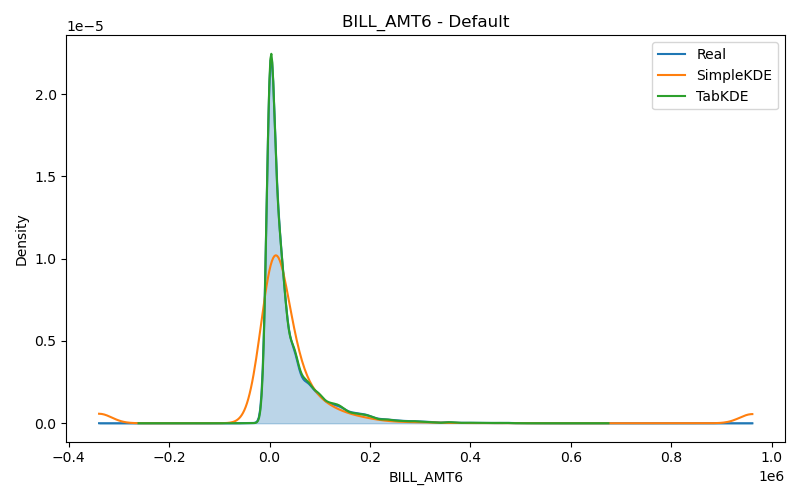} &
        \includegraphics[width=0.225\linewidth]{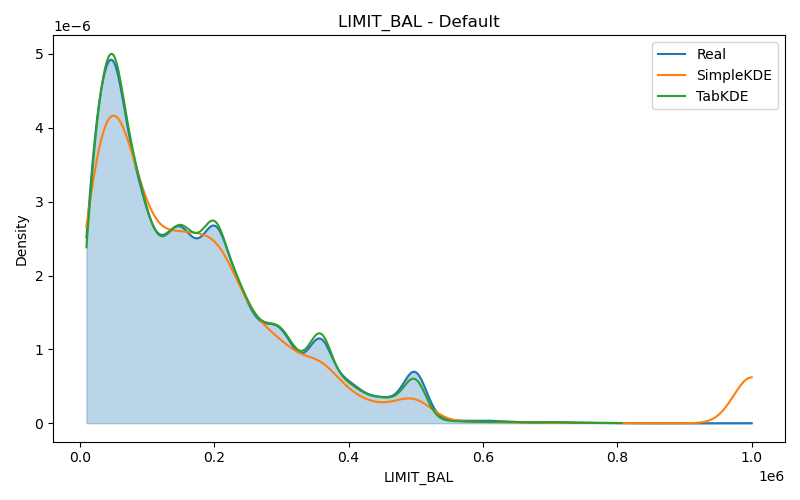} \\
        
        \includegraphics[width=0.225\linewidth]{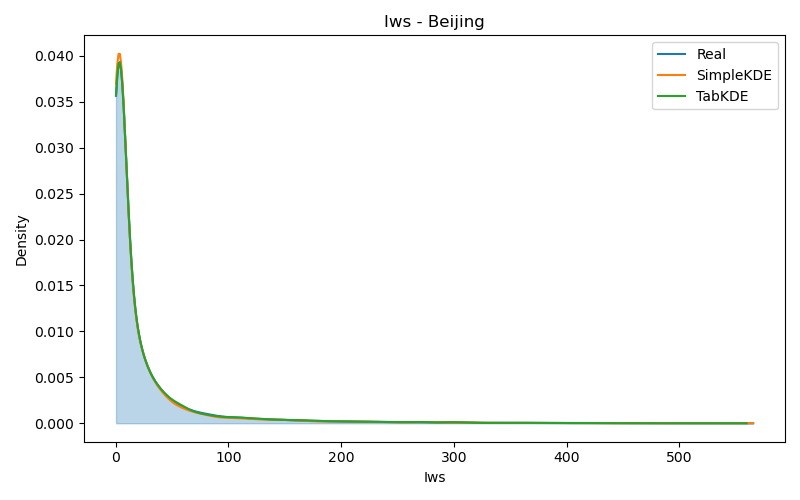} &
        \includegraphics[width=0.225\linewidth]{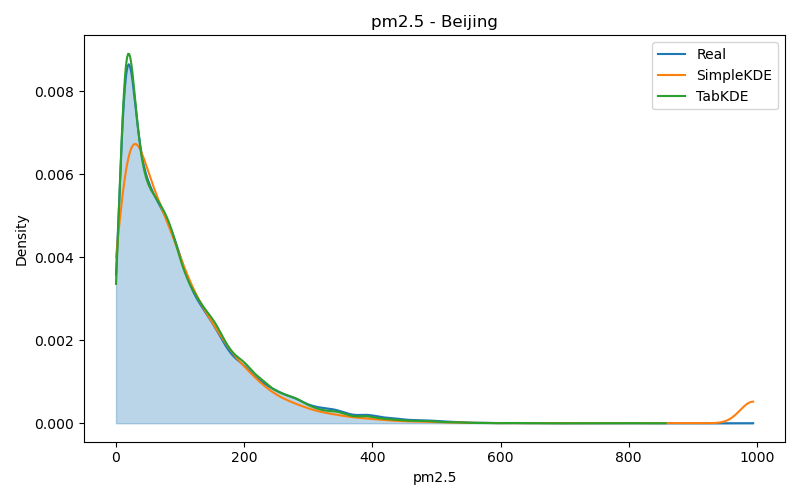} &
        \includegraphics[width=0.225\linewidth]{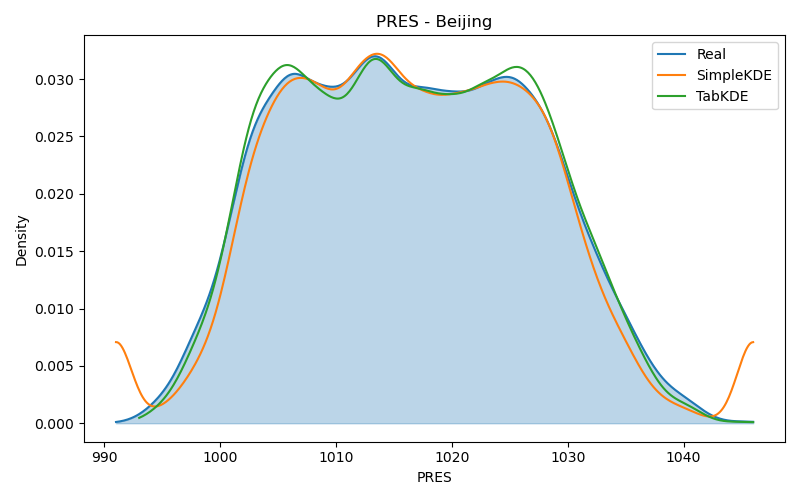} &
        \includegraphics[width=0.225\linewidth]{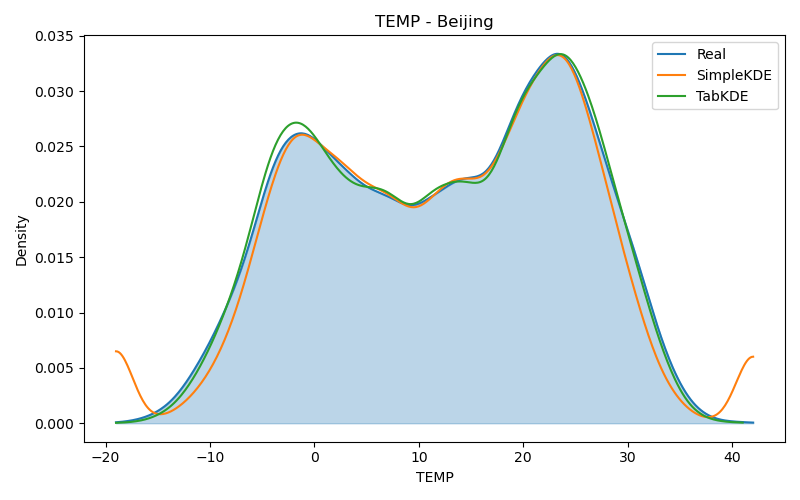} \\
        
        \includegraphics[width=0.225\linewidth]{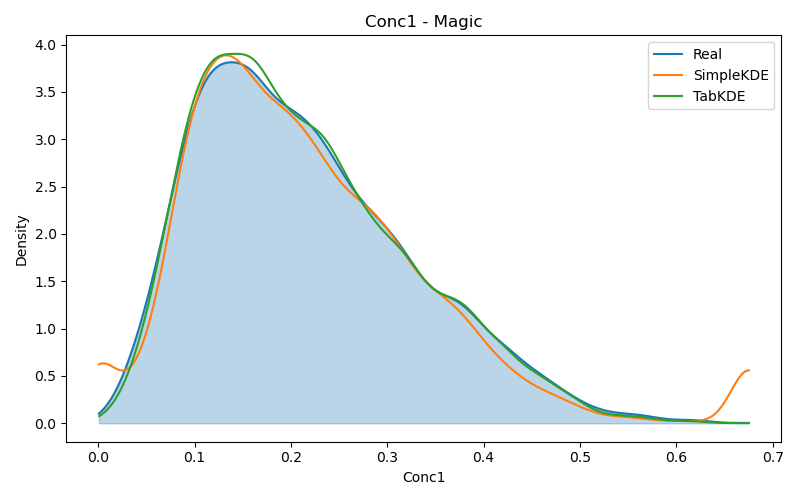} &
        \includegraphics[width=0.225\linewidth]{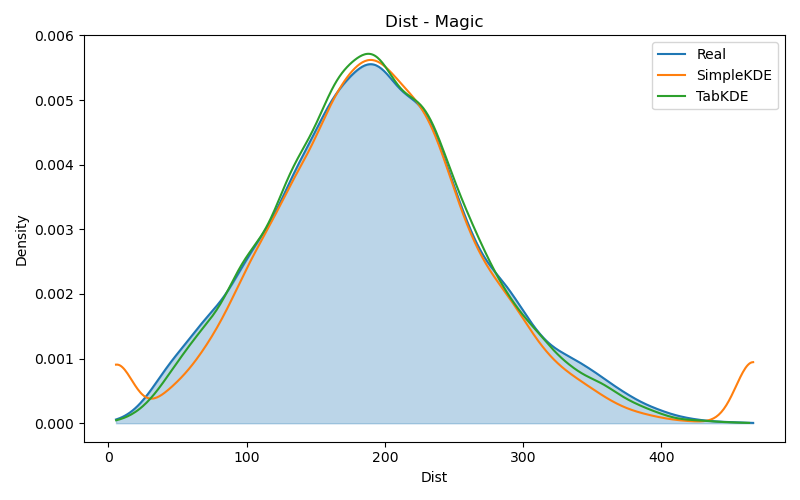} &
        \includegraphics[width=0.225\linewidth]{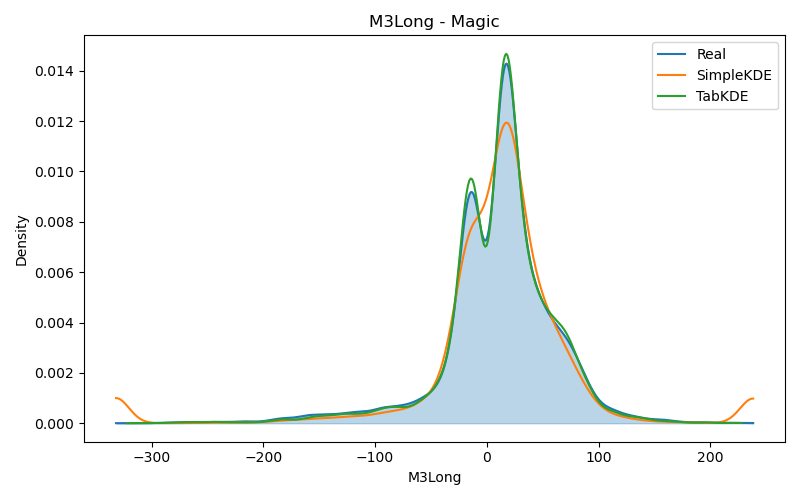} &
        \includegraphics[width=0.225\linewidth]{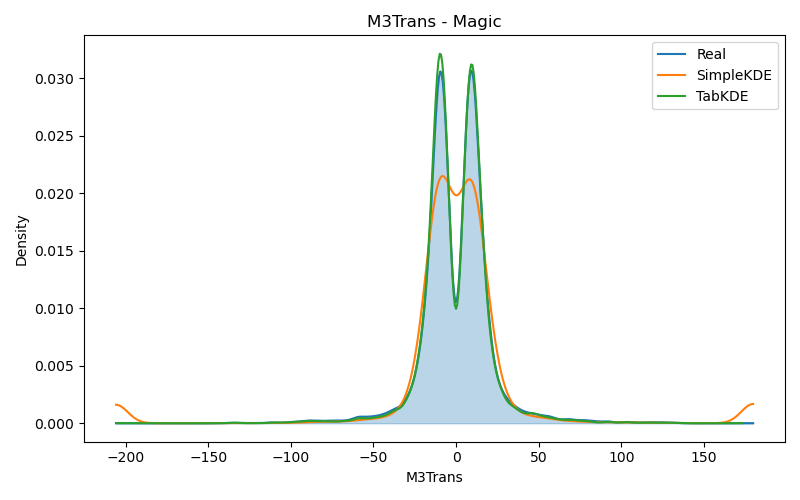} \\
        
        \includegraphics[width=0.225\linewidth]{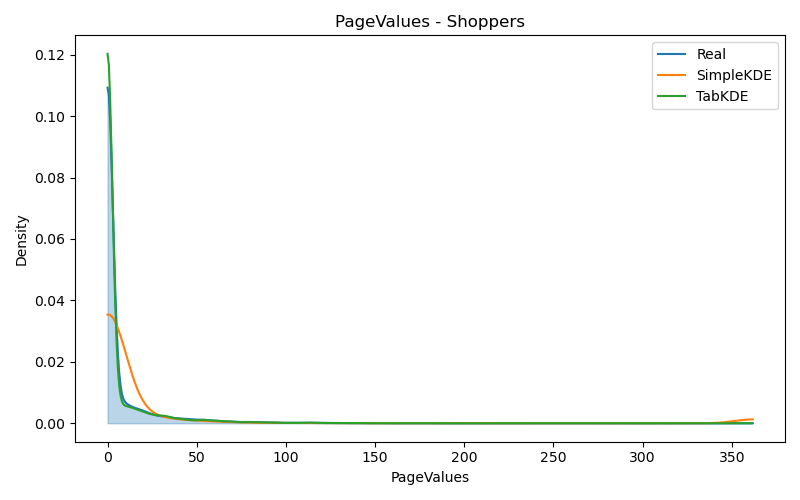} &
        \includegraphics[width=0.225\linewidth]{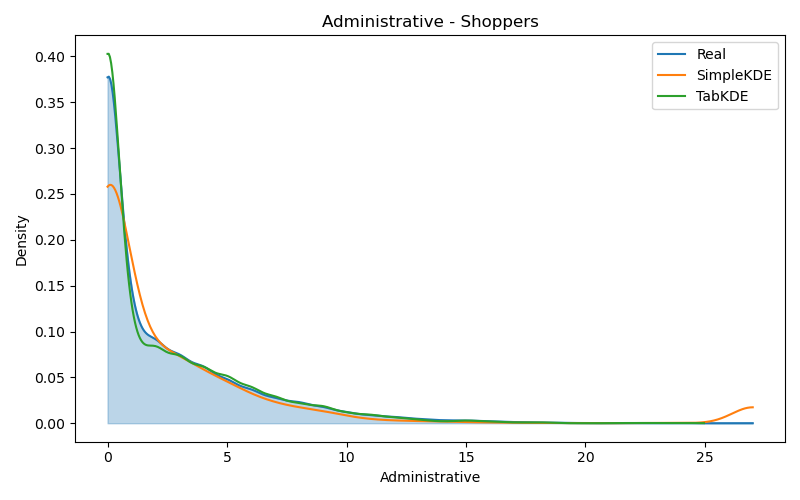} &
        \includegraphics[width=0.225\linewidth]{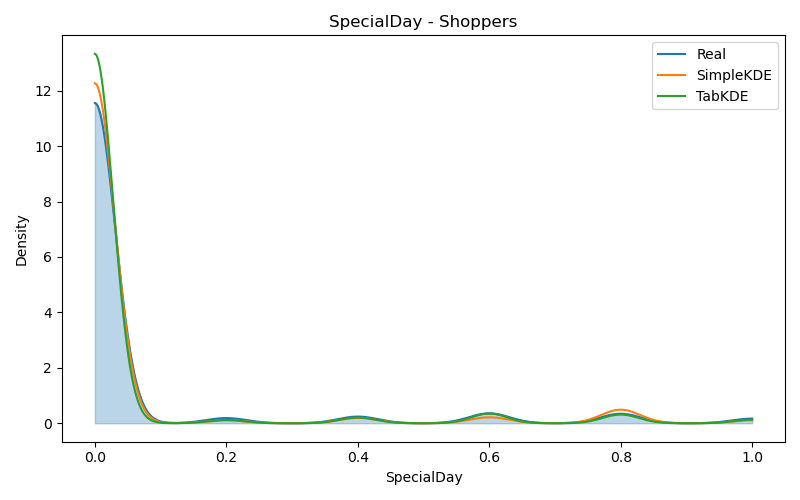} &
        \includegraphics[width=0.225\linewidth]{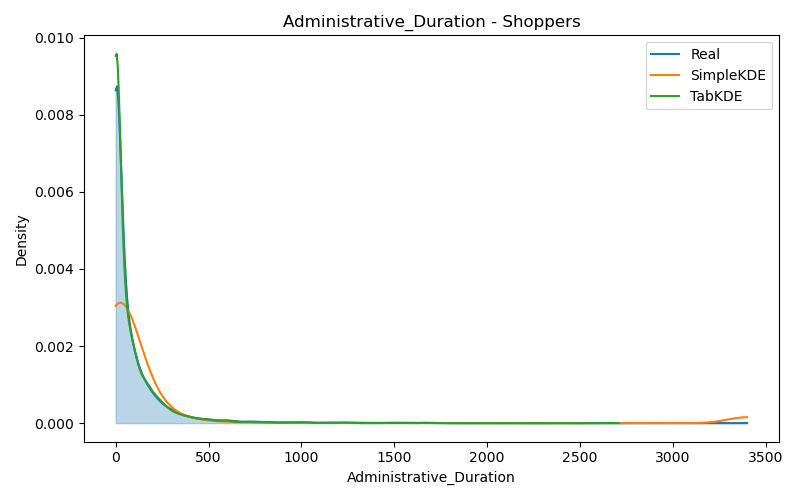} \\
        
        \includegraphics[width=0.225\linewidth]{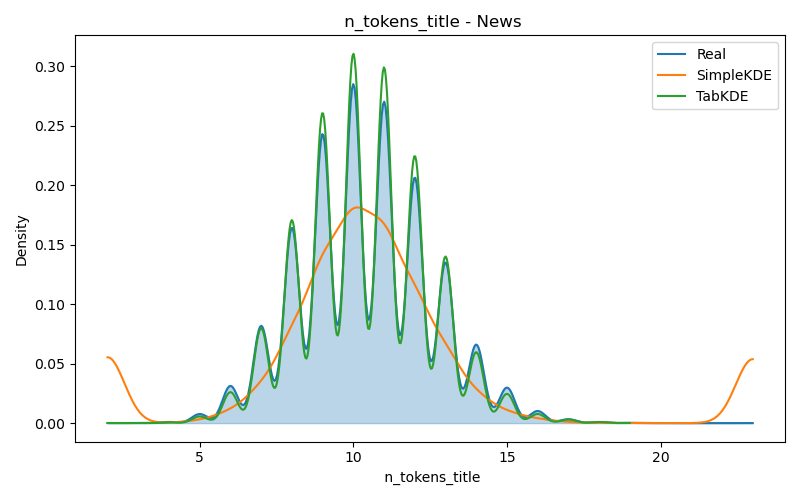} &
        \includegraphics[width=0.225\linewidth]{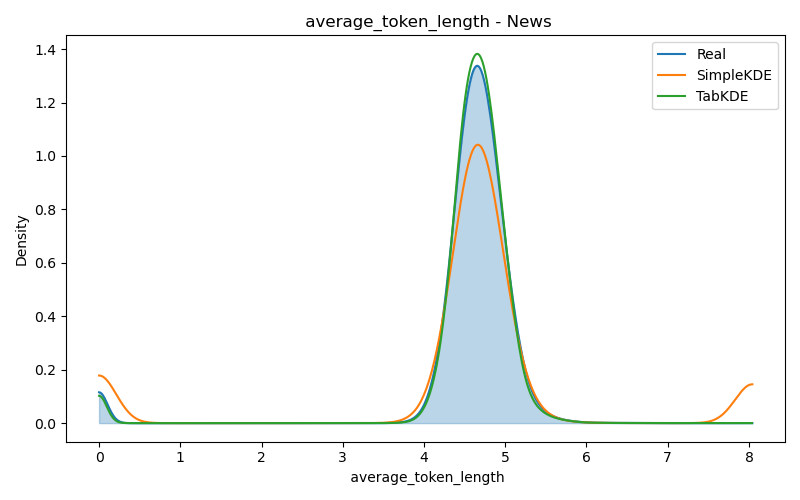} &
        \includegraphics[width=0.225\linewidth]{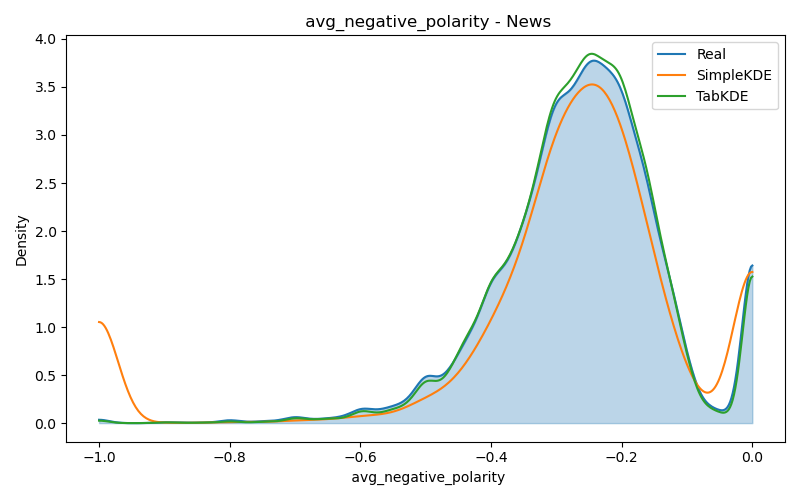} &
        \includegraphics[width=0.225\linewidth]{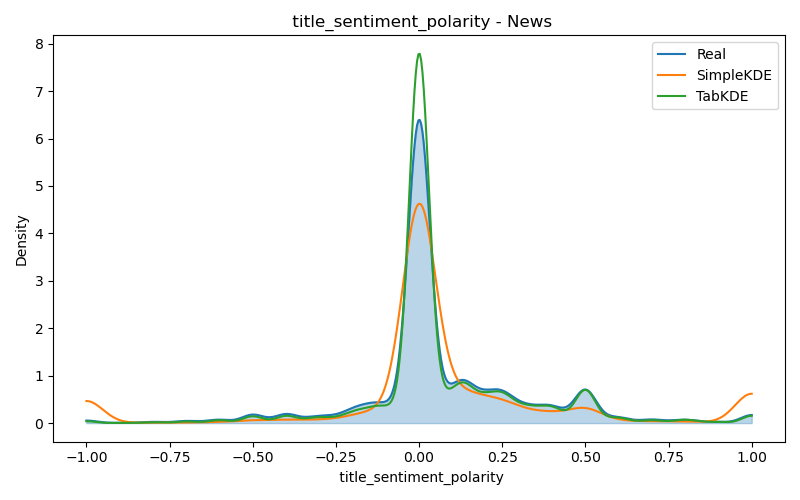} \\
    \end{tabular}
    \caption{Marginals comparison: Simple KDE vs \TabKDE}
    \label{app:fig:marginals:simpleVSTabKDE}
\end{figure}

In our initial exploration, we employed \textbf{SimpleKDE} to generate synthetic samples by perturbing numerical latent representation of the data points using a Kernel Density Estimation (KDE) model. While this approach is intuitive and straightforward, it presents a critical limitation—\textit{lack of boundary control} (see Figure~\ref{app:fig:marginals:simpleVSTabKDE}). The perturbed samples, generated by adding Gaussian noise to real points, often fall outside the convex hull of the original dataset. This results in synthetic records that do not reflect the valid range or domain constraints of the original data, leading to unrealistic samples that may violate the natural boundaries of the feature space.

To address this issue, we developed \TabKDE to better leverage the copula-transformed latent space, where all features reside. We ensure effective boundary control during sampling, maintaining the generated samples within valid limits. By resolving the boundary control problem, TabKDE produces synthetic data that more accurately preserves the statistical structure and integrity of the original dataset (see Figure~\ref{app:fig:marginals:simpleVSTabKDE}).

\subsection{Boundary Correction Ablation for \TabKDE Variants}\label{app:correction-stats}
As discussed in Section~\ref{subsec:ZtoSample}, we employ a boundary-aware sampler to avoid out-of-boundary samples.
Table~\ref{app:boundary-ablation} reports an ablation of boundary correction behavior across datasets, including the average and maximum number of coordinate resampling iterations, as well as the number of sampling failures, defined as cases where some coordinates remain out of bounds after $10 \times d$ resampling attempts.
We refer the reader to Appendix~\ref{app:datasets} for a description of the IBM datasets with and without the modeling trick.

Across all datasets, our boundary correction step is rarely needed: the mean number of coordinate-resampling iterations (denoted "Correction iterations" in Table \ref{app:boundary-ablation}) is below $1$ for every dataset except News, indicating that most sampled coordinates fall within valid bounds on the first draw. Even when correction is required, it remains modest--the maximum number of iterations is at most $9$ across datasets (excluding $1$ failure in News dataset out of the 35679 samples we generate), suggesting fast convergence of the boundary-aware sampler. Sampling failures are essentially absent: we observe zero failures on all datasets except News, which exhibits a single failure under the $10\times d$ resampling budget. For the IBM datasets, the modeling trick reduces the correction burden (mean iterations $0.81 \to 0.58$ and max iterations $8 \to 6$) which is expected since the dimension is decreased from $14$ to $11$ due to the modeling trick.

\begin{table}[ht]
\caption{
Boundary-correction statistics for the boundary-aware sampler used in \TabKDE. For each dataset, we report the mean $\pm$ standard deviation of the number of coordinate-resampling iterations required to obtain an in-bounds sample, the maximum number of iterations observed (for successful samples), and the number of sampling failures (coordinates still out of bounds after $10\times d$ resampling attempts).
}
\centering
\small
\begin{tabular}{lccccc}
\toprule
Dataset        & Size   & Features  &  Correction iterations & Max iterations   & Failed \\
\midrule
Adult          & 32561  & 15        & 0.30 \tiny{$\pm$ 0.52} & 4                & 0      \\
Default        & 27000  & 24        & 0.64 \tiny{$\pm$ 0.74} & 9                & 0      \\
Shoppers       & 11097  & 18        & 0.55 \tiny{$\pm$ 0.65} & 5                & 0      \\
Magic          & 17117  & 11        & 0.38 \tiny{$\pm$ 0.64} & 7                & 0      \\
Beijing        & 37581  & 12        & 0.36 \tiny{$\pm$ 0.56} & 6                & 0      \\
News           & 35679  & 48        & 1.74 \tiny{$\pm$ 0.83} & 8                & 1      \\
IBM(no trick)  & 176221 & 14        & 0.81 \tiny{$\pm$ 0.69} & 8                & 0      \\
IBM(w/ trick)  & 176221 & 11        & 0.58 \tiny{$\pm$ 0.63} & 6                & 0      \\
\bottomrule
\end{tabular}
\label{app:boundary-ablation}
\end{table}

\paragraph{More ``pure'' rejection sampling.}
To extend the ablation study, 
we added two more resampling strategies \textsf{keep-the-seed} resampling and \textsf{change-the-seed} resampling to \TabKDE aside the default coordinate-correction resampling. In all cases, a seed point is first drawn from the empirical training distribution in the transformed latent copula space, and a candidate synthetic point is then produced by perturbing this seed with a sampled radius and a random direction. Because the transformed space is bounded coordinate-wise in interval \([0,1]\), some candidate points might fall outside the valid domain and must be repaired or rejected.

In the coordinate-correction resampling strategy, only the coordinates that violate the domain bounds are resampled, while coordinates that are already valid are preserved. Among the three strategies, this is the most effective at obtaining valid samples. As shown in Table~\ref{app:boundary-ablation}, the average number of correction iterations remains low across all datasets, the observed maxima are modest, and failures are essentially eliminated, with only a single failure observed on News. This indicates that partial coordinate repair is highly robust even in higher-dimensional settings, where the probability that at least one coordinate leaves the feasible region is naturally larger.

In the \textsf{keep-the-seed} strategy, the original seed point is retained, but the entire perturbation is rejected whenever any coordinate falls outside the valid domain. The sampler then retries with a new radius and direction while keeping the same seed, up to a fixed attempt budget. Relative to coordinate correction, this substantially increases the number of retries and the maximum number of iterations, especially on more difficult datasets. Table~\ref{app:boundary-ablation-keep-the-seed} shows that this strategy remains manageable on lower-dimensional datasets such as Adult, Shoppers, and Beijing, but becomes much more expensive on challenging settings such as News and IBM, where the average number of retries and the maximum observed iterations increase sharply. Failures also become non-negligible for some datasets, especially News, indicating that preserving the same seed can make boundary-adjacent seeds costly to sample from successfully.

In the \textsf{change-the-seed} strategy, the sampler makes only one perturbation attempt per seed. If the resulting candidate is invalid, the seed is discarded and a new one is selected. This is the simplest correction rule operationally, but it is also the least tolerant of boundary violations. Table~\ref{app:boundary-ablation-change-the-seed} shows that the number of failures (when we change the seed) becomes very large for every dataset, including moderate-dimensional ones, and grows dramatically for harder cases such as News and IBM. In particular, in News the vast majority of seeds are rejected as seen by the Percentage Failed column.  Compared with the other two strategies, change-the-seed therefore sacrifices acceptance rate for simplicity: it avoids repeated repair attempts, but discards many otherwise usable seeds and strongly amplifies the cost of boundary effects.

Overall, the empirical comparison suggests a clear trade-off. Coordinate-correction resampling provides the best acceptance behavior and the strongest robustness to boundaries. Keep-the-seed preserves the association between an accepted sample and its original seed more strictly, but at a sometimes substantial computational cost. Change-the-seed is algorithmically simple, yet empirically the most wasteful, because it converts many boundary violations directly into failed samples rather than attempting to repair them.
Moreover, as shown in Tables \ref{app:tab:shape}, \ref{app:tab:trend}, \ref{app:tab:IBM-accuracy}, \ref{app:tab:mle}, \ref{app:tab:c2st}, and \ref{app:tab:dcr} the coordinate-correction has the best average accuracy across a wide variety of evaluations.  This supports our decision to use Coordinate-correction as the recommend approach for \TabKDE.

\begin{table}[ht]
\caption{
Boundary statistics for the keep-the-seed resampling strategy. For each selected seed, the sampler repeatedly draws a new radius and direction until an in-bounds perturbation is found or the attempt budget is exhausted. Compared with coordinate correction, this strategy is markedly more sensitive to boundary-adjacent seeds, especially on higher-dimensional or harder datasets.
}
\centering
\small
\begin{tabular}{lccccc}
\toprule
Dataset        & Size   & Features  & Correction iterations & Max iterations   & Failed \\
\midrule
Adult          & 32561  & 15        & 0.51 \tiny{$\pm$ 0.008} & 28.60 \tiny{$\pm$ 5.41}   & 0.00 \tiny{$\pm$ 0.00} \\
Default        & 27000  & 24        & 1.60 \tiny{$\pm$ 0.031} & 165.00 \tiny{$\pm$ 39.96} & 0.80 \tiny{$\pm$ 0.84} \\
Shoppers       & 11097  & 18        & 1.16 \tiny{$\pm$ 0.011} & 33.60 \tiny{$\pm$ 7.70}   & 0.00 \tiny{$\pm$ 0.00} \\
Magic          & 17117  & 11        & 1.04 \tiny{$\pm$ 0.032} & 90.20 \tiny{$\pm$ 14.13}  & 1.40 \tiny{$\pm$ 0.55} \\
Beijing        & 37581  & 12        & 0.63 \tiny{$\pm$ 0.004} & 41.80 \tiny{$\pm$ 9.96}   & 0.00 \tiny{$\pm$ 0.00} \\
News           & 35679  & 48        & 106.57 \tiny{$\pm$ 0.561} & 479.00 \tiny{$\pm$ 0.00} & 6832.40 \tiny{$\pm$ 174.33} \\
IBM (no trick) & 176221 & 14        & 3.12 \tiny{$\pm$ 0.014} & 133.80 \tiny{$\pm$ 4.15}  & 2.60 \tiny{$\pm$ 1.14} \\
IBM (w trick)  & 176221 & 11        & 3.10 \tiny{$\pm$ 0.004} & 134.00 \tiny{$\pm$ 5.70}  & 1.80 \tiny{$\pm$ 0.84} \\
\bottomrule
\end{tabular}
\label{app:boundary-ablation-keep-the-seed}
\end{table}

\begin{table}[ht]
\caption{
Failure statistics for the change-the-seed resampling strategy. In this variant, each seed receives only one perturbation attempt; if the candidate is out of bounds, the seed is discarded and a new seed is drawn. Since no iterative repair is performed, the relevant quantity is the number of failed samples caused by invalid perturbations.  Percentage Failed is $((\mathsf{Size} + \mathsf{Failed})/\mathsf{Size} - 1)\times 100$. 
}
\centering
\small
\begin{tabular}{lcccc}
\toprule
Dataset        & Size   & Features  & Failed & Percentage Failed\\
\midrule
Adult          & 32561  & 15        & 12564.60 \tiny{$\pm$ 194.89} & 38.59\%\\
Default        & 27000  & 24        & 28094.40 \tiny{$\pm$ 194.49} & 104.05\%\\
Shoppers       & 11097  & 18        & 9809.00 \tiny{$\pm$ 211.95} & 88.39\%\\
Magic          & 17117  & 11        & 7575.80 \tiny{$\pm$ 173.88} & 44.26\%\\
Beijing        & 37581  & 12        & 17427.00 \tiny{$\pm$ 197.25} & 46.37\%\\
News           & 35679  & 48        & 2717469.20 \tiny{$\pm$ 23234.80} & 7616.44\%\\
IBM(no trick)  & 176221 & 14        & 346221.80 \tiny{$\pm$ 1525.07} & 196.47\%\\
IBM (w trick)  & 176221 & 11        & 186346.80 \tiny{$\pm$ 383.37} & 105.75\%\\
\bottomrule
\end{tabular}
\label{app:boundary-ablation-change-the-seed}
\end{table}

\subsection{Different Categorical Encodings}
\label{app:ablation-cat-encode}
A defining design choice of \TabKDE is that the intermediate representation $E\in\mathbb{R}^{n\times d}$
(and thus the copula space $Z\subseteq[0,1]^d$) has \emph{one coordinate per original table column}.
This fixed per-column structure makes the copula mapping invertible column-wise and keeps KDE sampling scalable
for mixed data types. As a result, \TabKDE \emph{must} map each categorical column to a \emph{single real-valued coordinate}
(i.e., a continuous code per category) rather than using one-hot expansion. Our default is Principal-Guided Encoding (\PGE),
though the pipeline is modular: any single-scalar encoding can be substituted, potentially at the cost of fidelity.

\begin{figure}
    \centering
    \includegraphics[width=1\linewidth]{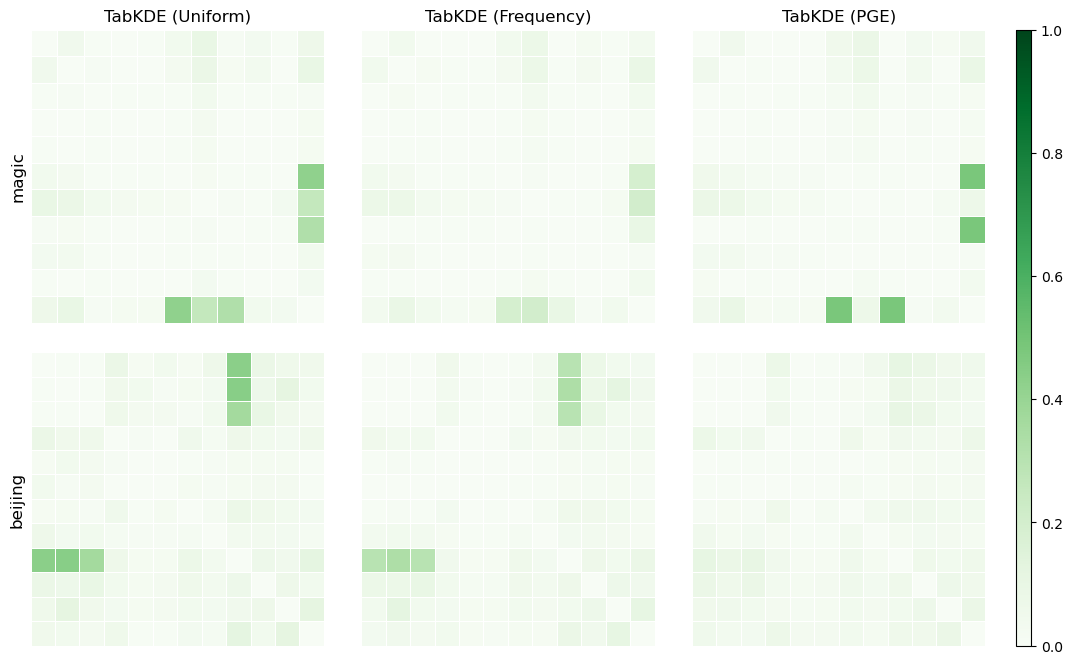}
    \caption{Representative pairwise correlation divergence heatmaps for the Magic and Beijing datasets using \TabKDE with different categorical encodings.
    }\label{fig:magic-beijing-diff-encoding}
\end{figure}

To isolate the effect of \PGE, we re-run \TabKDE and \TabSYN with standard single-scalar alternatives, including uniform and
frequency-based encodings. Uniform encoding assigns each category a sub-interval of $[0,1]$ with width proportional to its
marginal frequency and samples within that interval, yielding an approximately uniform transformed column. Frequency-based
encoding also uses only marginal frequencies, but typically maps each category to a fixed scalar derived from its frequency (or
a related interval statistic). Tables~\ref{tab:marginal_diff_encoding} and~\ref{tab:pairwise_diff_encoding} report marginal and
pairwise alignment errors.  We observe that all $\TabKDE$ variants perform significantly better (about $1.5 - 1.7$) than \TabSYN (about $8.5-10$) in marginal error, with still clear but less dramatic improvements in pairwise correlation error.  
Overall, \PGE and Frequency achieve the lowest average pairwise alignment errors (4.67 and 4.74;
Average column of Table~\ref{tab:pairwise_diff_encoding}), indicating the most consistent cross-type fidelity among single-scalar
encodings as a key contributor to \TabKDE's performance without one-hot expansion.
Figure~\ref{fig:magic-beijing-diff-encoding} further illustrates correlation divergence for Magic and Beijing via heatmaps, showing how the encoding choice affects the correlation structure.  Ultimately, we conclude that Frequency-based encoding could probably replace PGE-encoding without considerable change to the efficacy.  

\begin{table}[h!]
\caption{Marginal distribution alignment error (lower is better) across different categorical encodings that map each category to a single scalar.
}
\centering
\setlength{\tabcolsep}{2.8pt}
\begin{tabular}{lccccccc}
\toprule
\textbf{Method}     & 
\textbf{Adult}      & 
\textbf{Default}    & 
\textbf{Shoppers}   & 
\textbf{Magic}      & 
\textbf{Beijing}    & 
\textbf{News}       & 
\textbf{Average}    \\
\midrule
\FreqTabsyn & 12.12 & 5.53 & 10.02 & 1.62 & 3.82 & 17.99 & 8.52\\
\UniTabsyn  & 10.74 & 5.56 & 10.33 & 1.47 & 1.40 & 22.41 & 8.65\\
\PGETabsyn  & 11.21 & 7.66 & 12.98 & 1.36 & 2.65 & 22.24 & 9.68\\
\hline
\FreqTabKDE & 1.19 & 1.34 & 2.52 & 0.86 & 1.33 & 2.49 & 1.62\\
\UniTabKDE  & 1.58 & 1.01 & 2.07 & 0.75 & 1.44 & 2.48 & 1.56\\
\TabKDE     & 1.56 & 1.55 & 2.44 & 0.78 & 1.37 & 2.52 & 1.70\\
\bottomrule
    \end{tabular}
    \label{tab:marginal_diff_encoding}
\end{table}

\begin{table}[h!]
\caption{Pairwise correlation alignment error (lower is better) across different categorical encodings that map each category to a single scalar. 
}
\centering
\setlength{\tabcolsep}{2.8pt}
\begin{tabular}{lccccccc}
\toprule
\textbf{Method}     & 
\textbf{Adult}      & 
\textbf{Default}    & 
\textbf{Shoppers}   & 
\textbf{Magic}      & 
\textbf{Beijing}    & 
\textbf{News}       & 
\textbf{Average}    \\
\midrule
\FreqTabsyn & 11.03& 11.66& 6.38 & 3.25 & 5.76 & 5.07 & 7.19\\
\UniTabsyn  & 9.18 & 12.48& 6.62  & 3.36 & 4.10 & 7.08 & 7.14\\
\PGETabsyn  & 7.14 & 15.19& 7.56  & 2.67 & 3.49 & 7.12 & 7.20\\
\hline
\FreqTabKDE & 3.94 & 10.62 & 4.05 & 2.51 & 4.51 & 2.80 & 4.74\\
\UniTabKDE  & 6.28 & 15.03 & 4.39 & 3.65 & 5.78 & 2.91 & 6.34\\
\TabKDE     & 4.51 & 9.93 & 4.31 & 2.72 & 3.74 & 2.83 &  4.67\\
\bottomrule
    \end{tabular}
    \label{tab:pairwise_diff_encoding}
\end{table}

\section{Comparison with Copula/KDE Based generator Models}
\label{app:copula-comapre}

Copula-based models represent one of the earliest and most widely used approaches for synthetic tabular data generation. These methods typically rely on two steps: (i) learning marginal distributions for each feature, and (ii) coupling them via a copula function to capture dependencies across variables. Classic examples include the \textit{GaussianCopulaSynthesizer}~\citep{Patki2016}, which estimates univariate marginals and applies a Gaussian copula to model correlations, and its variants such as \textit{CopulaGAN} and vine-copula GANs, which map the copula-transformed data into a latent space before adversarial training~\citep{Xu2019}. While conceptually elegant, these methods often struggle with mixed-type tabular data. Categorical variables are usually handled by simplistic encodings (e.g., one-hot encoding or UniformEncoder), which can distort dependencies, and the Gaussian copula assumption may fail to capture higher-order interactions.

Recent research has extended this line of work. For example, \citep{Meyer2021} applied vine and Gaussian copulas to continuous weather data, while \citep{Majdara2020} integrated copula transforms with diffusion-based KDE for continuous density estimation. However, these methods are limited to continuous domains and do not address challenges of mixed categorical-numerical data or scalability. Other work, such as differentially private copula models (e.g., \citep{Gambs2021GrowingSD}), explicitly aims to strengthen privacy at the expense of accuracy and efficiency.


Against this backdrop, \TabKDE can be viewed as both building on and diverging from the copula tradition. Like classical copula generators, TabKDE maps data into a copula space (not the same as the classical copula method), standardizing marginals into the unit hypercube. However, rather than imposing parametric assumptions (e.g., Gaussian copula) or adversarial training, it employs a $d$-dimensional kernel density estimator directly in copula space. This design introduces several key innovations absent from prior copula models:
\begin{itemize}
    \item \textbf{Principal-Guided Encoding (PGE)} for categorical features, enabling faithful one-dimensional embeddings without using one-hot encodings, which does not increase the dimension and avoids sparsity.  
    
    \item \textbf{Covariance-aware geometry and boundary-respecting kernels}, which allow KDE sampling to preserve higher-order correlations and respect marginal supports.  
    \item \textbf{DCR-calibrated kernels}, which explicitly align synthetic samples to the empirical distance-to-closest-record distribution, thereby privacy protection.  
    \item \textbf{Coreset compression}, which produces compact, scalable generative models, in contrast to copula baselines that typically scale linearly with dataset size.  
\end{itemize}

Empirical comparisons reinforce these differences. Across UCI benchmarks (Adult, Default, Magic,) and a large IBM fraud dataset, TabKDE consistently achieves lower marginal and pairwise correlation errors and substantially higher C2ST fidelity than GaussianCopulaSynthesizer, while also surpassing CopulaGAN in distributional accuracy. GaussianCopulaSynthesizer maintains slightly stronger privacy under the DCR metric (hovering near the ideal $50\%$), but this advantage stems largely from its poorer fidelity. In practice, TabKDE achieves a more balanced tradeoff: reasonable privacy coupled with diffusion-level accuracy, while retaining orders-of-magnitude efficiency gains, training in minutes compared to hours for deep copula-hybrids like CopulaGAN.

To further contextualize our approach, we compare TabKDE against both the 
classical \textit{GaussianCopulaSynthesizer} and  
\textit{CopulaGANSynthesizer}, in addition to our \CopulaDiff variant. 
Table~\ref{tab:avg-copula} reports averages across Adult, Default, Shoppers, 
Beijing, and News, while Tables~\ref{tab:extended_baselines} provide 
per-dataset breakdowns for Adult, Default, Magic, and Beijing.

\begin{table}[ht]
\centering
\begin{tabular}{lccc}
\hline
\textbf{Metric ($\downarrow$ better except C2ST $\uparrow$)} &
\textbf{TabKDE} &
\textbf{CopulaDiff} &
\textbf{GaussianCopula}\\
\hline
Average Marginal error (\%)      & \textbf{1.70} & 1.91 & 14.9\\
Average Pairwise-corr error (\%) & 4.67          & \textbf{3.59} & 13.3\\
Average C2ST ($\uparrow$)                 & \textbf{0.94} & \textbf{0.94} & 0.22\\
DCR (ideal $\approx$ 50\%)       & 58.6          & 51.06 & \textbf{49.8}\\
Train / Sample time (s, CPU)     & 39.2 / 39.0   & --    & \textbf{0.9 / 6.8}\\
\hline
\end{tabular}
\caption{Comparison of \TabKDE and \CopulaDiff against the GaussianCopulaSynthesizer. 
Averages are computed over Adult, Default, Shoppers, Beijing, and News datasets.}
\label{tab:avg-copula}
\end{table}

\paragraph{Accuracy.} TabKDE achieves $7$--$10\times$ lower distribution-alignment errors and quadruples C2ST fidelity compared to the GaussianCopulaSynthesizer. This indicates that our encoding of the copula transform together with the KDE estimator drives the observed accuracy gains.

\paragraph{Speed.} GaussianCopulaSynthesizer is milliseconds-fast on CPU, effectively functioning as a restricted subset of TabKDE. Nonetheless, TabKDE remains far more efficient than deep generative models.

\paragraph{Privacy.} GaussianCopulaSynthesizer yields DCR scores near the ideal $50\%$, reflecting strong privacy but poor fidelity. \TabKDE (58\%) attains a more balanced trade-off, offering reasonable privacy while maintaining high fidelity. Note that DCR is an imperfect metric: it only evaluates nearest-neighbor distances, and broader distributional comparisons are more favorable for \TabKDE.

Overall, \TabKDE consistently outperforms both complex copula-based generators (e.g., CopulaGAN) and the simpler GaussianCopula model in terms of fidelity, while remaining efficient and maintaining reasonable privacy.

\subsection{Extended Copula+KDE Baselines}

We additionally benchmark TabKDE against several methods at the intersection of 
copulas and kernel density estimation (KDE). These include existing implementations in SDV 
and baselines we constructed, along with a differentially private Gaussian Copula model.

\paragraph{Copula+KDE Baselines.}
The comparison space includes a variety of copula-based and KDE-based extensions. Tables~\ref{tab:extended_baselines} present results on the Adult, Default, Magic, and Beijing datasets.

CopulaGAN applies an empirical copula transform followed by GAN-based generation, while GaussianCopula also relies on the copula transform but pairs it with Gaussian marginals.  The GaussianKDECopulaSynthesizer, implemented in SDV with \texttt{default\_distribution=``gaussian\_kde''}, fits Gaussian KDEs on marginals but is extremely slow in practice, requiring about 2.5 hours compared to roughly one minute for TabKDE. We also construct a CopulaKDE baseline, which uses an empirical copula transform followed by a $d$-dimensional Gaussian KDE in latent space $[0,1]^d$ with the bandwidth $\sigma$ set to the median pairwise distance. 

Below the lines in Tables \ref{tab:extended_baselines} are our variants.  
CopulaDiff represents our diffusion-based variant applied after a copula transform.
SimpleKDE is a variant of TabKDE that incorporates a DCR-calibrated kernel and covariance-aware directions, but omits boundary-aware sampling. 
Finally, \TabKDE is our proposed method. 

\begin{table}[ht]
\centering
\begin{tabular}{lcccc}
\hline
\textbf{Adult} & Marginal Err. ($\downarrow$) & Pairwise Corr. ($\downarrow$) & DCR ($\rightarrow$ 50) & C2ST ($\uparrow$) \\
\hline
CopulaGAN              & 8.53\%  & 16.75\% & 49.58\% & 0.60 \\
GaussianCopula         & 12.41\% & 19.24\% & \textbf{50.23\%} & 0.18 \\
CopulaKDE              & 7.39\%  & 14.14\% & 53.00\% & 0.76 \\
GaussianKDECopula (8,461s) & 8.30\% & 8.84\%  & 49.28\% & 0.89 \\
\hline
CopulaDiff             & 2.10\%  & 4.61\%  & 50.34\% & 0.86 \\
SimpleKDE              & 1.98\%  & 4.64\%  & 62.89\% & 0.90 \\
TabKDE                 & \textbf{1.56\%} & \textbf{4.51\%} & 62.23\% & \textbf{0.92} \\
\hline
\end{tabular}\\
\vspace{1cm}
\begin{tabular}{lcccc}
\hline
\textbf{Default} & Marginal Err. ($\downarrow$) & Pairwise Corr. ($\downarrow$) & DCR ($\rightarrow$ 50) & C2ST ($\uparrow$) \\
\hline
CopulaGAN              & 11.50\% & 21.13\% & 52.98\% & 0.74 \\
GaussianCopula         & 12.33\% & 21.90\% & \textbf{49.59\%} & 0.41 \\
CopulaKDE              & 8.99\%  & 12.48\% & 56.36\% & 0.58 \\
GaussianKDECopula (9,239s) & 7.02\% & 7.58\%  & 50.41\% & 0.95 \\
\hline
CopulaDiff             & \textbf{1.47\%} & \textbf{3.29\%} & 50.96\% & \textbf{0.98} \\
SimpleKDE              & 3.33\%  & 5.16\%  & 66.05\% & 0.87 \\
TabKDE                 & 1.55\%  & 9.93\%  & 63.46\% & 0.96 \\
\hline
\end{tabular}\\
\vspace{1cm}
\begin{tabular}{lcccc}
\hline
\textbf{Magic} & Marginal Err. ($\downarrow$) & Pairwise Corr. ($\downarrow$) & DCR ($\rightarrow$ 50) & C2ST ($\uparrow$) \\
\hline
CopulaGAN              & 10.21\% & 9.03\% & 51.23\% & 0.66 \\
GaussianCopula         & 11.19\% & 6.34\% & \textbf{50.10\%} & 0.51 \\
CopulaKDE              & 11.42\% & 7.20\% & 55.40\% & 0.73 \\
GaussianKDECopula (1,477s) & 2.30\% & 5.00\% & 50.22\% & \textbf{0.99} \\
\hline
CopulaDiff             & 0.94\%  & \textbf{1.72\%} & 52.03\% & 0.94 \\
SimpleKDE              & 3.12\%  & 3.30\% & 62.62\% & 0.97 \\
TabKDE                 & \textbf{0.78\%} & 2.72\% & 63.02\% & 0.94 \\
\hline
\end{tabular}\\
\vspace{1cm}
\begin{tabular}{lcccc}
\hline
\textbf{Beijing} & Marginal Err. ($\downarrow$) & Pairwise Corr. ($\downarrow$) & DCR ($\rightarrow$ 50) & C2ST ($\uparrow$) \\
\hline
CopulaGAN              & 7.79\%  & 12.11\% & 50.89\% & 0.78 \\
GaussianCopula         & 10.01\% & 6.00\%  & \textbf{50.05\%} & 0.11 \\
CopulaKDE              & 12.04\% & 17.07\% & 53.94\% & 0.74 \\
GaussianKDECopula (9,288s) & 2.69\% & 6.56\% & 50.66\% & \textbf{0.99} \\
\hline
CopulaDiff             & 2.13\%  & 4.50\%  & 50.29\% & 0.96 \\
SimpleKDE              & 2.06\%  & 4.68\%  & 55.45\% & 0.94 \\
TabKDE                 & \textbf{1.37\%} & \textbf{3.74\%} & 54.24\% & 0.95 \\
\hline
\end{tabular}
\caption{\TabKDE vs Copula+KDE Baselines.}
\label{tab:extended_baselines}
\end{table}

Three central observations arise from the experiments. In terms of accuracy, TabKDE consistently achieves the best or near-best fidelity. While some competitors such as CopulaDiff, SimpleKDE, or GaussianKDECopula perform strongly on C2ST, they typically fall short in marginal or pairwise errors. Regarding privacy, the SDV copula models and the DP Gaussian Copula reach DCR values close to the ideal 50\%, but they incur substantially higher errors. As expected, stronger privacy guarantees correlate with reduced fidelity. TabKDE offers a balanced tradeoff, maintaining high fidelity while still achieving moderate privacy with DCR around 58\%. Finally, in terms of scalability, GaussianKDECopulaSynthesizer is prohibitively slow, taking several hours compared to TabKDE’s single-minute runtime. TabKDE thus emerges as both more efficient and more accurate.

\subsection{Differentially Private Gaussian Copula}
\label{app:DP-GC}
In addition to the above, we benchmark a differentially private (DP) Gaussian Copula model with a specified privacy budget $\epsilon$. Table~\ref{tab:dp-adult} reports the results.  

Note that while this achieves strong DCR score (as do similar copula methods without DP guarantees), there is not a clear advantage as the $\epsilon$ parameter is decreased.  
However, the resulting Marginal Error (about 17\%) and Pairwise Correlation (about 28\%) scores are significantly higher than most other baselines; \TabKDE achieves $1.56\%$ and $4.51\%$.  Also C2ST (about $0.35$) is worse that most baselines; \TabKDE achieves $0.92$, where higher is better.  

So while this method does provide a DP guarantee, it appears to perform significantly worse in all accuracy measures, even for very large $\epsilon$ values.  

\begin{table}[ht]
\centering
\begin{tabular}{lcccc}
\hline
\textbf{DP Gaussian Copula (Adult)} & Marginal Err. ($\downarrow$) & Pairwise Corr. ($\downarrow$) & DCR ($\rightarrow$ 50) & C2ST ($\uparrow$)  \\
\hline
$\epsilon=0.1$   & 18.44\% & 32.56\% & 49.88\% & 0.37 \\
$\epsilon=1$     & 16.72\% & 29.04\% & 50.12\% & 0.36 \\
$\epsilon=5$     & 16.27\% & 28.69\% & 49.97\% & 0.37 \\
$\epsilon=10$    & 16.92\% & 27.76\% & 49.75\% & 0.33 \\
$\epsilon=100$   & 17.46\% & 29.80\% & 49.80\% & 0.29 \\
\hline
\end{tabular}
\caption{DP Gaussian Copula on Adult with different privacy budget $\epsilon$ (smaller implies stronger privacy).}
\label{tab:dp-adult}
\end{table}




\end{document}